\documentclass[runningheads]{llncs}

\usepackage{doc}

\usepackage{abbrv}
\usepackage{makecell}
\usepackage{float}
\usepackage[section]{placeins}
\usepackage{multirow}
\usepackage{multicol}
\usepackage{svg}
 \usepackage{bbm} 
\usepackage{booktabs}
\usepackage{colortbl}
\usepackage{xcolor}
\usepackage{caption}
\usepackage{array}
\usepackage{svg}
\usepackage{xcolor, colortbl}
\definecolor{PrimaryColumnColor}{rgb}{0.90,0.90,0.90}
\definecolor{SecondaryColumnColor}{rgb}{1,1,1}
\definecolor{CyanColor}{rgb}{0.75,1,1}
\definecolor{cvprblue}{rgb}{0.21,0.49,0.74}
\definecolor{GreyTable}{RGB}{225, 225, 225}

\usepackage{enumitem}

\newlength{\tablecaptionvspace}
\usepackage{graphicx}
\usepackage{booktabs}

\usepackage[accsupp]{axessibility}  

\usepackage{hyperref}

\usepackage{orcidlink}

\begin{document}


\title{Unsupervised Source-Free Ranking of Biomedical Segmentation Models Under Distribution Shift}

\titlerunning{Unsupervised Source-Free Model Ranking}

\author{Joshua Talks\inst{1}\and
Kevin Marchesini\inst{2}\and
Luca Lumetti\inst{2}
\and
Federico Bolelli\inst{2}
\and
Anna Kreshuk\inst{1}
}

\authorrunning{J.~Talks et al.}

\institute{EMBL Heidelberg, Meyerhofstraße 1, Germany\\ \email{first.last@embl.de}\and
University of Modena and Reggio Emilia, Italy\\
\email{first.last@unimore.it}\\}

\maketitle

\begin{abstract}
Model reuse offers a solution to the challenges of segmentation in biomedical imaging, where high data annotation costs remain a major bottleneck for deep learning. However, although many pretrained models are released through challenges, model zoos, and repositories, selecting the most suitable model for a new dataset remains difficult due to the lack of reliable model ranking methods. We introduce the first black-box-compatible framework for unsupervised and source-free ranking of semantic and instance segmentation models based on the consistency of predictions under perturbations. While ranking methods have been studied for classification and a few segmentation-related approaches exist, most target related tasks such as transferability estimation or model validation and typically rely on labelled data, feature-space access, or specific training assumptions. In contrast, our method directly addresses the repository setting and applies to both semantic and instance segmentation, for zero-shot reuse or after unsupervised domain adaptation. We evaluate the approach across a wide range of biomedical segmentation tasks in both 2D and 3D imaging, showing that our estimated rankings strongly correlate with true target-domain model performance rankings.


\end{abstract}
\section{Introduction}
\label{sec:intro}

Segmentation is a ubiquitous problem in biomedical images, vital for the interpretation of biological and anatomical structures. While modern deep learning has significantly improved segmentation accuracy, practical use is limited by the labour-intensive annotation of training data. Model transfer~\cite{weiss_survey_2016,mustafa_supervised_2021}, with or without fine-tuning, offers a potential solution by reusing pre-trained ``source'' networks for new ``target'' datasets of interest. In biomedical imaging, where comprehensive public training datasets are rare, and new imaging modalities continue to emerge, practitioners often train domain-specific models~\cite{isensee2021nnu}. Community initiatives such as the BioImage Model Zoo (BMZ)~\cite{ouyang_bioimage_2022}, public challenges such as ToothFairy~\cite{2024TMI}, and emerging large generalist models~\cite{archit_segment_2025,pachitariu_cellpose-sam_2025,stringer_cellpose_2020,wasserthal_totalsegmentator_2023} have created a growing ecosystem of reusable models. This raises the practical challenge of model selection when multiple candidate models---trained on the same task but with varying settings, architectures, or source data---are available.

\begin{figure}[tb]
  \centering
    \includegraphics[width=0.95\linewidth]{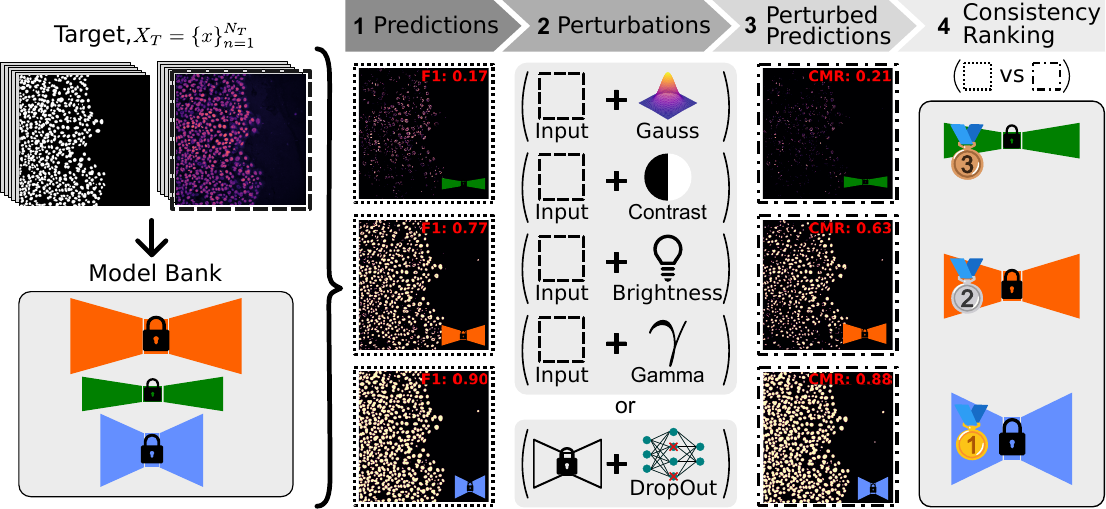}
  \caption{Unsupervised consistency-based model ranking (CMR) for a set of models via pairwise consistency between perturbed and unperturbed predictions.}
  \label{fig:intro_figure}
\end{figure}

Transferability estimation \cite{agostinelli_how_2022} aims to address this task, often in a source-free, model-agnostic setting appropriate for repository model ranking. Most prior work addresses the task of image classification, where multiple supervised methods have been introduced 
\cite{renggli_which_2020,ding_which_2024,ibrahim_newer_2023,you_logme_2021,nguyen_leep_2020,wang_how_2023,bao_information-theoretic_2022,li_ranking_2020} to rank post-fine-tuning performance. 
It has also been suggested that transferability ranking for semantic segmentation can be achieved by simple extension of classification approaches~\cite{agostinelli_how_2022}. In our experiments, this claim does not hold for biomedical imaging data, where classification-derived methods fail to reliably correlate with model target data performance. Furthermore, while semantic segmentation can be expressed as pixel classification, instance segmentation cannot. Instance segmentation outputs a permutation invariant set of labels that arise from model‑dependent post-processing (e.g., connected components analysis), so the labels have no stable, fixed‑dimensional correspondence to instance representation in feature space. The problem of instance segmentation transferability ranking remains largely unaddressed for both natural and biomedical images. 

While supervised transferability estimation can save fine-tuning costs, annotation, especially in 3D, can take weeks of expert time without guaranteeing that a representative set has been labelled. Unsupervised ranking, therefore, has significant practical advantages. Also, recent advances in Unsupervised Domain Adaptation (UDA) enable potential improvements to model target data performance without labels, but still require unsupervised ranking of the adapted models. First such methods have been introduced~\cite{saito_tune_2021,morerio_minimal-entropy_2017,yang_can_2024}, again largely for classification. However, in our experiments, the state-of-the-art (SOTA) for UDA validation~\cite{yang_can_2024} fails to reliably rank semantic segmentation models and, like other classification-derived approaches, cannot be applied to instance segmentation.

While model repositories exist---HuggingFace alone hosts over 2.7 million models—practical ranking methods remain scarce because real-world deployment requires them to be unsupervised, source-free, and model-agnostic---con\-straints that few existing approaches satisfy. To address this gap, we therefore consider the problem of selecting the best segmentation model from a repository for a new dataset without access to target labels.

\noindent \textbf{Contribution}. In summary, our contribution is fourfold:

\begin{enumerate}[label=(\roman*)]
    \item We introduce the first unsupervised, source-free method for ranking both semantic and instance segmentation models based on prediction consistency under feature- or input-space perturbations (\cref{fig:intro_figure});
    
    \item The method is training- and architecture-agnostic and operates only on model predictions. It can be applied to black-box models, making it compatible with heterogeneous model zoos even if using inference APIs or containerised deployments;

    \item We evaluate on realistic biomedical transfer scenarios spanning nuclei, cell, and mitochondria segmentation in light and electron microscopy. Our method successfully ranks models ranging from domain-specific architectures to large generalist segmentation models (e.g., $\mu$SAM~\cite{archit_segment_2025} and CellposeSAM~\cite{pachitariu_cellpose-sam_2025}), both for direct application and after unsupervised domain adaptation;

    \item We further demonstrate the practical applicability of our approach by ranking submissions from the ToothFairy challenge~\cite{2025CVPR}, a benchmark for multi-class 3D CBCT semantic segmentation. Without access to test dataset labels, our method successfully reproduces the challenge’s average model DICE ranking.

\end{enumerate}

\section{Related Work}
In the following, we review supervised transferability metrics, UDA validation, performance estimation, generalisation, and uncertainty estimation while highlighting the distinct challenges of unsupervised model ranking. These fields address distinct yet related problems. While they share many methodological approaches, they use different terminology and rarely cross-cite.

\smallskip
\noindent
\textbf{Supervised Transferability Metrics.}
Transferability metrics (TM) aim to rank a set of models by their post-fine-tuned target performance. Existing work follows two main categories: \textit{(i)} label-comparison-based methods, such as LEEP \cite{nguyen_leep_2020}, and \textit{(ii)} feature-embedding-based methods, which analyse the target data embedding space with guidance from target labels ($\mathcal{N}$LEEP~\cite{li_ranking_2020}, LogME~\cite{you_logme_2021},  H-score~\cite{bao_information-theoretic_2022}, Regularised H-score~\cite{ibrahim_newer_2023}, GBC~\cite{pandy_transferability_2021}, NCTI~\cite{wang_how_2023}). They all require target domain labels and were initially designed for classification, except CC-FV~\cite{yang_pick_2023}, which focuses on semantic segmentation. Our task and method are unsupervised, but fundamentally aim to answer a very similar question of selecting the most suitable source models for a given domain-shifted target dataset~\cite{agostinelli_how_2022}. Hence, many TM methods satisfy our requirements of being source-free and model agnostic and can serve as strong baselines for ranking `zero-shot transferability'. \textit{Our key difference:} label-free, no fine-tuning assumption.

\smallskip
\noindent
\textbf{UDA Validation.} Many popular UDA methods rely on self or adversarial training~\cite{kumar_understanding_2020,cao_multi-modal_2023} and are notoriously unstable~\cite{wang_deep_2018, wang_characterizing_2019}. Hence, UDA validation addresses a problem close to ours: how to select the best adapted model without target domain labels. The SOTA in this field is the Transfer Score (TS)~\cite{yang_can_2024}, developed for classification, but also tested on segmentation. It measures the post-UDA model classifier bias in the target domain along with the feature space transferability and discriminability. We use TS as a further baseline in our experiments. \textit{Our key difference:} ranking across heterogeneous models, not only UDA setting.

\smallskip
\noindent
\textbf{Performance Estimation.} Another related field is unlabelled performance estimation (PE), also known as quality control (QC), where the aim is to directly estimate performance metrics. However, in PE, the unit of analysis is typically a single model, and methods aim to assess the quality of the model's predictions in deployment, e.g., under distribution drift. In contrast, our method estimates the relative success of a candidate set of models applied to specific new domain-shifted target data. Thus we do not require the score to equate to a direct performance metric for each model, as we aim to recover a relative ranking across the model set.
PE methods need to rely on assumptions about the nature of the domain shift to uniquely identify the target conditional~\cite{garg_leveraging_2022}. At the same time, PE methods are not required to be target label-free~\cite{valindria_reverse_2017}, source-free~\cite{garg_leveraging_2022,bialek_estimating_2025} or model-training agnostic~\cite{lin_novel_2023,hoebel_exploration_2020,roy_bayesian_2019,robinson_real-time_2018,kivimaki_confidence-based_2025,mehrtash_confidence_2020}, making most of them inapplicable for model selection in the unsupervised model repository setting. Using a recent benchmark~\cite{yu_odp-bench_2025}, we identify a few exceptions that satisfy our requirements, namely Nuclear Norm~\cite{deng_confidence_2023} and Dispersion~\cite{xie_importance_2023}. The first seeks to quantify the confidence and dispersity of model predictions via the nuclear norm, while the second, similarly to~\cite{pandy_transferability_2021,yang_pick_2023} seeks to use inter-class feature dispersion as a surrogate for model performance. We implement both as baselines. \textit{Our key difference:} repository ranking vs single-model QC, source-free, model-training agnostic.

Ensemble-based PE methods~\cite{sims_seg_2023, baek_agreement---line_2023} incur a large computational overhead, but can in principle be applied to ranking, with \cite{sims_seg_2023} specifically proposed for segmentation. However, \cite{sims_seg_2023} reduces the evaluation to object centroids, making the method insensitive to object shape, especially for irregularly shaped objects. Missing pixelwise object evaluation, the method is closer to object detection than instance segmentation ranking, but still provides a useful baseline as the only other approach to address instance segmentation. 

\smallskip
\noindent
\textbf{Generalisation and Uncertainty Estimation.} Generalisability is another term used to describe the ability of a model to perform across new datasets. While often used for in-domain (ID) data, out-of-distribution (OOD) generalisability also exists. Typically, it is considered as an intrinsic model property that can be improved, e.g., by learning more robust and broadly applicable features. Our aim is not to improve target performance, yet we can exploit similar approaches to those measuring generalisability, but in a specific paired model-dataset setting. Ranking differs from generalisability as it explicitly measures the compatibility between a model and a target dataset either directly or post-UDA and is not a property of the model alone.
Consistency-based approaches~\cite{deng_strong_2022,schiff_predicting_2021,k_robustness_2021} provide potentially suitable unsupervised methods and have been explored for classification. Early works~\cite{k_robustness_2021,schiff_predicting_2021} suggest that prediction invariance under test-time-augmentation (TTA) can be linked to model generalisability, but do not explore feature perturbations. Furthermore, they only focus on the ID setting, relying on carefully weighted penalty scores~\cite{k_robustness_2021}, or directly using ground-truth~\cite{schiff_predicting_2021}  to evaluate prediction invariance. Effective Invariance (EI)~\cite{deng_strong_2022} addresses OOD generalisability, measuring both the consistency and confidence of probabilistic model outputs. 
However, EI cannot be extended to instance segmentation.

Consistency to perturbations has also been investigated in the uncertainty estimation (UE) field, where rich theory ties the variance of predictions to an approximation of Bayesian neural networks (BNNs)~\cite{gal_dropout_2016,loquercio_general_2020,ledda_dropout_2023,mi_training-free_2022,wang_aleatoric_2019,wang_epistemic_2024}. Calibrating model confidence to correlate with true performance is a key challenge and often requires access to source/target labelled validation sets~\cite{loquercio_general_2020,ledda_dropout_2023} or restriction of the training procedure~\cite{gal_dropout_2016}. The UE field itself is not concerned with assessing model applicability to target data, but rather aims to improve model interpretability via the quantification of uncertainty. However, UE has been widely used in many PE, QC, or failure detection methods~\cite{zenk_comparative_2025,lin_novel_2023,devries_leveraging_2018,hoebel_exploration_2020,kivimaki_confidence-based_2025,mehrtash_confidence_2020,jungo_analyzing_2020,roy_bayesian_2019} that explicitly conflate estimating performance and uncertainty, thus suffering the same limitations. These approaches are typically developed to assist medical experts in segmentation and provide per-image evaluation of a single model's predictions. While important, such methods are not suitable for model selection in model zoos, adding unnecessary, hard to satisfy uncertainty calibration requirements (i.e., restricting the model training~\cite{devries_leveraging_2018,hoebel_exploration_2020,roy_bayesian_2019,kivimaki_confidence-based_2025,mehrtash_confidence_2020}). \textit{Our key difference:} no uncertainty calibration, semantic and instance segmentation, model-agnostic.

Following source-free, model-agnostic, unsupervised constraints of a model repository setting, we use the building blocks from the above fields to create a ranking-specific approach, stripping away complexities associated with confidence calibration or feature space reasoning, and explicitly do not use perturbations for UE, as described in \cref{sec:methods}.
Our method, for the first time, can be applied in a unified framework for both semantic and instance segmentation ranking.
\section{Methods}
\label{sec:methods}
We rank model performance via the consistency of model outputs under perturbation. Focusing on output analysis enables a fully unsupervised and source-free ranking.
Since the output spaces of candidate models are fixed and directly comparable---unlike their feature spaces, which may differ in composition and dimensionality---our approach provides a uniform evaluation basis. Moreover, our method, for the first time, directly assesses pixel-wise instance segmentation performance across arbitrary instantiation methods. SEG~\cite{sims_seg_2023} has previously attempted to address instance segmentation, but reduces the problem to object detection based on centroids. This constrains SEG to circular objects and disregards pixel-wise information. We motivate our consistency-based approach as follows: a source model learns decision boundaries to partition data by class. When applied to a distribution-shifted target domain, these boundaries may fail to separate classes effectively or may intersect embedding clusters. To assess model-dataset compatibility, we introduce small perturbations---either to the model input or its features---and observe predicted class flips as pixel embeddings cross decision boundaries. Measuring such pixel class flips provides a model-agnostic estimate of compatibility, free from feature space-based reasoning used in prior methods (e.g., feature-space compactness or separability~\cite{pandy_transferability_2021,li_ranking_2020,xie_importance_2023,deng_confidence_2023,wang_how_2023,yang_can_2024,yang_pick_2023}).
This is advantageous for semantic/instance segmentation, where common losses (e.g., Dice and cross-entropy) do not explicitly enforce feature compactness.

Prediction consistency can be viewed as a proxy for the margin of a model’s decision boundaries on target data. Models that transfer well are expected to exhibit larger margins, leading to better-separated classes and less perturbation sensitive predictions. This intuition aligns with prior discussions on manifold smoothness and margin distribution~\cite{jiang_predicting_2018,ng_predicting_2023}.
Without access to labels we rely on perturbation-induced class changes to indicate separability.
From this perspective, input-image perturbations (commonly in generalization) and feature-space perturbations (popular in uncertainty estimation) are conceptually equivalent.
 
Consistency perturbations need not represent realistic image changes between source and target distributions; e.g., feature-space perturbations do not necessarily have realistic image interpretations. Their purpose is to perturb pixel embeddings, testing the suitability of a model's decision boundary for target data by measuring the number of output class label flips. 
We explicitly do not use perturbations to estimate uncertainty, and do not use MC-dropout~\cite{gal_dropout_2016}, which is used in many PE methods for calibrated confidence scores and must be applied during both training and inference. Instead, we employ test-time-dropout (TTD)~\cite{tompson_efficient_2015,ledda_dropout_2023} which can be applied to arbitrary layers at inference time only. Thus remaining model agnostic and setting no model training requirements. 

\subsection{Problem Definition}

Considering a pre-trained model $M_{S}: X_{S} \rightarrow Y_{S}$, trained on a source domain $\mathcal{D}_{S}= \{(x_{n}, y_{n})\}^{N_{S}}_{n=1}$ we apply $M_S$ to an unlabelled target dataset $X_{T}$, sampled from a shifted target domain $\mathcal{D}_{T} = \{x_{n}\}^{N_{T}}_{n=1}$. Inference on $X_{T}$ can be performed directly or after UDA, which further trains $M_{S}$ on a subset of $\mathcal{D}_T$.
We enforce that the source and target task are the same (e.g., nuclei segmentation). Given a set of candidate source models $\mathcal{M} = \{M_{m}\}_{m=1}^{N_{M}}$, our goal is to rank the models on $X_T$ so that the ranking correlates with true performance $\mathcal{P}_{M_{m}(X_{T})}$ (e.g., F1 score) ranking, which is not available without labels. We introduce an unsupervised consistency-based model ranking method $\mathcal{R}_{M_{m}(X_{T})}$ (CMR) that aims to preserve the ranking $\mathcal{R}_{M_{i}(X_{T})} > \mathcal{R}_{M_{j}(X_{T})}$ if and only if $\mathcal{P}_{M_{i}(X_{T})} > \mathcal{P}_{M_{j}(X_{T})}$.

\subsection{Our Consistency Model Ranking (CMR)}
\label{subsec:cte}

Considering a target dataset $X_{T} = \{x_{n}\}_{n=1}^{N_{T}}$, we rank the transfer performance of a set of models $\mathcal{M} = \{M_{m}\}_{m=1}^{N_{M}}$ on $X_{T}$ by measuring pixelwise prediction consistency between perturbed and unperturbed predictions (\cref{fig:intro_figure}).
Firstly, we propose to extend the Effective Invariance (EI)~\cite{deng_strong_2022}, originally formulated for classification tasks, to semantic segmentation. CMR-EI measures the per-class invariance of a single segmentation prediction $\hat{y}_{n}$ to test-time perturbations,
\begin{equation}
    \text{CMR-EI}_n^{(c)} = \frac{1}{|\mathcal{I}^{(c)}_{n}|} \sum_{i \in \mathcal{I}^{(c)}_{n}} \sqrt{\hat{p}_{n,i}^{(c)} \cdot \tilde{p}_{n,i}^{(c)}} \cdot \mathbbm{1}[\tilde{y}_{n,i}^{(c)}=\hat{y}_{n,i}^{(c)}],
\end{equation} where $\hat{p}_{n,i}^{(c)}$ and $\hat{y}_{n,i}^{(c)}$ represent the pixelwise softmax and thresholded output of a model for class $c$, with $\tilde{p}_{n,i}^{(c)}$, $\tilde{y}_{n,i}^{(c)}$ the perturbed equivalents. To counteract high-class imbalance in biomedical data and allow for multiclass problems, we calculate a foreground restricted per-class consistency score, limiting iteration to the pixel set $\mathcal{I}^{(c)}_{n}=\{i:\tilde{y}^{(c)}_{n,i}=1 \; \text{or} \; \hat{y}^{(c)}_{n,i}=1\}$, thus preventing background pixels from dominating. CMR-EI is a ``soft'' metric as it utilises both output consistency and confidence, introducing a reliance on confidence calibration which is not guaranteed in transfer~\cite{ovadia_can_2019}. We therefore also propose a ``hard'' consistency measure based on the thresholded per-class binary model output. Using the normalised Hamming distance (NHD), we count the number of foreground restricted pixelwise label changes between the perturbed and unperturbed predictions,

\begin{equation}
    \text{CMR-NHD}_n^{(c)} = 1 - \frac{1}{|\mathcal{I}_n^{(c)}|}\sum_{i\in \mathcal{I}_n^{(c)}}
\mathbbm{1}[\tilde{y}_{n,i}^{(c)}\neq \hat{y}_{n,i}^{(c)}].
\end{equation}
This is equivalent to the $IoU$, but considering pixel flips adds intuition. The consistency score for a single prediction $\hat{y}_{n}$ is the average of the per-class scores.

For instance segmentation, direct individual pixel comparison is impossible as instance labels are arbitrary. Instead, we measure the consistency between $\tilde{y}_n$ and $\hat{y}_n$ via the agreement of paired pixels in each prediction, as measured by the Rand Index~\cite{rand_objective_1971,unnikrishnan_measures_2005}. Intuitively, if a pair of pixels is assigned to a single instance in $\tilde{y}_n$, then to be consistent, this should also be true in $\hat{y}_n$. To account for class imbalance, we use the foreground-restricted Adapted Rand Score (ARS)~\cite{arganda-carreras_crowdsourcing_2015},

\begin{equation} \text{CMR-ARS}_n = \frac{\sum_{k,\ell} w_{k\ell}^2}{\alpha \sum_k \tilde{s}_k^2 + (1 - \alpha) \sum_\ell \hat{s}_\ell^2},
\label{eq:ARS}
\end{equation} 
where $w_{k \ell}$ is the proportion of pixels that belong to instance $k$ in $\tilde{y}_n$ and instance $\ell$ in $\hat{y}_n$, restricted to the pairwise union of foreground pixels in $\tilde{y}_n$ and $\hat{y}_n$. $\tilde{s}_{k}=\sum_{\ell}w_{k \ell}$ represents the proportion of pixels with label $k$ in $\tilde{y}_n$ and $\hat{s}_{\ell}=\sum_{k}w_{k\ell}$ is similarly defined for $\hat{y}_n$. Thus, $\sum_{k,\ell}w^{2}_{k\ell}$ is the proportion of pixel pairs belonging to a single instance in both $\hat{y}_n$ and $\tilde{y}_n$. From the perspective of either prediction, \cref{eq:ARS} can then be considered as the weighted harmonic mean between two terms: 
\begin{equation}
    \text{ARS}_{\text{split}}=\frac{\sum_{k,\ell}w_{k\ell}^{2}}{\sum_{\ell}\hat{s}_{\ell}}, \;\;\;\;\;\;\;\;
    \text{ARS}_{\text{merge}}=\frac{\sum_{k,\ell}w_{k\ell}^{2}}{\sum_{k}\tilde{s}_{k}}.
\end{equation}
$\text{ARS}_{\text{split}}$ penalises split disagreements and is the proportion of pixels pairs in one instance in $\tilde{y}_n$ given they are in one instance in $\hat{y}_n$. $\text{ARS}_{\text{merge}}$ penalises merge disagreements and is the proportion of pixel pairs in one instance in $\hat{y}_n$ given they are in one instance in $\tilde{y}_n$. The weighting $\alpha=0.5$ is used by default. Following convention, every background pixel is treated as an instance containing a single pixel.
The consistency score for a single transfer $M_{m}(X_{T})$ is then calculated as: 
\begin{equation}
    \text{CMR}_{M_{m}(X_{T})} = \underset{n=1,\ldots,N_T}{\text{median}}\left( \frac{1}{N_{\text{pert}}}\sum\limits_{j=1}^{N_{\text{pert}}}\text{consis}\!\left(\hat{y}_{n,m}, \tilde{y}^{(j)}_{n,m}\right) \right),
\end{equation}
where $\text{consis}\!\left(\hat{y}_{n,m}, \tilde{y}^{(j)}_{n,m}\right)$ is calculated using one of the above metrics (CMR-EI, CMR-NHD, CMR-ARS), and we average over $N_\text{pert}$ perturbations. The final ranking within $\mathcal{M}$ is the descending order of  $\text{CMR}_{M_{m}(X_{T})}$.

\subsection{Model Perturbation}
\label{sec:model_perturbation}
Perturbations can be applied to inputs via TTA or directly to feature space, provided model access. We aim to assess model decision boundaries via target prediction consistency. This differs from evaluating the model’s learned invariance to perturbations within the source domain (e.g., due to training augmentations).
To examine this, we rank models trained both with and without augmentation, showing perturbations still expose informative variations on target data. Perturbations must remain tolerable~\cite{mi_training-free_2022}: if too strong, even good models become inconsistent, while if overly weak all models stay stable. Empirically, we find a broad range of perturbations enable effective ranking via CMR (see \textit{Supp.}).

\smallskip
\noindent
\textbf{Input Space Perturbations.}
We explored several popular image transformations: additive Gaussian noise, gamma correction, and changes in brightness and contrast. In our experiments, we found similar performance across all transformations. Thus, for brevity, in the main text we mostly only report  Gaussian noise, with strength controlled by $\sigma$.
See \textit{Supp.}\ 
and \cref{tab:multiple_metrics_trimmed} 
for additional TTAs.

\smallskip
\noindent
\textbf{Feature Perturbation.}
Perturbations can be applied directly to intermediate feature layers, while keeping the network weights frozen. Motivated by findings in semi-supervised learning~\cite{ouali_semi-supervised_2020} and uncertainty estimation~\cite{mi_training-free_2022}, we apply test-time dropout~\cite{tompson_efficient_2015,ledda_dropout_2023} (TTD). Specifically, we apply TTD \textit{(i)} across all layers of the network, \textit{(ii)} only at the bottleneck, where information is the most redundant~\cite{he_reshaping_2014} (\cref{tab:semantic_seg_correlations,tab:Instance_seg_corr}) or \textit{(iii)} at the bottleneck and skip connection layers (\cref{tab:multiple_metrics_trimmed}).

\section{Experimental Setting}
\label{sec:experiments}
\subsection{Datasets}

We evaluate our approach on a range of public segmentation datasets, covering six distinct tasks and many modalities. For each model ranking run, we fix one target dataset and evaluate all task-specific models (see \textit{Supp.}\ for more details).

\noindent \textbf{Electron Microscopy (EM), Mitochondria.}
We use four semantic segmentation datasets with neural tissues from different species and EM modalities: ~\cite{lucchi_learning_2013,franco-barranco_current_2023,phelps_reconstruction_2021}. We also reformulated these datasets as a per-patch binary classification task, where a positive label indicates a mitochondrion is present. This auxiliary task allows us to disentangle biomedical domain effects~\cite{chaves_performance_2023} from the intrinsic suitability of classification-based approaches for semantic segmentation.

\noindent {\textbf{Light Microscopy (LM), Nuclei.}
We use nine nuclei datasets for instance and semantic segmentation, taken from a variety of species and LM modalities: ~\cite{vijayan_deep_2024,von_chamier_democratising_2021,kromp_annotated_2020,hawkins_rescu-nets_2024,ljosa_annotated_2012,caicedo_nucleus_2019,arvidsson_annotated_2023,de_helacytonuc_2024,leal-taixe_benchmark_2019}} (see \textit{Supp.} for detailed source target splits).

\noindent {\textbf{Light Microscopy, Cells.}
We consider four instance cell segmentation datasets, taken from fruit fly, \textit{Arabidopsis thaliana} and human samples:~\cite{leal-taixe_benchmark_2019,wolny_accurate_2020,willis_cell_2016,pape_microscopy-based_2021}}.

\noindent {\textbf{Cone-Beam Computed Tomography (CBCT), ToothFairy2.}
We use the ToothFairy2 datasets taken from the MICCAI2024 challenge~\cite{2024TMI,2025CVPR}, comprising 3D CBCT volumes of the human jaw with 42 labelled classes. Model ranking is assessed on a held-out test set acquired at another institution.
}

\subsection{Models and Correlation Evaluation Metrics}
\label{sec:models}
Our diverse set of models include self-trained and pre-trained candidates. For classification: ResNet18/50~\cite{he_deep_2015}, DenseNet121~\cite{huang_densely_2018}, MobileNetV2/V3~\cite{sandler_mobilenetv2_2018,Qian_MobileNetV3}. For binary semantic and instance segmentation: 2D U-Net~\cite{ronneberger_u-net_2015}, Residual U-Net~\cite{lee_superhuman_2017}, UNETR~\cite{hatamizadeh_unetr_2022}, $\mu \text{SAM}$~\cite{archit_segment_2025}, Cellpose-SAM~\cite{pachitariu_cellpose-sam_2025}, and BMZ~\cite{ouyang_bioimage_2022} model, id: 5092850 `powerful-chipmunk'~\cite{pape_microscopy-based_2021}. For multi-class semantic segmentation, we select top-models from the ToothFairy2 challenge~\cite{wang2025supervised} (Isensee \etal and Wang \etal) and further add architectures covering CNNs (2D nnU-Net, residual-encoder 3D nnU-Net~\cite{isensee2021nnu}), Transformers (SwinUNETR~\cite{hatamizadeh2021swin}, TotalSegmentator~\cite{wasserthal_totalsegmentator_2023}) and State Space Models (UMamba~\cite{Ma2024}, VMamba~\cite{vmamba}).

For all ranking methods $\mathcal{R}_{M_{m}(X_{T})}$, we calculate the correlation between the metric ranking and the performance score $\mathcal{P}_{M_{m}(X_{T})}$ ranking. As standard~\cite{agostinelli_how_2022}, we evaluate the correlation scores using three complementary measures: Pearson correlation coefficient $(P_{r})$~\cite{benesty_pearson_2009}, which captures linear relationships; Spearman's rank correlation coefficient $(S_{\rho})$~\cite{spearman_proof_1904}, which measures monotonicity; and Kendall tau $(K_{\tau})$~\cite{1972RankCM}, which assesses ordinal association and is more sensitive to local pairwise rank disagreements. All three metrics range from $[-1, 1]$ with $K_{\tau} = P_{r} = S_{\rho} = 1$ indicating perfect positive correlation. While a perfect $\mathcal{R}_{M_{m}(X_{T})}$ should be monotonically related to $\mathcal{P}_{M_{m}(X_{T})}$, linear correlation is not required.

\section{Results}
\subsection{Semantic Segmentation}

\begin{figure*}[tb]
  \centering
    \includegraphics[width=0.95\linewidth]{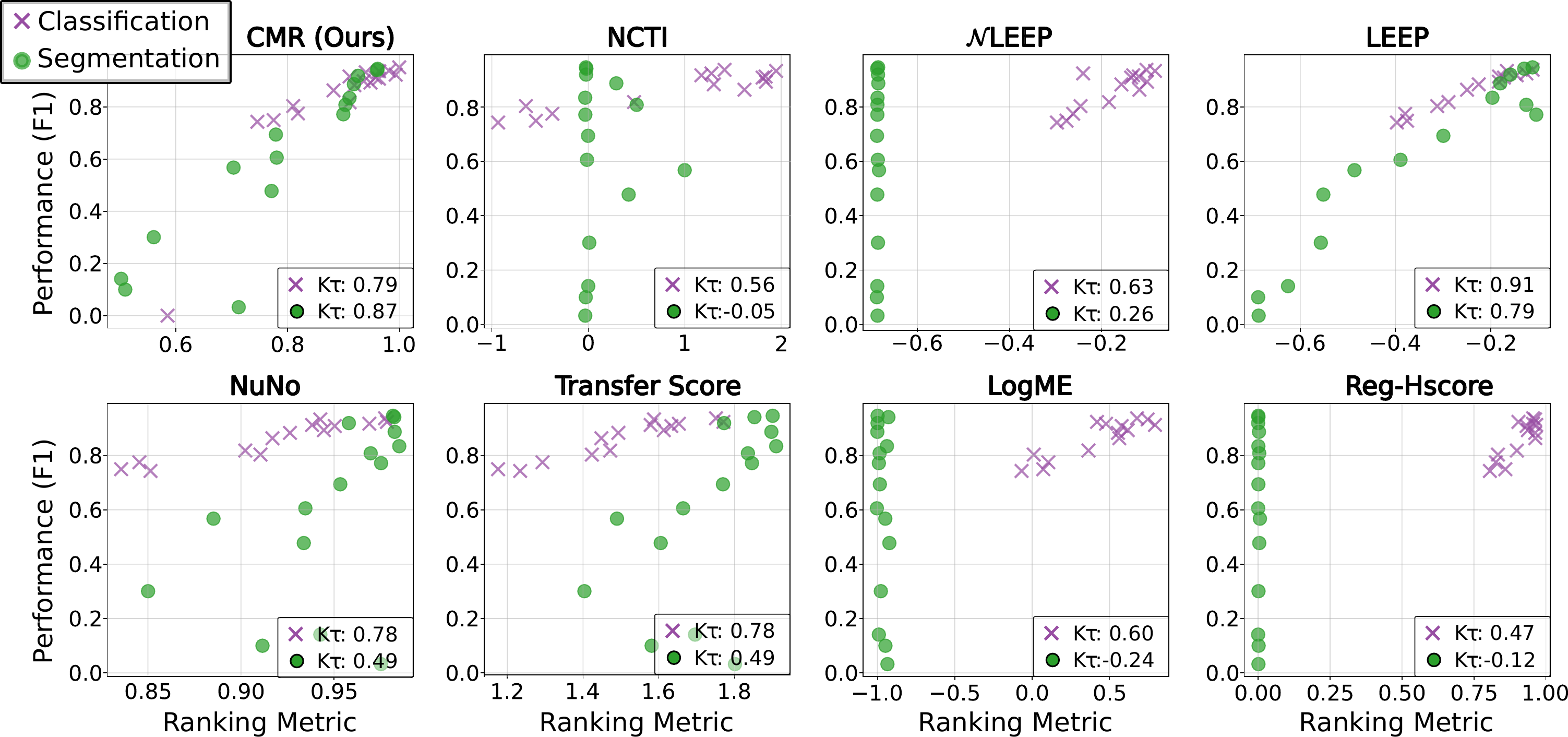}
    \label{fig:class_vs_seg}
  \caption{Classification \& semantic segmentation ranking (EPFL~\cite{lucchi_learning_2013}). Point per model.}
  \label{fig:class_seg_trasnfer_metrics}
\end{figure*}

Previous work~\cite{agostinelli_how_2022,yang_pick_2023,pandy_transferability_2021} suggested that supervised transferability metrics developed for classification can be extended to semantic segmentation by sampling a class-balanced set of pixels and treating them as a classification dataset. However, for our biomedical datasets this approach fails: existing transfer metrics, even with supervision, do not correlate with either direct transfer performance (\cref{tab:semantic_seg_correlations_supervised}) or post–fine-tuning performance (see \textit{Supp.}).
To determine whether this limitation arises from the biomedical domain~\cite{chaves_performance_2023} or from the shift to segmentation, we additionally evaluated the metrics on mitochondria classification. \cref{fig:class_seg_trasnfer_metrics} shows model performance (F1) against each metric for a single target dataset (EPFL~\cite{lucchi_learning_2013}) for both classification and semantic segmentation. While the metrics correlate well with classification performance (violet crosses), they fail for segmentation (green circles), where only CMR and supervised LEEP remain reliable.

\begin{table*}[tbp]
    \centering
    \scriptsize
    \caption{Semantic-segmentation. Mean correlation to F1 ranking over 4 target datasets each. Mito~\cite{lucchi_learning_2013,franco-barranco_current_2023,phelps_reconstruction_2021} ($|\mathcal{M}|$=15) and Nuclei~\cite{von_chamier_democratising_2021,caicedo_nucleus_2019,ljosa_annotated_2012,arvidsson_annotated_2023}($|\mathcal{M}|$=7). $\dagger$ notes our metrics.}\vspace{\tablecaptionvspace}
    \label{tab:semantic_seg_correlations}
    
    \begin{subtable}[t]{0.48\textwidth}
        \centering
        \tiny
        \setlength{\tabcolsep}{2.5pt}
        \begin{tabular}{@{}c l 
       >{\columncolor{GreyTable}}c
       >{\columncolor{GreyTable}}c
       >{\columncolor{GreyTable}}c 
       c c c@{}}
        \toprule
        \multirow{2}{*}{Metric} & {} & \multicolumn{3}{c}{Mito} & \multicolumn{3}{c}{Nuclei} \\
        
        \cmidrule(lr){3-5}
        \cmidrule(lr){6-8}
          &  & \cellcolor{SecondaryColumnColor}K$\tau$ & \cellcolor{SecondaryColumnColor}S$\rho$ & \cellcolor{SecondaryColumnColor}P$r$ & K$\tau$ & $\rho$ & P$r$ \\
        \midrule
        \multirow{2}{*}{\begin{tabular}{@{}c@{}} CMR-EI $\dagger$ \\ (Gauss)\end{tabular}} & \textit{$\mu$} & \textbf{0.77} & 0.85 & 0.83 & 0.62 & 0.77 & \textbf{0.98} \\
          & \textit{$\sigma$} & ±\textbf{0.1} & ±0.1 & ±0.1 & ±0.1 & ±0.1 & ±\textbf{0.0} \\
        \cmidrule{1-8}
         \multirow{2}{*}{\begin{tabular}{@{}c@{}} CMR-NHD $\dagger$\\ (Gauss)\end{tabular}} & \textit{$\mu$} & 0.71 & 0.83 & 0.79 & 0.69 & 0.82 & 0.97 \\
          & \textit{$\sigma$} & ±0.1 & ±0.1 & ±0.1 & ±0.1 & ±0.1 & ±0.0 \\
        \cmidrule{1-8}
         \multirow{2}{*}{\begin{tabular}{@{}c@{}} CMR-EI $\dagger$\\ (DropOut)\end{tabular}} & \textit{$\mu$} & 0.73 & \textbf{0.86} & \textbf{0.86} & \textbf{0.74} & \textbf{0.85} & 0.90 \\
          & \textit{$\sigma$} & ±0.1 & ±\textbf{0.1} & ±\textbf{0.1} & ±\textbf{0.2} & ±\textbf{0.1} & ±0.1 \\
        \cmidrule{1-8}
         \multirow{2}{*}{\begin{tabular}{@{}c@{}} CMR-NHD $\dagger$\\ (DropOut)\end{tabular}} & \textit{$\mu$} & 0.69 & 0.85 & 0.84 & 0.71 & 0.84 & 0.62 \\
          & \textit{$\sigma$} & ±0.1 & ±0.1 & ±0.1 & ±0.2 & ±0.1 & ±0.4 \\
        \cmidrule{1-8}
         \multirow{2}{*}{TS} & \textit{$\mu$} & 0.25 & 0.30 & 0.23 & 0.02 & 0.10 & -0.09 \\
          & \textit{$\sigma$} & ±0.2 & ±0.3 & ±0.3 & ±0.3 & ±0.4 & ±0.4 \\
          \cmidrule{1-8}
         \multirow{2}{*}{NuNo} & \textit{$\mu$} & 0.17 & 0.20 & 0.13 & 0.09 & 0.17 & 0.08 \\
          & \textit{$\sigma$} & ±0.2 & ±0.3 & ±0.3 & ±0.5 & ±0.5 & ±0.5 \\
          \cmidrule{1-8}
         \multirow{2}{*}{Dispersion} & \textit{$\mu$} & -0.03 & -0.07 & -0.17 & -0.18 & -0.29 & -0.19 \\
          & \textit{$\sigma$} & ±0.0 & ±0.1 & ±0.1 & ±0.3 & ±0.4 & ±0.5 \\
         \bottomrule
        \end{tabular}
        \subcaption{Unsupervised}
        \label{tab:semantic_seg_correlations_unsupervised}
    \end{subtable}
    \hfill
    \begin{subtable}[t]{0.48\textwidth}
        \centering
        \tiny
        \setlength{\tabcolsep}{2.5pt}
        \begin{tabular}{@{}c l 
       >{\columncolor{GreyTable}}c
       >{\columncolor{GreyTable}}c
       >{\columncolor{GreyTable}}c 
       c c c@{}}
        \toprule
        \multirow{2}{*}{Metric} & {} & \multicolumn{3}{c}{Mito} & \multicolumn{3}{c}{Nuclei} \\
        
        \cmidrule(lr){3-5}
        \cmidrule(lr){6-8}
          &  & \cellcolor{SecondaryColumnColor}K$\tau$ & \cellcolor{SecondaryColumnColor}S$\rho$ & \cellcolor{SecondaryColumnColor}P$r$ & K$\tau$ & S$\rho$ & P$r$ \\
        \midrule
        \multirow{2}{*}{CCFV} & \textit{$\mu$} & -0.11 & -0.16 & -0.18 & -0.03 & -0.05 & -0.16 \\
          & \textit{$\sigma$} & ±0.1 & ±0.1 & ±0.1 & ±0.3 & ±0.4 & ±0.5 \\
        \cmidrule{1-8}
         \multirow{2}{*}{NLEEP} & \textit{$\mu$} & 0.41 & 0.49 & 0.39 & 0.11 & 0.12 & 0.32 \\
          & \textit{$\sigma$} & ±0.4 & ±0.5 & ±0.3 & ±0.1 & ±0.1 & ±0.2 \\
        \cmidrule{1-8}
        \multirow{2}{*}{LEEP} & \textit{$\mu$} & 0.89 & 0.96 & 0.97 & 0.67 & 0.77 & 0.94 \\
        & \textit{$\sigma$} & ±0.1 & ±0.0 & ±0.0 & ±0.4 & ±0.3 & ±0.0 \\
        \cmidrule{1-8}
         \multirow{2}{*}{GBC} & \textit{$\mu$} & 0.39 & 0.52 & 0.47 & -0.13 & -0.16 & -0.08 \\
          & \textit{$\sigma$} & ±0.1 & ±0.1 & ±0.2 & ±0.3 & ±0.4 & ±0.4 \\
        \cmidrule{1-8}
         \multirow{2}{*}{LogME} & \textit{$\mu$} & 0.08 & 0.10 & 0.06 & -0.20 & -0.28 & -0.49 \\
          & \textit{$\sigma$} & ±0.2 & ±0.3 & ±0.3 & ±0.3 & ±0.4 & ±0.4 \\
        \cmidrule{1-8}
         \multirow{2}{*}{NCTI} & \textit{$\mu$} & 0.21 & 0.28 & 0.37 & 0.29 & 0.25 & 0.30 \\
          & \textit{$\sigma$} & ±0.2 & ±0.2 & ±0.2 & ±0.5 & ±0.6 & ±0.3 \\
        \cmidrule{1-8}
         \multirow{2}{*}{RegHscore} & \textit{$\mu$} & 0.24 & 0.34 & 0.37 & 0.35 & 0.36 & 0.34 \\
          & \textit{$\sigma$} & ±0.2 & ±0.4 & ±0.3 & ±0.4 & ±0.4 & ±0.3 \\
         \bottomrule
        \end{tabular}
        \subcaption{Supervised}
        \label{tab:semantic_seg_correlations_supervised}
    \end{subtable}
    \vspace{-10pt}
\end{table*}

\cref{tab:semantic_seg_correlations} reports the mean correlation across sets of 4 target datasets for mitochondria segmentation in EM~\cite{lucchi_learning_2013,franco-barranco_current_2023,phelps_reconstruction_2021} and nuclei segmentation in LM~\cite{von_chamier_democratising_2021,caicedo_nucleus_2019,ljosa_annotated_2012,arvidsson_annotated_2023} (see \textit{Supp.} for per-dataset results). We evaluate CMR using both ``hard'' (CMR-NHD) and ``soft'' (CMR-EI) consistency metrics under input (Gauss) and feature (DropOut) perturbations. In all cases, CMR shows strong correlation with segmentation F1, outperforming the other unsupervised baselines—Transfer Score (TS), NuNo, and Dispersion. Most supervised transferability metrics also fail despite access to target labels, with LEEP as the only consistent exception. Most baseline methods operate in feature space and rely on assumptions about feature space geometry, such as inter-class separability or intra-class compactness~\cite{yang_can_2024,xie_importance_2023,deng_confidence_2023,pandy_transferability_2021,wang_how_2023}. The contrast between LEEP and $\mathcal{N}$LEEP, which adds feature-space reasoning, illustrates the difficulty of extending such assumptions from classification to segmentation. In contrast, CMR operates directly in the output space, comparing predictions rather than labels and thus enabling fully unsupervised ranking. Among the evaluated approaches, CMR provides the most reliable method for ranking models in practical repository settings.

\begin{figure}[tb]
    \centering
    \includegraphics[width=0.95\linewidth]{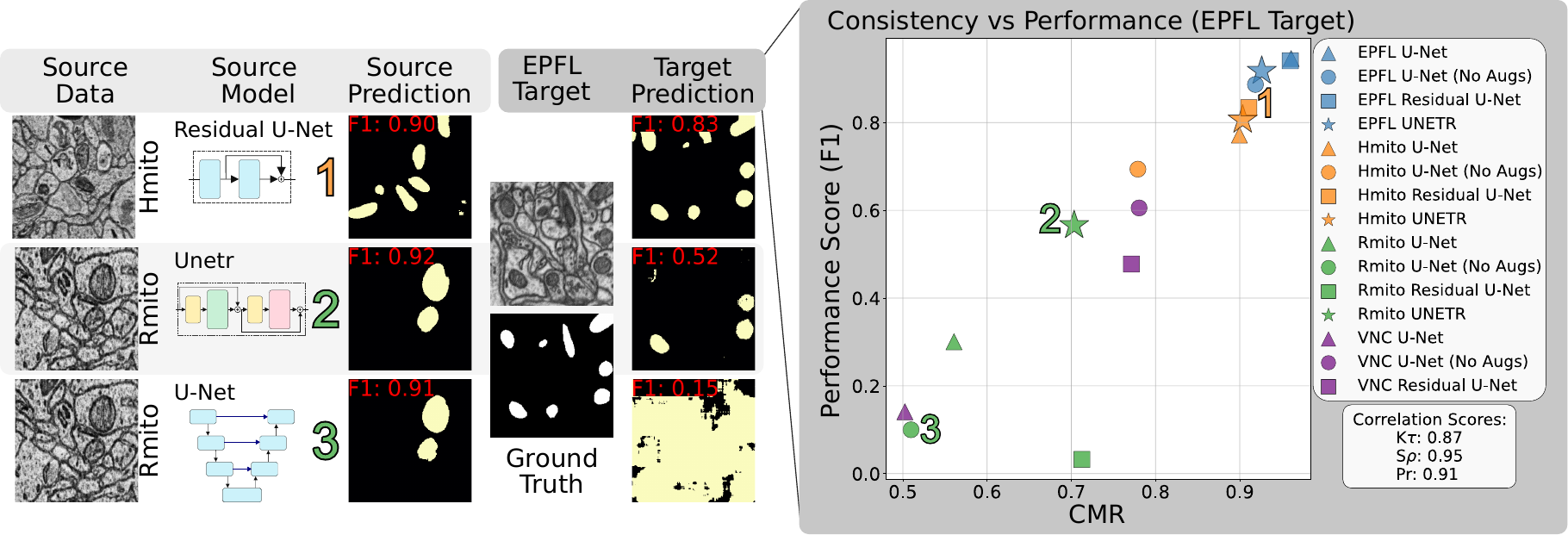}
    \caption{Correlation between F1 \& CMR for Semantic segmentation (EPFL~\cite{lucchi_learning_2013} target).}
    \label{fig:Semantic_EPFL}
\end{figure}

\cref{tab:semantic_seg_correlations} shows all CMR variants exhibit strong linear and monotonic correlations with performance across all target datasets. Comparable results between input- and feature-space perturbations confirm both approaches provide effective ranking. For feature-space perturbations, we examined where in the network they should be applied. Prior work~\cite{mi_training-free_2022,he_reshaping_2014} suggests intermediate layers, where representations contain the most redundant information. To test generality, we applied TTD either to all layers (Mito) or only to the bottleneck (Nuclei). As shown in \cref{tab:semantic_seg_correlations} and the \textit{Supp.}, both strategies yield similarly strong correlations, but with different effective $p_d$ ranges.

Our model sets include networks trained both with and without data augmentations; the consistently high CMR correlations indicate that ranking is unaffected by source-domain augmentation strategies. ``Hard'' (CMR-NHD) and ``soft'' (CMR-EI) metrics perform equivalently well for semantic segmentationlikely because the large number of pixel-wise predictions per image provides a strong signal for the ranking score. UE methods~\cite{mi_training-free_2022} typically require multiple inference runs ($\approx10$). In contrast, we find that measuring consistency with a single perturbation per test image is sufficient for reliable model ranking, requiring only two inference passes per image. For both input and feature-space perturbations, a broad range of perturbation strengths produces stable rankings (see \textit{Supp.}); therefore all main tables report results for a single perturbation setting.

\cref{fig:Semantic_EPFL} illustrates CMR performance on the EPFL~\cite{lucchi_learning_2013} mitochondria segmentation dataset across a diverse set of architectures (U-Net, Residual U-Net, UNETR) trained on multiple source datasets with and without input augmentations. Although models perform similarly on their respective source datasets, transfer performance varies substantially, and simple source–target similarity does not reliably predict ranking. For example, UNETR (2) and U-Net (3), trained on the same Rmito dataset, show markedly different target performance. CMR captures both architectural properties and source-domain effects through the learned weights, directly assessing model suitability for the target dataset.
While CMR generally correlates strongly with transfer performance, pathological cases can cause mis-rankings. These occur primarily among low-performing transfers where models exhibit ``mode collapse'' (predicting all foreground or background), producing artificially high consistency. Such cases could be mitigated by combining consistency with cross-model prediction diversity to distinguish stable models from collapsed predictions.

\begin{table}[tb]
    \centering
    \tiny
    \caption{Correlation on multi-class semantic segmentation of the ToothFairy2 dataset $|\mathcal{M}|=8$. \textit{pval.} $<$ 0.05 (*), \textit{pval.} $<$ 0.01 (**).\vspace{\tablecaptionvspace}}
    \setlength{\tabcolsep}{1pt}
    \begin{tabular}{@{}c c
   >{\columncolor{GreyTable}}c
   >{\columncolor{GreyTable}}c
   cc
   >{\columncolor{GreyTable}}c
   >{\columncolor{GreyTable}}c
   cc
   >{\columncolor{GreyTable}}c
   >{\columncolor{GreyTable}}c
   cc
   >{\columncolor{GreyTable}}c
   >{\columncolor{GreyTable}}c
   cc
   >{\columncolor{GreyTable}}c
   >{\columncolor{GreyTable}}c@{}}
    \toprule
    \makecell{\\Class} & &
    \multicolumn{2}{c}{\makecell{CMR-EI  \\ (Gauss)}} &
    \multicolumn{2}{c}{\makecell{CMR-NHD  \\ (Gauss)}} &
    \multicolumn{2}{c}{\makecell{CMR-EI  \\ (DropOut)}} &
    \multicolumn{2}{c}{\makecell{CMR-NHD  \\ (DropOut)}} &
    \multicolumn{2}{c}{\makecell{CMR-EI  \\ (Gamma)}} &
    \multicolumn{2}{c}{\makecell{CMR-NHD  \\ (Gamma)}} &
    \multicolumn{2}{c}{TS} &
    \multicolumn{2}{c}{NuNo} &
    \multicolumn{2}{c}{Dispersion} \\
    \midrule
    \multirow{3}{*}{\rotatebox[origin=c]{90}{IACs}}
      & K$\tau$ & 0.90 & (**) & 0.81 & (**) & 0.87 & (*) & 0.87 & (*) & \textbf{1.00} & (**) & 0.90 & (*) & 0.21 & (0.2)  & 0.14 & (0.4) & 0.07  & (0.4) \\
      & S$\rho$ & 0.96 & (**) & 0.93 & (*) & 0.94 & (**) & 0.94 & (**) & \textbf{1.00} & (**) & 0.96 & (**) & 0.22 & (0.1) & 0.14 & (0.2) & 0.17 &  (0.3)\\
      & P$r$    & \textbf{0.99} & (**) & 0.98 & (**) & 0.83 & (*) & 0.80 & (0.1) & \textbf{0.99} & (**) & 0.98 & (**) & 0.45 & (0.1)  & 0.01 & (0.4)  & 0.08 & (0.3)  \\
    \midrule
    \multirow{3}{*}{\rotatebox[origin=c]{90}{Teeth}}
      & K$\tau$ & 0.81 & (**) & \textbf{1.00} & (**) & 0.67 & (0.1) & 0.75 & (0.1) & 0.71 & (*) & 0.90 & (**) & 0.36 & (0.1) & 0.07 & (0.3) & 0.43 & (*) \\
      & S$\rho$ & 0.89 & (*) & \textbf{1.00} & (**) & 0.77 & (0.1) & 0.77 & (0.1) & 0.86 & (*) & 0.96 & (**) & 0.55 & (0.1)  & 0.21 & (0.2) & 0.60  & (*)  \\
      & P$r$    & 0.95 & (**) & \textbf{0.98} & (**) & 0.85 & (**) & 0.94 & (**) & 0.89 & (**) & 0.95 & (**) & 0.55 & (*)  & -0.01 & (0.4) & 0.49  & (0.1)  \\
    \midrule
    \multirow{3}{*}{\rotatebox[origin=c]{90}{Mand.}}
      & K$\tau$ & 0.62 & (*) & 0.52 & (0.2) & \textbf{0.87} & (*) & \textbf{0.87} & (**) & 0.43 & (0.3) & 0.81 & (**) & 0.09 & (0.3)  & 0.07 & (0.3) & 0.18 & (0.2)  \\
      & S$\rho$ & 0.82 & (*) & 0.68 & (0.1) & 0.87 & (*) & \textbf{0.94} & (*) & 0.54 & (0.2) & 0.89 & (**) & 0.17 & (0.2)  & 0.12 & (0.3) & 0.25 & (0.2)  \\
      & P$r$    & 0.73 & (0.1) & 0.65 & (0.1) & 0.89 & (**) & \textbf{0.96} & (**) & 0.74 & (0.1) & 0.80 & (*) & 0.02 & (0.3)  & 0.23 & (0.2)  & 0.38 & (0.1) \\
    \midrule
    \multirow{3}{*}{\rotatebox[origin=c]{90}{Sinus.}}
      & K$\tau$ & 0.43 & (0.3) & 0.81 & (*) & 0.87 & (*) & 0.83 & (*) & 0.43 & (0.2) & \textbf{0.90} & (**) & 0.57 & (*)  &  -0.07 & (0.3) & 0.34 & (0.1)  \\
      & S$\rho$ & 0.50 & (0.3) & 0.89 & (*) & 0.84 & (*) & 0.94 & (*) & 0.57 & (0.2) & \textbf{0.96} & (**) & 0.64 & (*)  & -0.05 & (0.4) & 0.35  & (0.2)  \\
      & P$r$    & 0.38 & (0.4) & 0.73 & (0.1) & 0.88 & (**) & 0.90 & (**) & 0.38 & (0.4) & \textbf{0.98} & (**) & 0.53 & (0.1)  & -0.01 & (0.4) & 0.54  & (*)  \\
    \midrule
    \multirow{3}{*}{\rotatebox[origin=c]{90}{Overall\hspace{-2pt}}}
      & K$\tau$ & 0.71 & (*) & 0.90 & (**) & 0.93 & (*) & 0.73 & (0.1) & 0.81 & (**) & \textbf{1.00} & (**) & 0.33 & (0.1) & 0.25 & (0.1) & 0.28 & (0.1) \\
      & S$\rho$ & 0.86 & (*) & 0.96 & (**) & 0.95 & (*) & 0.83 & (0.1) & 0.89 & (*) & \textbf{1.00} & (**) & 0.49 & (0.2)  & 0.28 & (0.2) & 0.50 & (0.1)  \\
      & P$r$    & 0.93 & (**) & \textbf{0.98} & (**) & 0.93 & (**) & 0.93 & (**) & 0.90 & (**) & 0.96 & (**) & 0.55 & (0.2)  & 0.26 & (0.2) & 0.44 & (0.1) \\
    \bottomrule
    \end{tabular}
    \label{tab:multiple_metrics_trimmed}
\end{table}

\cref{tab:multiple_metrics_trimmed} shows highly multi-class 3D experiments on a public challenge dataset ToothFairy2~\cite{2024TMI,2025CVPR}. Here, both our CMR-EI and CMR-NHD metrics remain reliable, although CMR-NHD is slightly better, likely due to poor model calibration on the challenge test set~\cite{ovadia_can_2019}. Correlation scores are computed after aggregating semantically similar classes, i.e., averaging left/right Inferior Alveolar Canals into IACs, the 32 teeth into Teeth, and left/right Maxillary Sinuses into Sinus. The Overall score shows the mean across all 42 classes in the dataset. \cref{tab:multiple_metrics_trimmed} also reports three competitor metrics adapted to the 3D multi-class setting; since these metrics are inherently multi-class, we report the correlation between the overall metric ranking and the corresponding per-group model performance. Once again, both input and feature space perturbations employed by our method yield the strongest and most coherent rank correlations. 
Splitting the class-grouped results reveals our metrics are effective and stable across a morphologically heterogeneous set of classes. Our results largely reproduce the challenge rankings without access to target labels, suggesting a practical approach for self-assessing challenge submissions when the test dataset is available. Notably, our metrics are substantially more memory efficient than competitors: also, the CMR-EI variant only requires the maximum probability of each voxel, i.e., predicted class, whereas TS, NuNo, and Dispersion require storing all softmax probabilities, which is prohibitive in 3D (e.g., more than 200GB for ToothFairy2 dataset).

\smallskip
\noindent
\textbf{UDA Validation.} UDA may be applied to improve model performance, but still requires ranking the adapted models. \cref{tab:finetuned_semantic} shows ranking correlation scores across four mitochondria datasets and two UDA methods: Mean Teacher (MT)~\cite{tarvainen_mean_2018} self-supervised training and adapted batch normalisation (AdaBN)~\cite{li_adaptive_2018}. We observe strong correlation between CMR and post-UDA F1-score on EPFL, Hmito and Rmito for both MT and AdaBN, outperforming the previous SOTA for UDA validation, Transfer Score (TS)~\cite{yang_can_2024}. VNC exhibits lower correlation scores for both CMR and TS metrics, particularly under the MT approach. For CMR, correlation scores are reduced by a few pathological models, all with EPFL source (see \textit{Supp.}), that consistently segment non-mitochondrial structures in the target data, yielding stable but incorrect predictions and, consequently, low task performance. Consistency regularisation in MT training reinforces this class confusion between visually similar, but distinct structures present in EPFL and VNC, violating the assumption that source and target models address the same task. Hence, models can be ranked well under direct transfer but exhibit outlier behaviour after MT self-training. This is further highlighted by the AdaBN UDA approach, which lacks a consistency regularisation component and thus remains less affected, resulting in higher correlation scores. Post-UDA ranking is inherently more challenging, as target data becomes part of the model's adapted `source' distribution, potentially reducing the impact of perturbations used in CMR estimation, yet our approach maintains good correlation.

\begin{table}[tb]
\centering
\begin{minipage}[t]{0.56\textwidth}
    \centering
    \tiny
    \caption{Post-UDA correlation to semantic F1-score $|\mathcal{M}|=12$. $\dagger$ notes our metric. Best per group (Mean Teacher/AdaBN) and dataset in bold. \textit{pval.} $<$ 0.05 (*), \textit{pval.} $<$ 0.01 (**).\vspace{\tablecaptionvspace}}
    \setlength{\tabcolsep}{1.7pt}
    \begin{tabular}{@{}c c c
       >{\columncolor{GreyTable}}c
       >{\columncolor{GreyTable}}c
       c c
       >{\columncolor{GreyTable}}c
       >{\columncolor{GreyTable}}c
       cc@{}}
    \toprule
    {} & \multirow{2}{*}{Metric} & {} & \multicolumn{2}{c}{EPFL} & \multicolumn{2}{c}{Hmito} & \multicolumn{2}{c}{Rmito} & \multicolumn{2}{c}{VNC} \\
    {} &  &  &  \cellcolor{SecondaryColumnColor} & \cellcolor{SecondaryColumnColor}\textit{pval.} &  & \textit{pval.} &  \cellcolor{SecondaryColumnColor} & \cellcolor{SecondaryColumnColor}\textit{pval.} &  & \cellcolor{SecondaryColumnColor}\textit{pval.} \\
    \cmidrule{1-11}
    {} & \multirow{3}{*}{\begin{tabular}{@{}c@{}}CMR-EI $\dagger$ \\ \textit{Gauss}\end{tabular}} & K$\tau$ & \textbf{0.78} & (**) & \textbf{0.75} & (**) & \textbf{0.64} & (*) & \textbf{0.21} & (0.4) \\
    {} &  & S$\rho$ & \textbf{0.91} & (**) & \textbf{0.83} & (**) & \textbf{0.81} & (**) & \textbf{0.24} & (0.4) \\
    {} &  & P$r$ & \textbf{0.93} & (**) & \textbf{0.76} & (**) & \textbf{0.70} & (*) & \textbf{0.28} & (0.4) \\
    \cmidrule{2-11}
    {} & \multirow{3}{*}{\begin{tabular}{@{}c@{}}TS\end{tabular}} & K$\tau$ & 0.45 & (0.1) & 0.42 & (0.1) & 0.38 & (0.1) & 0.00 & (1.0) \\
    \multirow{-5}{*}{\rotatebox[origin=c]{90}{\parbox{1cm}{\centering Mean Teacher}}} &  & S$\rho$ & 0.57 & (0.1) & 0.57 & (0.1) & 0.50 & (0.1) & 0.05 & (0.9) \\
     &  & P$r$ & 0.44 & (0.2) & 0.56 & (0.1) & \textbf{0.70} & (*) & 0.03 & (0.9) \\
    \midrule
    {} & \multirow{3}{*}{\begin{tabular}{@{}c@{}}CMR-EI $\dagger$ \\ \textit{Gauss}\end{tabular}} & K$\tau$ & \textbf{0.70} & (**) & \textbf{0.73} & (**) & \textbf{0.55} & (*) & \textbf{0.39} & (0.2) \\
    {} &  & S$\rho$ & \textbf{0.86} & (**) & \textbf{0.88} & (**) & \textbf{0.73} & (**) & \textbf{0.47} & (0.2) \\
    {} &  & P$r$ & 0.76 & (**) & \textbf{0.91} & (**) & 0.87 & (**) & \textbf{0.71} & (*) \\
    \cmidrule{2-11}
    {} & \multirow{3}{*}{\begin{tabular}{@{}c@{}}TS\end{tabular}} & K$\tau$ & 0.50 & (0.1) & 0.50 & (0.1) & 0.39 & (0.2) & 0.17 & (0.6) \\
    \multirow{-5}{*}{\rotatebox[origin=c]{90}{\parbox{1cm}{\centering AdaBN}}} &  & S$\rho$ & 0.67 & (0.1) & 0.73 & (*) & 0.65 & (0.1) & 0.27 & (0.5) \\
     &  & P$r$ & \textbf{0.90} & (**) & 0.81 & (**) & \textbf{0.89} & (**) & 0.34 & (0.4) \\
    \bottomrule
    \end{tabular}
    \label{tab:finetuned_semantic}
\end{minipage}%
\hfill
\begin{minipage}[t]{0.41\textwidth}
    \centering
    \tiny
    \caption{\sloppy Mean correlation to Instance mAP over target datasets. $\dagger$ notes ours. $|\mathcal{M}_{Cells}|=8$ to~\cite{leal-taixe_benchmark_2019,wolny_accurate_2020,willis_cell_2016}), $|\mathcal{M}_{Nuclei}|=5$ to~\cite{von_chamier_democratising_2021,kromp_annotated_2020,ljosa_annotated_2012,arvidsson_annotated_2023}).\vspace{\tablecaptionvspace}}
    \setlength{\tabcolsep}{2.5pt}
    \begin{tabular}{@{}c c c 
       >{\columncolor{GreyTable}}c
       >{\columncolor{GreyTable}}c
       c c@{}}
    \toprule
    \multirow{2}{*}{Metric} & \multirow{2}{*}{} & \multirow{2}{*}{} & \multicolumn{2}{c}{\cellcolor{SecondaryColumnColor}Cells} & \multicolumn{2}{c}{Nuclei} \\
    \cmidrule(lr){4-5}
    \cmidrule(lr){6-7}
     &  &  & \cellcolor{SecondaryColumnColor}\textit{$\mu$} & \cellcolor{SecondaryColumnColor}\textit{$\sigma$} & \textit{$\mu$} & \textit{$\sigma$} \\
    \cmidrule{1-7}
    \multirow{3}{*}{\begin{tabular}{@{}c@{}} CMR-ARS $\dagger$ \\ (Gauss)\end{tabular}} & \multirow{3}{*}{} & K$\boldsymbol{\tau}$ & 0.69 & ±0.15 & 0.72 & ±0.25 \\
     &  & S$\boldsymbol{\rho}$ & 0.83 & ±0.09 & 0.83 & ±0.17 \\
     &  & P$r$ & 0.90 & ±0.04 & 0.79 & ±0.21 \\
    \cmidrule{1-7}
    \multirow{3}{*}{\begin{tabular}{@{}c@{}} CMR-ARS $\dagger$ \\ (DropOut)\end{tabular}} & \multirow{3}{*}{} & K$\boldsymbol{\tau}$ & \textbf{0.69} & ±\textbf{0.09} & \textbf{0.80} & ±\textbf{0.28} \\
     &  & S$\boldsymbol{\rho}$ & \textbf{0.83} & ±\textbf{0.06} & \textbf{0.85} & ±\textbf{0.24} \\
     &  & P$r$ & \textbf{0.91} & ±\textbf{0.04} & \textbf{0.84} & ±\textbf{0.11} \\
     \cmidrule{1-7}
     \multirow{3}{*}{SEG} & \multirow{3}{*}{} & K$\boldsymbol{\tau}$ & 0.12 & ±0.49 & 0.48 & ±0.42 \\
     &  & S$\boldsymbol{\rho}$ & 0.16 & ±0.68 & 0.53 & ±0.49 \\
     &  & P$r$ & 0.31 & ±1.02 & 0.13 & ±0.72 \\
    \bottomrule
    \end{tabular}
    \label{tab:Instance_seg_corr}
\end{minipage}
\end{table}

\subsection{Instance Segmentation}
Instance segmentation model ranking remains largely unexplored. Unlike semantic segmentation, instance outputs consist of unordered sets of masks with no fixed correspondence to feature representations, making previously compared transferability metrics inapplicable.
\cref{tab:Instance_seg_corr} reports correlation with mean Average Precision (mAP@[0.5:0.95])~\cite{lin_microsoft_2015} for our CMR-ARS method and the ensemble-based SEG method across two tasks and seven target datasets (see \textit{Supp.} for model details). SEG requires two hyperparameters: the agreement ratio $a_r$ defining ensemble consensus for pseudo ground truth and the centroid dilation radius $r$. Following~\cite{sims_seg_2023} we set $a_r=0.75$ and estimate $r$ as half the mean instance size from a small random sample, reflecting a realistic unsupervised setting. However, SEG proves highly sensitive to these parameters, as reflected by the large variance ($\sigma$) values (see \textit{Supp.}). In contrast, CMR-ARS shows consistently strong correlations with performance under both input and feature-space perturbations, outperforming SEG across all targets.
A key advantage of CMR-ARS is that it evaluates instance segmentations directly at the pixel level using the Rand score. In contrast, SEG reduces each instance to a circularly dilated centroid and measures object detections only, which is less sensitive to pixel-level quality differences and problematic for elongated or irregular biomedical objects.

\begin{figure}[tb]
    \centering
    \includegraphics[width=0.9\linewidth]{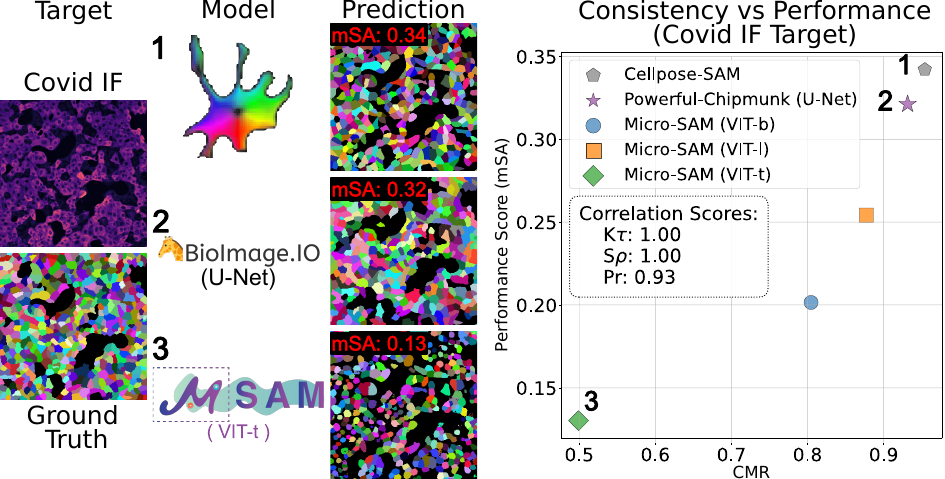}
    \caption{Instance segmentation: Correlation to mSA on Covid\_IF~\cite{pape_microscopy-based_2021} target.}
    \label{fig:Instance_seg_covid}
\end{figure}

Since CMR relies only on output consistency, it can compare models with arbitrary instantiation strategies. \cref{fig:Instance_seg_covid} illustrates a realistic ranking scenario on the Covid-IF cell segmentation dataset using several widely used models: $\mu$SAM~\cite{archit_segment_2025} with 3 transformer backbones (ViT-T, ViT-B, ViT-L), Cellpose-SAM~\cite{pachitariu_cellpose-sam_2025}, and the specialist U-Net model “Powerful-Chipmunk”~\cite{pape_microscopy-based_2021} from the BioImage Model Zoo~\cite{ouyang_bioimage_2022}. Despite substantial differences in architecture and instance generation strategies, CMR accurately reproduces the true mean Segmentation Accuracy (mSA)~\cite{everingham_pascal_2010} ranking across this diverse model set (\cref{fig:Instance_seg_covid}), demonstrating effective unsupervised model selection in a realistic repository setting.
\section{Limitations and Conclusion}

This work introduces a consistency-based model ranking (CMR) method for semantic and instance segmentation that is source-free, unsupervised, and model-agnostic. Our rankings strongly correlate with true target-domain model performance across diverse segmentation tasks and model types, both under direct application and after unsupervised domain adaptation, while requiring only two inference passes on unlabelled target data. The resulting combination of accuracy, efficiency, and conceptual simplicity makes CMR well suited for practical model selection in repository settings with heterogeneous pretrained models.

Like other transferability metrics, CMR assumes alignment between source and target tasks, which may be violated for highly generalist models without explicit task specification (e.g., $\mu\text{SAM}$ and Cellpose-SAM). In addition, the current formulation aligns most closely with pixel-wise evaluation metrics such as F1 or mAP; boundary-sensitive metrics could be incorporated through pixel-importance weighting within the consistency calculation. More broadly, we believe that unsupervised ranking methods such as CMR can enable scalable model reuse and benchmarking in emerging model repositories, supporting wider adoption of segmentation models in domains where labeled data remains scarce.

\section{Acknowledgements}
The research was carried out as part of the AI4Life consortium. AI4Life receives funding from the European Union’s Horizon Europe research and innovation programme under grant agreement number 101057970. Computational resources were provided by the High-Performance Computing (HPC) cluster at EMBL. The authors acknowledge the support of these resources, including technical assistance and computational infrastructure, which were essential for this work. Special thanks also to Fynn Beuttenmueller for many constructive conversations.


\bibliographystyle{splncs04}
\bibliography{paper}

@String(CVPR= {IEEE Conf. Comput. Vis. Pattern Recog.})

@String(ICCV= {Int. Conf. Comput. Vis.})

@String(ECCV= {Eur. Conf. Comput. Vis.})

@String(ICASSP=	{ICASSP})

@String(CVPR  = {CVPR})

@String(ICCV  = {ICCV})

@String(ECCV  = {ECCV})

@article{wang_aleatoric_2019,
	title = {Aleatoric uncertainty estimation with test-time augmentation for medical image segmentation with convolutional neural networks.},
	doi = {10.1016/j.neucom.2019.01.103},
	journal = {Neurocomputing},
	author = {Wang, Guotai and Li, Wenqi and Li, Wenqi and Aertsen, Michael and Deprest, Jan and Ourselin, Sebastien and Vercauteren, Tom},
	year = {2019},
}

@article{isensee2021nnu,
  title={{nnU-Net: a self-configuring method for deep learning-based biomedical image segmentation}},
  author={Isensee, Fabian and Jaeger, Paul F and Kohl, Simon AA and Petersen, Jens and Maier-Hein, Klaus H},
  journal={Nature methods},
  volume={18},
  number={2},
  pages={203--211},
  year={2021},
  publisher={Nature Publishing Group}
}

@misc{bailoni_gasp_2022,
	title = {{GASP}, a generalized framework for agglomerative clustering of signed graphs and its application to {Instance} {Segmentation}},
	url = {http://arxiv.org/abs/1906.11713},
	doi = {10.48550/arXiv.1906.11713},
	urldate = {2025-11-12},
	publisher = {arXiv},
	author = {Bailoni, Alberto and Pape, Constantin and Hütsch, Nathan and Wolf, Steffen and Beier, Thorsten and Kreshuk, Anna and Hamprecht, Fred A.},
	month = jun,
	year = {2022},
	note = {arXiv:1906.11713 [cs]},
}

@misc{he_deep_2015,
	title = {Deep {Residual} {Learning} for {Image} {Recognition}},
	url = {https://arxiv.org/abs/1512.03385v1},
	language = {en},
	urldate = {2025-11-13},
	journal = {arXiv.org},
	author = {He, Kaiming and Zhang, Xiangyu and Ren, Shaoqing and Sun, Jian},
	month = dec,
	year = {2015},
}

@misc{huang_densely_2018,
	title = {Densely {Connected} {Convolutional} {Networks}},
	url = {http://arxiv.org/abs/1608.06993},
	doi = {10.48550/arXiv.1608.06993},
	urldate = {2025-11-13},
	publisher = {arXiv},
	author = {Huang, Gao and Liu, Zhuang and Maaten, Laurens van der and Weinberger, Kilian Q.},
	month = jan,
	year = {2018},
	note = {arXiv:1608.06993 [cs]},
}

@misc{sandler_mobilenetv2_2018,
	title = {{MobileNetV2}: {Inverted} {Residuals} and {Linear} {Bottlenecks}},
	shorttitle = {{MobileNetV2}},
	url = {https://arxiv.org/abs/1801.04381v4},
	language = {en},
	urldate = {2025-11-13},
	journal = {arXiv.org},
	author = {Sandler, Mark and Howard, Andrew and Zhu, Menglong and Zhmoginov, Andrey and Chen, Liang-Chieh},
	month = jan,
	year = {2018},
}

@INPROCEEDINGS{Qian_MobileNetV3,
  author={Qian, Siying and Ning, Chenran and Hu, Yuepeng},
  booktitle={2021 IEEE 2nd International Conference on Big Data, Artificial Intelligence and Internet of Things Engineering (ICBAIE)}, 
  title={MobileNetV3 for Image Classification}, 
  year={2021},
  volume={},
  number={},
  pages={490-497},
  doi={10.1109/ICBAIE52039.2021.9389905}}

@article{2024TMI,
  title={{Segmenting the Inferior Alveolar Canal in CBCTs Volumes: the ToothFairy Challenge}},
  author={Bolelli, Federico and Lumetti, Luca and Vinayahalingam, Shankeeth and Di Bartolomeo, Mattia and Pellacani, Arrigo and Marchesini, Kevin and van Nistelrooij, Niels and van Lierop, Pieter and Xi, Tong and Liu, Yusheng and Xin, Rui and Yang, Tao and Wang, Lisheng and Wang, Haoshen and Xu, Chenfan and Cui, Zhiming and Wodzinski, Marek and Müller, Henning and Kirchhoff, Yannick and R. Rokuss, Maximilian and Maier-Hein, Klaus and Han, Jaehwan and Kim, Wan and Ahn, Hong-Gi and Szczepański, Tomasz and Grzeszczyk, Michal K. and Korzeniowski, Przemyslaw and Caselles Ballester, Vicent amd Paolo Burgos-Artizzu, Xavier and Prados Carrasco, Ferran and Berge’, Stefaan and van Ginneken, Bram and Anesi, Alexandre and Grana, Costantino},
  year={2024},
  month={Dec},
  journal={IEEE Transactions on Medical Imaging},
  doi={https://doi.org/10.1109/TMI.2024.3523096}
}

@inproceedings{2025CVPR,
  title={{Segmenting Maxillofacial Structures in CBCT Volumes}},
  author={Bolelli, Federico and Marchesini, Kevin and van Nistelrooij, Niels and Lumetti, Luca and Pipoli, Vittorio and Ficarra, Elisa and Vinayahalingam, Shankeeth and Grana, Costantino},
  year={2025},
  month={Mar},
  booktitle={IEEE/CVF Conference on Computer Vision and Pattern Recognition (CVPR)},
  venue={Nashville, Tennessee, USA}
}

@book{wang2025supervised,
  title={{Supervised and Semi-supervised Multi-structure Segmentation and Landmark Detection in Dental Data: MICCAI 2024 Challenges: ToothFairy 2024, 3DTeethLand 2024, and STS 2024, Held in Conjunction with MICCAI 2024, Marrakesh, Morocco, October 6, 2024, Proceedings}},
  author={Wang, Yaqi and Qian, Dahong and Wang, Shuai and Ben-Hamadou, Achraf Achraf Pujades, Sergi and   Lumetti, Luca and   Grana, Costantino and Bolelli, Federico},
  year={2025},
  publisher={Springer Nature}
}

@article{you_logme_2021,
	title = {{LogME}: {Practical} {Assessment} of {Pre}-trained {Models} for {Transfer} {Learning}},
	doi = {null},
	journal = {International Conference on Machine Learning},
	author = {You, Kaichao and Liu, Yong and Long, Mingsheng and Wang, Jianmin},
	year = {2021},
	pmid = {null},
	pmcid = {null},
}

@article{nguyen_leep_2020,
	title = {{LEEP}: {A} {New} {Measure} to {Evaluate} {Transferability} of {Learned} {Representations}},
	doi = {null},
	journal = {arXiv: Learning},
	author = {Nguyen, Cuong V. and Nguyen, Cuong V. and Hassner, Tal and Archambeau, Cédric and Seeger, Matthias},
	year = {2020},
	pmid = {null},
	pmcid = {null},
}

@article{agostinelli_how_2022,
	title = {How stable are {Transferability} {Metrics} evaluations?},
	doi = {10.48550/arxiv.2204.01403},
	journal = {European Conference on Computer Vision},
	author = {Agostinelli, A. and P'andy, Michal and Uijlings, J. and Mensink, Thomas and Ferrari, V.},
	year = {2022},
	pmid = {null},
	pmcid = {null},
}

@article{pandy_transferability_2021,
	title = {Transferability {Estimation} using {Bhattacharyya} {Class} {Separability}},
	doi = {10.1109/cvpr52688.2022.00896},
	journal = {Computer Vision and Pattern Recognition},
	author = {P'andy, Michal and Agostinelli, A. and Uijlings, J. and Ferrari, V. and Mensink, Thomas},
	year = {2021},
	pmid = {null},
	pmcid = {null},
}

@article{ouyang_bioimage_2022,
	title = {{BioImage} {Model} {Zoo}: {A} {Community}-{Driven} {Resource} for {Accessible} {Deep} {Learning} in {BioImage} {Analysis}},
	doi = {10.1101/2022.06.07.495102},
	journal = {bioRxiv},
	author = {Ouyang, Wei and Beuttenmueller, Fynn and Gómez-de-Mariscal, Estibaliz and Pape, Constantin and Burke, Tom and Garcia-López-de-Haro, Carlos and Russell, Craig and Moya-Sans, Lucía and de-la-Torre-Gutiérrez, Cristina and Schmidt, Deborah and Kutra, Dominik and Novikov, Maksim and Weigert, Martin and Schmidt, Uwe and Bankhead, Peter and Jacquemet, Guillaume and Sage, Daniel and Henriques, Ricardo and Muñoz-Barrutia, Arrate and Lundberg, Emma and Jug, Florian and Kreshuk, Anna},
	year = {2022},
	pmid = {null},
	pmcid = {null},
}

@article{stringer_cellpose_2020,
	title = {Cellpose: a generalist algorithm for cellular segmentation},
	doi = {10.1101/2020.02.02.931238},
	journal = {bioRxiv},
	author = {Stringer, Carsen and Wang, Timothy C. and Michaelos, Michalis and Pachitariu, Marius},
	year = {2020},
	pmid = {33318659},
	pmcid = {null},
}

@article{ronneberger_u-net_2015,
	title = {U-{Net}: {Convolutional} {Networks} for {Biomedical} {Image} {Segmentation}},
	doi = {10.1007/978-3-319-24574-4_28},
	journal = {arXiv: Computer Vision and Pattern Recognition},
	author = {Ronneberger, Olaf and Fischer, Philipp and Brox, Thomas},
	year = {2015},
	pmid = {null},
	pmcid = {null},
}

@article{weiss_survey_2016,
	title = {A survey of transfer learning},
	doi = {10.1186/s40537-016-0043-6},
	journal = {Journal of Big Data},
	author = {Weiss, Karl R. and Khoshgoftaar, Taghi M. and Wang, Dingding},
	year = {2016},
	pmid = {null},
	pmcid = {null},
}

@article{renggli_which_2020,
	title = {Which {Model} to {Transfer}? {Finding} the {Needle} in the {Growing} {Haystack}},
	doi = {10.1109/cvpr52688.2022.00899},
	journal = {arXiv: Learning},
	author = {Renggli, Cedric and Pinto, André Susano and Rimanic, Luka and Puigcerver, Joan and Ruiz, Carlos Riquelme and Riquelme, Carlos and Zhang, Ce and Lucic, Mario},
	year = {2020},
	pmid = {null},
	pmcid = {null},
}

@article{mustafa_supervised_2021,
	title = {Supervised {Transfer} {Learning} at {Scale} for {Medical} {Imaging}},
	doi = {null},
	journal = {arXiv.org},
	author = {Mustafa, Basil and Loh, Aaron and Freyberg, J. and MacWilliams, Patricia and Wilson, Megan and McKinney, S. M. and Sieniek, M. and Winkens, Jim and Liu, Yuan and Bui, P. and Prabhakara, Shruthi and Telang, Umesh and Karthikesalingam, A. and Houlsby, N. and Natarajan, Vivek},
	year = {2021},
	pmid = {null},
	pmcid = {null},
}

@misc{chaves_performance_2023,
	title = {The {Performance} of {Transferability} {Metrics} does not {Translate} to {Medical} {Tasks}},
	url = {http://arxiv.org/abs/2308.07444},
	doi = {10.48550/arXiv.2308.07444},
	urldate = {2024-01-23},
	publisher = {arXiv},
	author = {Chaves, Levy and Bissoto, Alceu and Valle, Eduardo and Avila, Sandra},
	month = aug,
	year = {2023},
	note = {arXiv:2308.07444 [cs]},
}

@misc{saito_tune_2021,
	title = {Tune it the {Right} {Way}: {Unsupervised} {Validation} of {Domain} {Adaptation} via {Soft} {Neighborhood} {Density}},
	shorttitle = {Tune it the {Right} {Way}},
	url = {http://arxiv.org/abs/2108.10860},
	doi = {10.48550/arXiv.2108.10860},
	urldate = {2024-01-25},
	publisher = {arXiv},
	author = {Saito, Kuniaki and Kim, Donghyun and Teterwak, Piotr and Sclaroff, Stan and Darrell, Trevor and Saenko, Kate},
	month = aug,
	year = {2021},
	note = {arXiv:2108.10860 [cs]},

}

@misc{garg_leveraging_2022,
	title = {Leveraging {Unlabeled} {Data} to {Predict} {Out}-of-{Distribution} {Performance}},
	url = {http://arxiv.org/abs/2201.04234},
	doi = {10.48550/arXiv.2201.04234},
	urldate = {2024-05-07},
	author = {Garg, Saurabh and Balakrishnan, Sivaraman and Lipton, Zachary C. and Neyshabur, Behnam and Sedghi, Hanie},
	month = oct,
	year = {2022},
	note = {arXiv:2201.04234 [cs, stat]},

}

@inproceedings{jiang_predicting_2018,
	title = {Predicting the {Generalization} {Gap} in {Deep} {Networks} with {Margin} {Distributions}},
	url = {https://openreview.net/forum?id=HJlQfnCqKX},
	urldate = {2024-05-07},
	author = {Jiang, Yiding and Krishnan, Dilip and Mobahi, Hossein and Bengio, Samy},
	month = sep,
	year = {2018},
}

@misc{deng_strong_2022,
	title = {On the {Strong} {Correlation} {Between} {Model} {Invariance} and {Generalization}},
	url = {http://arxiv.org/abs/2207.07065},
	doi = {10.48550/arXiv.2207.07065},
	urldate = {2024-06-16},
	author = {Deng, Weijian and Gould, Stephen and Zheng, Liang},
	month = jul,
	year = {2022},
	note = {arXiv:2207.07065 [cs]},
}

@incollection{ibrahim_newer_2023,
	title = {Newer is not always better: {Rethinking} transferability metrics, their peculiarities, stability and performance},
	shorttitle = {Newer is not always better},
	url = {http://arxiv.org/abs/2110.06893},
	urldate = {2024-09-17},
	author = {Ibrahim, Shibal and Ponomareva, Natalia and Mazumder, Rahul},
	year = {2023},
	doi = {10.1007/978-3-031-26387-3_42},
	note = {ISSN: 0302-9743, 1611-3349
arXiv:2110.06893 [cs]},
	pages = {693--709},
}

@article{li_adaptive_2018,
	title = {Adaptive {Batch} {Normalization} for practical domain adaptation},
	doi = {10.1016/j.patcog.2018.03.005},
	journal = {Pattern Recognition},
	author = {Li, Yanghao and Wang, Naiyan and Shi, Jianping and Hou, Xiaodi and Liu, Jiaying},
	year = {2018},
	pmid = {null},
	pmcid = {null},
}

@misc{tarvainen_mean_2018,
	title = {Mean teachers are better role models: {Weight}-averaged consistency targets improve semi-supervised deep learning results},
	shorttitle = {Mean teachers are better role models},
	url = {http://arxiv.org/abs/1703.01780},
	doi = {10.48550/arXiv.1703.01780},
	urldate = {2024-09-25},
	author = {Tarvainen, Antti and Valpola, Harri},
	month = apr,
	year = {2018},
	note = {arXiv:1703.01780 [cs, stat]},
}

@misc{ouali_semi-supervised_2020,
	title = {Semi-{Supervised} {Semantic} {Segmentation} with {Cross}-{Consistency} {Training}},
	url = {http://arxiv.org/abs/2003.09005},
	doi = {10.48550/arXiv.2003.09005},
	urldate = {2025-01-15},
	publisher = {arXiv},
	author = {Ouali, Yassine and Hudelot, Céline and Tami, Myriam},
	month = jun,
	year = {2020},
	note = {arXiv:2003.09005 [cs]},

}

@article{arganda-carreras_crowdsourcing_2015,
	title = {Crowdsourcing the creation of image segmentation algorithms for connectomics},
	volume = {9},
	issn = {1662-5129},
	url = {https://www.ncbi.nlm.nih.gov/pmc/articles/PMC4633678/},
	doi = {10.3389/fnana.2015.00142},
	urldate = {2025-01-23},
	journal = {Frontiers in Neuroanatomy},
	author = {Arganda-Carreras, Ignacio and Turaga, Srinivas C. and Berger, Daniel R. and Cireşan, Dan and Giusti, Alessandro and Gambardella, Luca M. and Schmidhuber, Jürgen and Laptev, Dmitry and Dwivedi, Sarvesh and Buhmann, Joachim M. and Liu, Ting and Seyedhosseini, Mojtaba and Tasdizen, Tolga and Kamentsky, Lee and Burget, Radim and Uher, Vaclav and Tan, Xiao and Sun, Changming and Pham, Tuan D. and Bas, Erhan and Uzunbas, Mustafa G. and Cardona, Albert and Schindelin, Johannes and Seung, H. Sebastian},
	month = nov,
	year = {2015},
	pmid = {26594156},
	pmcid = {PMC4633678},
	pages = {142},
}

@article{spearman_proof_1904,
	title = {The {Proof} and {Measurement} of {Association} between {Two} {Things}},
	volume = {15},
	url = {https://doi.org/10.2307/1412159},
	doi = {10.2307/1412159},
	journal = {Am. J. Psychol.},
	author = {Spearman, C.},
	year = {1904},
}

@inproceedings{1972RankCM,
  title={Rank Correlation Methods}, booktitle={4th edition (1970)},
  author={M. G. Kendall},
  year={1972},
  url={https://api.semanticscholar.org/CorpusID:160270288}
}

@inproceedings{unnikrishnan_measures_2005,
	title = {Measures of {Similarity}},
	volume = {1},
	url = {https://ieeexplore.ieee.org/document/4129508},
	doi = {10.1109/ACVMOT.2005.71},
	urldate = {2025-11-12},
	booktitle = {2005 {Seventh} {IEEE} {Workshops} on {Applications} of {Computer} {Vision} ({WACV}/{MOTION}'05) - {Volume} 1},
	author = {Unnikrishnan, Ranjith and Hebert, Martial},
	month = jan,
	year = {2005},
	pages = {394--394},
}

@article{li_ranking_2020,
	title = {Ranking {Neural} {Checkpoints}},
	doi = {10.1109/cvpr46437.2021.00269},
	journal = {arXiv: Learning},
	author = {Li, Yandong and Jia, Xuhui and Sang, Ruoxin and Zhu, Yukun and Green, Bradley and Green, Bradley Ray and Wang, Liqiang and Gong, Boqing and Gong, Boqing},
	year = {2020},
	pmid = {null},
	pmcid = {null},
}

@misc{k_robustness_2021,
	title = {Robustness to {Augmentations} as a {Generalization} metric},
	ignore_publisher = {arXiv},
	author = {Aithal, Sumukh K  and Kashyap, Dhruva and Subramanyam, Natarajan},
	ignore_month = jan,
	year = {2021},
	note = {arXiv:2101.06459},
}

@misc{schiff_predicting_2021,
	title = {Predicting {Deep} {Neural} {Network} {Generalization} with {Perturbation} {Response} {Curves}},
	ignore_publisher = {arXiv},
	author = {Schiff, Yair and Quanz, Brian and Das, Payel and Chen, Pin-Yu},
	ignore_month = oct,
	year = {2021},
	note = {arXiv:2106.04765},
}

@article{loquercio_general_2020,
	title = {A {General} {Framework} for {Uncertainty} {Estimation} in {Deep} {Learning}},
	volume = {5},
	issn = {2377-3766, 2377-3774},
	url = {http://arxiv.org/abs/1907.06890},
	doi = {10.1109/LRA.2020.2974682},
	number = {2},
	urldate = {2025-01-27},
	journal = {IEEE Robotics and Automation Letters},
	author = {Loquercio, Antonio and Segù, Mattia and Scaramuzza, Davide},
	month = apr,
	year = {2020},
	note = {arXiv:1907.06890 [cs]},
	pages = {3153--3160},

}

@article{ledda_dropout_2023,
	title = {Dropout injection at test time for post hoc uncertainty quantification in neural networks},
	volume = {645},
	issn = {0020-0255},
	url = {https://www.sciencedirect.com/science/article/pii/S0020025523009416},
	doi = {10.1016/j.ins.2023.119356},
	urldate = {2025-01-27},
	journal = {Information Sciences},
	author = {Ledda, Emanuele and Fumera, Giorgio and Roli, Fabio},
	month = oct,
	year = {2023},
	pages = {119356},

}

@inproceedings{wang_epistemic_2024,
	title = {Epistemic {Uncertainty} {Quantification} {For} {Pre}-{Trained} {Neural} {Networks}},
	url = {https://openaccess.thecvf.com/content/CVPR2024/html/Wang_Epistemic_Uncertainty_Quantification_For_Pre-Trained_Neural_Networks_CVPR_2024_paper.html},
	urldate = {2025-01-26},
	author = {Wang, Hanjing and Ji, Qiang},
	year = {2024},
	pages = {11052--11061},

}

@inproceedings{gal_dropout_2016,
	title = {Dropout as a {Bayesian} {Approximation}: {Representing} {Model} {Uncertainty} in {Deep} {Learning}},
	shorttitle = {Dropout as a {Bayesian} {Approximation}},
	url = {https://proceedings.mlr.press/v48/gal16.html},
	language = {en},
	urldate = {2025-01-28},
	booktitle = {Proceedings of {The} 33rd {International} {Conference} on {Machine} {Learning}},
	publisher = {PMLR},
	author = {Gal, Yarin and Ghahramani, Zoubin},
	month = jun,
	year = {2016},
	note = {ISSN: 1938-7228},
	pages = {1050--1059},

}

@inproceedings{ovadia_can_2019,
	title = {Can you trust your model' s uncertainty? {Evaluating} predictive uncertainty under dataset shift},
	volume = {32},
	shorttitle = {Can you trust your model' s uncertainty?},
	url = {https://proceedings.neurips.cc/paper_files/paper/2019/hash/8558cb408c1d76621371888657d2eb1d-Abstract.html},
	urldate = {2025-01-29},
	booktitle = {Advances in {Neural} {Information} {Processing} {Systems}},
	publisher = {Curran Associates, Inc.},
	author = {Ovadia, Yaniv and Fertig, Emily and Ren, Jie and Nado, Zachary and Sculley, D. and Nowozin, Sebastian and Dillon, Joshua and Lakshminarayanan, Balaji and Snoek, Jasper},
	year = {2019},

}

@misc{mi_training-free_2022,
	title = {Training-{Free} {Uncertainty} {Estimation} for {Dense} {Regression}: {Sensitivity} as a {Surrogate}},
	shorttitle = {Training-{Free} {Uncertainty} {Estimation} for {Dense} {Regression}},
	url = {http://arxiv.org/abs/1910.04858},
	doi = {10.48550/arXiv.1910.04858},
	urldate = {2025-01-29},
	publisher = {arXiv},
	author = {Mi, Lu and Wang, Hao and Tian, Yonglong and He, Hao and Shavit, Nir},
	month = jan,
	year = {2022},
	note = {arXiv:1910.04858 [cs]},
}

@misc{ding_which_2024,
	title = {Which {Model} to {Transfer}? {A} {Survey} on {Transferability} {Estimation}},
	shorttitle = {Which {Model} to {Transfer}?},
	url = {http://arxiv.org/abs/2402.15231},
	doi = {10.48550/arXiv.2402.15231},
	urldate = {2025-01-29},
	publisher = {arXiv},
	author = {Ding, Yuhe and Jiang, Bo and Yu, Aijing and Zheng, Aihua and Liang, Jian},
	month = feb,
	year = {2024},
	note = {arXiv:2402.15231 [cs]},
}

@misc{yang_pick_2023,
	title = {Pick the {Best} {Pre}-trained {Model}: {Towards} {Transferability} {Estimation} for {Medical} {Image} {Segmentation}},
	shorttitle = {Pick the {Best} {Pre}-trained {Model}},
	url = {http://arxiv.org/abs/2307.11958},
	doi = {10.48550/arXiv.2307.11958},
	urldate = {2025-01-29},
	publisher = {arXiv},
	author = {Yang, Yuncheng and Wei, Meng and He, Junjun and Yang, Jie and Ye, Jin and Gu, Yun},
	month = jul,
	year = {2023},
	note = {arXiv:2307.11958 [cs]},
}

@misc{bao_information-theoretic_2022,
	title = {An {Information}-{Theoretic} {Approach} to {Transferability} in {Task} {Transfer} {Learning}},
	url = {http://arxiv.org/abs/2212.10082},
	doi = {10.48550/arXiv.2212.10082},
	urldate = {2025-01-29},
	publisher = {arXiv},
	author = {Bao, Yajie and Li, Yang and Huang, Shao-Lun and Zhang, Lin and Zheng, Lizhong and Zamir, Amir and Guibas, Leonidas},
	month = dec,
	year = {2022},
	note = {arXiv:2212.10082 [cs]},
}

@misc{yang_can_2024,
	title = {Can {We} {Evaluate} {Domain} {Adaptation} {Models} {Without} {Target}-{Domain} {Labels}?},
	url = {http://arxiv.org/abs/2305.18712},
	doi = {10.48550/arXiv.2305.18712},
	urldate = {2025-01-30},
	publisher = {arXiv},
	author = {Yang, Jianfei and Qian, Hanjie and Xu, Yuecong and Wang, Kai and Xie, Lihua},
	month = feb,
	year = {2024},
	note = {arXiv:2305.18712 [cs]},
}

@misc{morerio_minimal-entropy_2017,
	title = {Minimal-{Entropy} {Correlation} {Alignment} for {Unsupervised} {Deep} {Domain} {Adaptation}},
	url = {http://arxiv.org/abs/1711.10288},
	doi = {10.48550/arXiv.1711.10288},
	urldate = {2025-01-30},
	publisher = {arXiv},
	author = {Morerio, Pietro and Cavazza, Jacopo and Murino, Vittorio},
	month = nov,
	year = {2017},
	note = {arXiv:1711.10288 [cs]},
}

@inproceedings{he_reshaping_2014,
	address = {Florence, Italy},
	title = {Reshaping deep neural network for fast decoding by node-pruning},
	isbn = {978-1-4799-2893-4},
	url = {http://ieeexplore.ieee.org/document/6853595/},
	doi = {10.1109/ICASSP.2014.6853595},
	language = {en},
	urldate = {2025-02-24},
	booktitle = {2014 {IEEE} {International} {Conference} on {Acoustics}, {Speech} and {Signal} {Processing} ({ICASSP})},
	publisher = {IEEE},
	author = {He, Tianxing and Fan, Yuchen and Qian, Yanmin and Tan, Tian and Yu, Kai},
	month = may,
	year = {2014},
	pages = {245--249},
}

@misc{tompson_efficient_2015,
	title = {Efficient {Object} {Localization} {Using} {Convolutional} {Networks}},
	url = {http://arxiv.org/abs/1411.4280},
	doi = {10.48550/arXiv.1411.4280},
	urldate = {2025-02-12},
	publisher = {arXiv},
	author = {Tompson, Jonathan and Goroshin, Ross and Jain, Arjun and LeCun, Yann and Bregler, Christopher},
	month = jun,
	year = {2015},
	note = {arXiv:1411.4280 [cs]},
}

@inproceedings{lucchi_learning_2013,
	title = {Learning for {Structured} {Prediction} {Using} {Approximate} {Subgradient} {Descent} with {Working} {Sets}},
	url = {https://ieeexplore.ieee.org/stampPDF/getPDF.jsp?tp=&arnumber=6619103&ref=},
	doi = {10.1109/CVPR.2013.259},
	urldate = {2025-02-13},
	booktitle = {2013 {IEEE} {Conference} on {Computer} {Vision} and {Pattern} {Recognition}},
	author = {Lucchi, Aurélien and Li, Yunpeng and Fua, Pascal},
	month = jun,
	year = {2013},
	note = {ISSN: 1063-6919},
	pages = {1987--1994},
}

@article{franco-barranco_current_2023,
	title = {Current {Progress} and {Challenges} in {Large}-{Scale} {3D} {Mitochondria} {Instance} {Segmentation}},
	volume = {42},
	issn = {1558-254X},
	url = {https://ieeexplore.ieee.org/ielx7/42/10336247/10266382.pdf?tp=&arnumber=10266382&isnumber=10336247&ref=aHR0cHM6Ly9pZWVleHBsb3JlLmllZWUub3JnL2RvY3VtZW50LzEwMjY2Mzgy},
	doi = {10.1109/TMI.2023.3320497},
	number = {12},
	urldate = {2025-02-13},
	journal = {IEEE Transactions on Medical Imaging},
	author = {Franco-Barranco, Daniel and Lin, Zudi and Jang, Won-Dong and Wang, Xueying and Shen, Qijia and Yin, Wenjie and Fan, Yutian and Li, Mingxing and Chen, Chang and Xiong, Zhiwei and Xin, Rui and Liu, Hao and Chen, Huai and Li, Zhili and Zhao, Jie and Chen, Xuejin and Pape, Constantin and Conrad, Ryan and Nightingale, Luke and de Folter, Joost and Jones, Martin L. and Liu, Yanling and Ziaei, Dorsa and Huschauer, Stephan and Arganda-Carreras, Ignacio and Pfister, Hanspeter and Wei, Donglai},
	month = dec,
	year = {2023},
	note = {Conference Name: IEEE Transactions on Medical Imaging},
	pages = {3956--3971},
}

@article{phelps_reconstruction_2021,
	title = {Reconstruction of motor control circuits in adult {Drosophila} using automated transmission electron microscopy},
	volume = {184},
	issn = {0092-8674},
	url = {https://www.ncbi.nlm.nih.gov/pmc/articles/PMC8312698/},
	doi = {10.1016/j.cell.2020.12.013},
	number = {3},
	urldate = {2025-02-13},
	journal = {Cell},
	author = {Phelps, Jasper S. and Colburn Hildebrand, David Grant and Graham, Brett J. and Kuan, Aaron T. and Thomas, Logan A. and Nguyen, Tri M. and Buhmann, Julia and Azevedo, Anthony W. and Sustar, Anne and Agrawal, Sweta and Liu, Mingguan and Shanny, Brendan L. and Funke, Jan and Tuthill, John C. and Allen Lee, Wei-Chung},
	month = feb,
	year = {2021},
	pmid = {33400916},
	pmcid = {PMC8312698},
	pages = {759--774.e18},
}

@article{vijayan_deep_2024,
	title = {A deep learning-based toolkit for {3D} nuclei segmentation and quantitative analysis in cellular and tissue context},
	volume = {151},
	issn = {0950-1991},
	url = {https://www.ncbi.nlm.nih.gov/pmc/articles/PMC11273294/},
	doi = {10.1242/dev.202800},
	number = {14},
	urldate = {2025-02-13},
	journal = {Development (Cambridge, England)},
	author = {Vijayan, Athul and Mody, Tejasvinee Atul and Yu, Qin and Wolny, Adrian and Cerrone, Lorenzo and Strauss, Soeren and Tsiantis, Miltos and Smith, Richard S. and Hamprecht, Fred A. and Kreshuk, Anna and Schneitz, Kay},
	month = jul,
	year = {2024},
	pmid = {39036998},
	pmcid = {PMC11273294},
	pages = {dev202800},
}

@article{von_chamier_democratising_2021,
	title = {Democratising deep learning for microscopy with {ZeroCostDL4Mic}},
	volume = {12},
	copyright = {2021 The Author(s)},
	issn = {2041-1723},
	url = {https://www.nature.com/articles/s41467-021-22518-0},
	doi = {10.1038/s41467-021-22518-0},
	language = {en},
	number = {1},
	urldate = {2025-02-13},
	journal = {Nature Communications},
	author = {von Chamier, Lucas and Laine, Romain F. and Jukkala, Johanna and Spahn, Christoph and Krentzel, Daniel and Nehme, Elias and Lerche, Martina and Hernández-Pérez, Sara and Mattila, Pieta K. and Karinou, Eleni and Holden, Séamus and Solak, Ahmet Can and Krull, Alexander and Buchholz, Tim-Oliver and Jones, Martin L. and Royer, Loïc A. and Leterrier, Christophe and Shechtman, Yoav and Jug, Florian and Heilemann, Mike and Jacquemet, Guillaume and Henriques, Ricardo},
	month = apr,
	year = {2021},
	note = {Publisher: Nature Publishing Group},
	pages = {2276},
}

@article{kromp_annotated_2020,
	title = {An annotated fluorescence image dataset for training nuclear segmentation methods},
	volume = {7},
	copyright = {2020 The Author(s)},
	issn = {2052-4463},
	url = {https://www.nature.com/articles/s41597-020-00608-w},
	doi = {10.1038/s41597-020-00608-w},
	language = {en},
	number = {1},
	urldate = {2025-02-13},
	journal = {Scientific Data},
	author = {Kromp, Florian and Bozsaky, Eva and Rifatbegovic, Fikret and Fischer, Lukas and Ambros, Magdalena and Berneder, Maria and Weiss, Tamara and Lazic, Daria and Dörr, Wolfgang and Hanbury, Allan and Beiske, Klaus and Ambros, Peter F. and Ambros, Inge M. and Taschner-Mandl, Sabine},
	month = aug,
	year = {2020},
	note = {Publisher: Nature Publishing Group},
	pages = {262},
}

@article{ljosa_annotated_2012,
	title = {Annotated high-throughput microscopy image sets for validation},
	volume = {9},
	copyright = {2012 Springer Nature America, Inc.},
	issn = {1548-7105},
	url = {https://www.nature.com/articles/nmeth.2083},
	doi = {10.1038/nmeth.2083},
	language = {en},
	number = {7},
	urldate = {2025-02-13},
	journal = {Nature Methods},
	author = {Ljosa, Vebjorn and Sokolnicki, Katherine L. and Carpenter, Anne E.},
	month = jul,
	year = {2012},
	note = {Publisher: Nature Publishing Group},
	pages = {637--637},
}

@article{caicedo_nucleus_2019,
	title = {Nucleus segmentation across imaging experiments: the 2018 {Data} {Science} {Bowl}},
	volume = {16},
	copyright = {2019 The Author(s)},
	issn = {1548-7105},
	shorttitle = {Nucleus segmentation across imaging experiments},
	url = {https://www.nature.com/articles/s41592-019-0612-7},
	doi = {10.1038/s41592-019-0612-7},
	language = {en},
	number = {12},
	urldate = {2025-02-13},
	journal = {Nature Methods},
	author = {Caicedo, Juan C. and Goodman, Allen and Karhohs, Kyle W. and Cimini, Beth A. and Ackerman, Jeanelle and Haghighi, Marzieh and Heng, CherKeng and Becker, Tim and Doan, Minh and McQuin, Claire and Rohban, Mohammad and Singh, Shantanu and Carpenter, Anne E.},
	month = dec,
	year = {2019},
	note = {Publisher: Nature Publishing Group},
	pages = {1247--1253},
}

@article{arvidsson_annotated_2023,
	title = {An annotated high-content fluorescence microscopy dataset with {Hoechst} 33342-stained nuclei and manually labelled outlines},
	volume = {46},
	issn = {2352-3409},
	doi = {10.1016/j.dib.2022.108769},
	language = {eng},
	journal = {Data in Brief},
	author = {Arvidsson, Malou and Rashed, Salma Kazemi and Aits, Sonja},
	month = feb,
	year = {2023},
	pmid = {36506804},
	pmcid = {PMC9727632},
	pages = {108769},
}

@misc{de_helacytonuc_2024,
	title = {{HeLaCytoNuc}: fluorescence microscopy dataset with segmentation masks for cell nuclei and cytoplasm},
	shorttitle = {{HeLaCytoNuc}},
	url = {https://rodare.hzdr.de/record/3001/export/hx},
	doi = {10.14278/rodare.3001},
	language = {eng},
	urldate = {2025-02-13},
	publisher = {Rodare},
	author = {De, Trina and Urbanski, Adrian and Thangamani, Subasini and Wyrzykowska, Maria and Yakimovich, Artur},
	month = jun,
	year = {2024},
}

@article{wolny_accurate_2020,
	title = {Accurate and versatile {3D} segmentation of plant tissues at cellular resolution},
	volume = {9},
	issn = {2050-084X},
	url = {https://doi.org/10.7554/eLife.57613},
	doi = {10.7554/eLife.57613},
	urldate = {2025-02-13},
	journal = {eLife},
	author = {Wolny, Adrian and Cerrone, Lorenzo and Vijayan, Athul and Tofanelli, Rachele and Barro, Amaya Vilches and Louveaux, Marion and Wenzl, Christian and Strauss, Sören and Wilson-Sánchez, David and Lymbouridou, Rena and Steigleder, Susanne S and Pape, Constantin and Bailoni, Alberto and Duran-Nebreda, Salva and Bassel, George W and Lohmann, Jan U and Tsiantis, Miltos and Hamprecht, Fred A and Schneitz, Kay and Maizel, Alexis and Kreshuk, Anna},
	editor = {Hardtke, Christian S and Bergmann, Dominique C and Bergmann, Dominique C and Graeff, Moritz},
	month = jul,
	year = {2020},
	note = {Publisher: eLife Sciences Publications, Ltd},
	pages = {e57613},
}

@article{willis_cell_2016,
	title = {Cell size and growth regulation in the {Arabidopsis} thaliana apical stem cell niche},
	volume = {113},
	issn = {1091-6490},
	doi = {10.1073/pnas.1616768113},
	language = {eng},
	number = {51},
	journal = {Proceedings of the National Academy of Sciences of the United States of America},
	author = {Willis, Lisa and Refahi, Yassin and Wightman, Raymond and Landrein, Benoit and Teles, José and Huang, Kerwyn Casey and Meyerowitz, Elliot M. and Jönsson, Henrik},
	month = dec,
	year = {2016},
	pmid = {27930326},
	pmcid = {PMC5187701},
	pages = {E8238--E8246},
}

@misc{hawkins_rescu-nets_2024,
	title = {{ReSCU}-{Nets}: recurrent {U}-{Nets} for segmentation of multidimensional microscopy data},
	copyright = {© 2024, Posted by Cold Spring Harbor Laboratory. This pre-print is available under a Creative Commons License (Attribution-NonCommercial 4.0 International), CC BY-NC 4.0, as described at http://creativecommons.org/licenses/by-nc/4.0/},
	shorttitle = {{ReSCU}-{Nets}},
	url = {https://www.biorxiv.org/content/10.1101/2024.11.28.625889v1},
	doi = {10.1101/2024.11.28.625889},
	language = {en},
	urldate = {2025-02-13},
	publisher = {bioRxiv},
	author = {Hawkins, Raymond and Balaghi, Negar and Rothenberg, Katheryn E. and Ly, Michelle and Fernandez-Gonzalez, Rodrigo},
	month = dec,
	year = {2024},
}

@incollection{benesty_pearson_2009,
	address = {Berlin, Heidelberg},
	title = {Pearson {Correlation} {Coefficient}},
	isbn = {978-3-642-00296-0},
	url = {https://doi.org/10.1007/978-3-642-00296-0_5},
	language = {en},
	urldate = {2025-02-14},
	booktitle = {Noise {Reduction} in {Speech} {Processing}},
	publisher = {Springer},
	author = {Benesty, Jacob and Chen, Jingdong and Huang, Yiteng and Cohen, Israel},
	editor = {Cohen, Israel and Huang, Yiteng and Chen, Jingdong and Benesty, Jacob},
	year = {2009},
	doi = {10.1007/978-3-642-00296-0_5},
	pages = {1--4},
}

@article{rand_objective_1971,
	title = {Objective {Criteria} for the {Evaluation} of {Clustering} {Methods}},
	volume = {66},
	issn = {0162-1459},
	url = {https://www.tandfonline.com/doi/abs/10.1080/01621459.1971.10482356},
	doi = {10.1080/01621459.1971.10482356},
	number = {336},
	urldate = {2025-02-24},
	journal = {Journal of the American Statistical Association},
	author = {Rand, William M.},
	month = dec,
	year = {1971},
	pages = {846--850},
}

@incollection{leal-taixe_benchmark_2019,
	address = {Cham},
	title = {A {Benchmark} for {Epithelial} {Cell} {Tracking}},
	volume = {11134},
	isbn = {978-3-030-11023-9 978-3-030-11024-6},
	url = {https://link.springer.com/10.1007/978-3-030-11024-6_33},
	language = {en},
	urldate = {2025-02-24},
	booktitle = {Computer {Vision} – {ECCV} 2018 {Workshops}},
	publisher = {Springer International Publishing},
	author = {Funke, Jan and Mais, Lisa and Champion, Andrew and Dye, Natalie and Kainmueller, Dagmar},
	editor = {Leal-Taixé, Laura and Roth, Stefan},
	year = {2019},
	doi = {10.1007/978-3-030-11024-6_33},
	note = {Series Title: Lecture Notes in Computer Science},
	pages = {437--445},
}

@inproceedings{wang_how_2023,
	address = {Paris, France},
	title = {How {Far} {Pre}-trained {Models} {Are} from {Neural} {Collapse} on the {Target} {Dataset} {Informs} their {Transferability}},
	copyright = {https://doi.org/10.15223/policy-029},
	url = {https://ieeexplore.ieee.org/document/10377311/},
	doi = {10.1109/iccv51070.2023.00511},
	language = {en},
	urldate = {2025-07-17},
	booktitle = {2023 {IEEE}/{CVF} {International} {Conference} on {Computer} {Vision} ({ICCV})},
	publisher = {IEEE},
	author = {Wang, Zijian and Luo, Yadan and Zheng, Liang and Huang, Zi and Baktashmotlagh, Mahsa},
	month = oct,
	year = {2023},
	pages = {5526--5535},
}

@inproceedings{hatamizadeh_unetr_2022,
	title = {{UNETR}: {Transformers} for {3D} {Medical} {Image} {Segmentation}},
	shorttitle = {{UNETR}},
	url = {https://ieeexplore.ieee.org/stampPDF/getPDF.jsp},
	doi = {10.1109/WACV51458.2022.00181},
	urldate = {2025-07-23},
	booktitle = {2022 {IEEE}/{CVF} {Winter} {Conference} on {Applications} of {Computer} {Vision} ({WACV})},
	author = {Hatamizadeh, Ali and Tang, Yucheng and Nath, Vishwesh and Yang, Dong and Myronenko, Andriy and Landman, Bennett and Roth, Holger R. and Xu, Daguang},
	month = jan,
	year = {2022},
	note = {ISSN: 2642-9381},
	pages = {1748--1758},
}

@misc{lee_superhuman_2017,
	title = {Superhuman {Accuracy} on the {SNEMI3D} {Connectomics} {Challenge}},
	url = {http://arxiv.org/abs/1706.00120},
	doi = {10.48550/arXiv.1706.00120},
	urldate = {2025-07-25},
	publisher = {arXiv},
	author = {Lee, Kisuk and Zung, Jonathan and Li, Peter and Jain, Viren and Seung, H. Sebastian},
	month = may,
	year = {2017},
	note = {arXiv:1706.00120 [cs]},
}

@misc{pachitariu_cellpose-sam_2025,
	title = {Cellpose-{SAM}: superhuman generalization for cellular segmentation},
	copyright = {© 2025, Posted by Cold Spring Harbor Laboratory. This pre-print is available under a Creative Commons License (Attribution-NonCommercial 4.0 International), CC BY-NC 4.0, as described at http://creativecommons.org/licenses/by-nc/4.0/},
	shorttitle = {Cellpose-{SAM}},
	url = {https://www.biorxiv.org/content/10.1101/2025.04.28.651001v1},
	doi = {10.1101/2025.04.28.651001},
	language = {en},
	urldate = {2025-08-02},
	author = {Pachitariu, Marius and Rariden, Michael and Stringer, Carsen},
	month = may,
	year = {2025},
}

@misc{cao_multi-modal_2023,
	title = {Multi-{Modal} {Continual} {Test}-{Time} {Adaptation} for {3D} {Semantic} {Segmentation}},
	url = {http://arxiv.org/abs/2303.10457},
	doi = {10.48550/arXiv.2303.10457},
	urldate = {2025-09-11},
	publisher = {arXiv},
	author = {Cao, Haozhi and Xu, Yuecong and Yang, Jianfei and Yin, Pengyu and Yuan, Shenghai and Xie, Lihua},
	month = mar,
	year = {2023},
	note = {arXiv:2303.10457 [cs]},
}

@article{archit_segment_2025,
	title = {Segment {Anything} for {Microscopy}},
	volume = {22},
	copyright = {2025 The Author(s)},
	issn = {1548-7105},
	url = {https://www.nature.com/articles/s41592-024-02580-4},
	doi = {10.1038/s41592-024-02580-4},
	language = {en},
	number = {3},
	urldate = {2025-09-26},
	journal = {Nature Methods},
	author = {Archit, Anwai and Freckmann, Luca and Nair, Sushmita and Khalid, Nabeel and Hilt, Paul and Rajashekar, Vikas and Freitag, Marei and Teuber, Carolin and Spitzner, Melanie and Tapia Contreras, Constanza and Buckley, Genevieve and von Haaren, Sebastian and Gupta, Sagnik and Grade, Marian and Wirth, Matthias and Schneider, Günter and Dengel, Andreas and Ahmed, Sheraz and Pape, Constantin},
	month = mar,
	year = {2025},
	note = {Publisher: Nature Publishing Group},
	pages = {579--591},
}

@article{wang_deep_2018,
	title = {Deep {Visual} {Domain} {Adaptation}: {A} {Survey}},
	volume = {312},
	issn = {09252312},
	shorttitle = {Deep {Visual} {Domain} {Adaptation}},
	url = {http://arxiv.org/abs/1802.03601},
	doi = {10.1016/j.neucom.2018.05.083},
	urldate = {2025-11-12},
	journal = {Neurocomputing},
	author = {Wang, Mei and Deng, Weihong},
	month = oct,
	year = {2018},
	note = {arXiv:1802.03601 [cs]},
	pages = {135--153},
}

@inproceedings{wang_characterizing_2019,
	title = {Characterizing and {Avoiding} {Negative} {Transfer}},
	url = {https://openaccess.thecvf.com/content_CVPR_2019/html/Wang_Characterizing_and_Avoiding_Negative_Transfer_CVPR_2019_paper.html},
	urldate = {2025-11-12},
	author = {Wang, Zirui and Dai, Zihang and Poczos, Barnabas and Carbonell, Jaime},
	year = {2019},
	pages = {11293--11302},
}

@inproceedings{kumar_understanding_2020,
	title = {Understanding {Self}-{Training} for {Gradual} {Domain} {Adaptation}},
	url = {https://proceedings.mlr.press/v119/kumar20c.html},
	language = {en},
	urldate = {2025-11-12},
	booktitle = {Proceedings of the 37th {International} {Conference} on {Machine} {Learning}},
	publisher = {PMLR},
	author = {Kumar, Ananya and Ma, Tengyu and Liang, Percy},
	month = nov,
	year = {2020},
	note = {ISSN: 2640-3498},
	pages = {5468--5479},
}

@article{pape_microscopy-based_2021,
	title = {Microscopy-based assay for semi-quantitative detection of {SARS}-{CoV}-2 specific antibodies in human sera: {A} semi-quantitative, high throughput, microscopy-based assay expands existing approaches to measure {SARS}-{CoV}-2 specific antibody levels in human sera},
	volume = {43},
	issn = {1521-1878},
	shorttitle = {Microscopy-based assay for semi-quantitative detection of {SARS}-{CoV}-2 specific antibodies in human sera},
	doi = {10.1002/bies.202000257},
	language = {eng},
	number = {3},
	journal = {BioEssays: News and Reviews in Molecular, Cellular and Developmental Biology},
	author = {Pape, Constantin and Remme, Roman and Wolny, Adrian and Olberg, Sylvia and Wolf, Steffen and Cerrone, Lorenzo and Cortese, Mirko and Klaus, Severina and Lucic, Bojana and Ullrich, Stephanie and Anders-Össwein, Maria and Wolf, Stefanie and Cerikan, Berati and Neufeldt, Christopher J. and Ganter, Markus and Schnitzler, Paul and Merle, Uta and Lusic, Marina and Boulant, Steeve and Stanifer, Megan and Bartenschlager, Ralf and Hamprecht, Fred A. and Kreshuk, Anna and Tischer, Christian and Kräusslich, Hans-Georg and Müller, Barbara and Laketa, Vibor},
	month = mar,
	year = {2021},
	pmid = {33377226},
	pmcid = {PMC7883048},
	pages = {e2000257},
}

@inproceedings{vmamba,
  title={{Vmamba: Visual State Space Model}},
  author={Liu, Yue and Tian, Yunjie and Zhao, Yuzhong and Yu, Hongtian and Xie, Lingxi and Wang, Yaowei and Ye, Qixiang and Jiao, Jianbin and Liu, Yunfan},
  booktitle={The Thirty-eighth Annual Conference on Neural Information Processing Systems},
  year={2024}
}

@article{Ma2024,
  title={{U-Mamba: Enhancing Long-range Dependency for Biomedical Image Segmentation}},
  author={Ma, Jun and Li, Feifei and Wang, Bo},
  journal={arXiv preprint arXiv:2401.04722},
  year={2024}
}

@inproceedings{hatamizadeh2021swin,
  title={{Swin UNETR: Swin Transformers for Semantic
Segmentation of Brain Tumors in MRI Images}},
  author={Hatamizadeh, Ali and Nath, Vishwesh and Tang, Yucheng and Yang, Dong and Roth, Holger R and Xu, Daguang},
  booktitle={International MICCAI Brainlesion Workshop},
  pages={272--284},
  year={2021},
  organization={Springer}
}

@article{kivimaki_confidence-based_2025,
	title = {Confidence-based {Estimators} for {Predictive} {Performance} in {Model} {Monitoring}},
	volume = {82},
	issn = {1076-9757},
	url = {http://arxiv.org/abs/2407.08649},
	doi = {10.1613/jair.1.16709},
	urldate = {2026-01-23},
	journal = {Journal of Artificial Intelligence Research},
	author = {Kivimäki, Juhani and Białek, Jakub and Nurminen, Jukka K. and Kuberski, Wojtek},
	month = jan,
	year = {2025},
	note = {arXiv:2407.08649 [cs]},
	pages = {209--240},

}

@misc{valindria_reverse_2017,
	title = {Reverse {Classification} {Accuracy}: {Predicting} {Segmentation} {Performance} in the {Absence} of {Ground} {Truth}},
	shorttitle = {Reverse {Classification} {Accuracy}},
	url = {http://arxiv.org/abs/1702.03407},
	doi = {10.48550/arXiv.1702.03407},
	urldate = {2026-01-23},
	publisher = {arXiv},
	author = {Valindria, Vanya V. and Lavdas, Ioannis and Bai, Wenjia and Kamnitsas, Konstantinos and Aboagye, Eric O. and Rockall, Andrea G. and Rueckert, Daniel and Glocker, Ben},
	month = feb,
	year = {2017},
	note = {arXiv:1702.03407 [cs]},
}

@misc{bialek_estimating_2025,
	title = {Estimating {Model} {Performance} {Under} {Covariate} {Shift} {Without} {Labels}},
	url = {http://arxiv.org/abs/2401.08348},
	doi = {10.48550/arXiv.2401.08348},
	urldate = {2026-01-23},
	publisher = {arXiv},
	author = {Białek, Jakub and Kivimäki, Juhani and Kuberski, Wojtek and Perrakis, Nikolaos},
	month = oct,
	year = {2025},
	note = {arXiv:2401.08348 [cs]},
}

@article{roy_bayesian_2019,
	title = {Bayesian {QuickNAT}: {Model} uncertainty in deep whole-brain segmentation for structure-wise quality control},
	volume = {195},
	issn = {1053-8119},
	shorttitle = {Bayesian {QuickNAT}},
	url = {https://www.sciencedirect.com/science/article/pii/S1053811919302319},
	doi = {10.1016/j.neuroimage.2019.03.042},
	urldate = {2026-01-23},
	journal = {NeuroImage},
	author = {Roy, Abhijit Guha and Conjeti, Sailesh and Navab, Nassir and Wachinger, Christian},
	month = jul,
	year = {2019},
	pages = {11--22},
}

@article{sims_seg_2023,
	title = {{SEG}: {Segmentation} {Evaluation} in absence of {Ground} truth labels},
	issn = {2692-8205},
	shorttitle = {{SEG}},
	url = {https://pmc.ncbi.nlm.nih.gov/articles/PMC9980141/},
	doi = {10.1101/2023.02.23.529809},
	urldate = {2026-01-23},
	journal = {bioRxiv},
	author = {Sims, Zachary and Strgar, Luke and Thirumalaisamy, Dharani and Heussner, Robert and Thibault, Guillaume and Chang, Young Hwan},
	month = feb,
	year = {2023},
	pages = {2023.02.23.529809},
}

@article{lin_novel_2023,
	title = {A {Novel} {Quality} {Control} {Algorithm} for {Medical} {Image} {Segmentation} {Based} on {Fuzzy} {Uncertainty}},
	volume = {31},
	issn = {1941-0034},
	url = {https://ieeexplore.ieee.org/stampPDF/getPDF.jsp},
	doi = {10.1109/TFUZZ.2022.3228332},
	number = {8},
	urldate = {2026-01-26},
	journal = {IEEE Transactions on Fuzzy Systems},
	author = {Lin, Qiao and Chen, Xin and Chen, Chao and Garibaldi, Jonathan M.},
	month = aug,
	year = {2023},
	pages = {2532--2544},
}

@misc{deng_confidence_2023,
	title = {Confidence and {Dispersity} {Speak}: {Characterising} {Prediction} {Matrix} for {Unsupervised} {Accuracy} {Estimation}},
	shorttitle = {Confidence and {Dispersity} {Speak}},
	url = {http://arxiv.org/abs/2302.01094},
	doi = {10.48550/arXiv.2302.01094},
	urldate = {2026-01-26},
	publisher = {arXiv},
	author = {Deng, Weijian and Suh, Yumin and Gould, Stephen and Zheng, Liang},
	month = feb,
	year = {2023},
	note = {arXiv:2302.01094 [cs]},
}

@misc{devries_leveraging_2018,
	title = {Leveraging {Uncertainty} {Estimates} for {Predicting} {Segmentation} {Quality}},
	url = {http://arxiv.org/abs/1807.00502},
	doi = {10.48550/arXiv.1807.00502},
	urldate = {2026-01-26},
	publisher = {arXiv},
	author = {DeVries, Terrance and Taylor, Graham W.},
	month = jul,
	year = {2018},
	note = {arXiv:1807.00502 [cs]},
}

@inproceedings{yu_odp-bench_2025,
	title = {{ODP}-{Bench}: {Benchmarking} {Out}-of-{Distribution} {Performance} {Prediction}},
	shorttitle = {{ODP}-{Bench}},
	url = {https://openaccess.thecvf.com/content/ICCV2025/html/Yu_ODP-Bench_Benchmarking_Out-of-Distribution_Performance_Prediction_ICCV_2025_paper.html},
	language = {en},
	urldate = {2026-01-27},
	author = {Yu, Han and Li, Kehan and Li, Dongbai and He, Yue and Zhang, Xingxuan and Cui, Peng},
	year = {2025},
	pages = {1846--1858},
}

@misc{ng_predicting_2023,
	title = {Predicting {Out}-of-{Domain} {Generalization} with {Neighborhood} {Invariance}},
	url = {http://arxiv.org/abs/2207.02093},
	doi = {10.48550/arXiv.2207.02093},
	urldate = {2026-01-27},
	publisher = {arXiv},
	author = {Ng, Nathan and Hulkund, Neha and Cho, Kyunghyun and Ghassemi, Marzyeh},
	month = jul,
	year = {2023},
	note = {arXiv:2207.02093 [cs]},
}

@misc{baek_agreement---line_2023,
	title = {Agreement-on-the-{Line}: {Predicting} the {Performance} of {Neural} {Networks} under {Distribution} {Shift}},
	shorttitle = {Agreement-on-the-{Line}},
	url = {http://arxiv.org/abs/2206.13089},
	doi = {10.48550/arXiv.2206.13089},
	urldate = {2026-01-27},
	publisher = {arXiv},
	author = {Baek, Christina and Jiang, Yiding and Raghunathan, Aditi and Kolter, Zico},
	month = may,
	year = {2023},
	note = {arXiv:2206.13089 [cs]},
}

@misc{xie_importance_2023,
	title = {On the {Importance} of {Feature} {Separability} in {Predicting} {Out}-{Of}-{Distribution} {Error}},
	url = {http://arxiv.org/abs/2303.15488},
	doi = {10.48550/arXiv.2303.15488},
	urldate = {2026-01-27},
	publisher = {arXiv},
	author = {Xie, Renchunzi and Wei, Hongxin and Feng, Lei and Cao, Yuzhou and An, Bo},
	month = oct,
	year = {2023},
	note = {arXiv:2303.15488 [cs]},
}

@article{robinson_real-time_2018,
	series = {Lecture {Notes} in {Computer} {Science} (including subseries {Lecture} {Notes} in {Artificial} {Intelligence} and {Lecture} {Notes} in {Bioinformatics})},
	title = {Real-{Time} {Prediction} of {Segmentation} {Quality}: 21st {International} {Conference} on {Medical} {Image} {Computing} and {Computer} {Assisted} {Intervention}, {MICCAI} 2018},
	issn = {9783030009366},
	shorttitle = {Real-{Time} {Prediction} of {Segmentation} {Quality}},
	url = {https://www.scopus.com/pages/publications/85053850621},
	doi = {10.1007/978-3-030-00937-3_66},
	urldate = {2026-02-03},
	journal = {Medical Image Computing and Computer Assisted Intervention – MICCAI 2018 - 21st International Conference, 2018, Proceedings},
	publisher = {Springer Verlag},
	author = {Robinson, Robert and Oktay, Ozan and Bai, Wenjia and Valindria, Vanya V. and Sanghvi, Mihir M. and Aung, Nay and Paiva, José M. and Zemrak, Filip and Fung, Kenneth and Lukaschuk, Elena and Lee, Aaron M. and Carapella, Valentina and Kim, Young Jin and Kainz, Bernhard and Piechnik, Stefan K. and Neubauer, Stefan and Petersen, Steffen E. and Page, Chris and Rueckert, Daniel and Glocker, Ben},
	editor = {Frangi, Alejandro F. and Fichtinger, Gabor and Schnabel, Julia A. and Alberola-López, Carlos and Davatzikos, Christos},
	year = {2018},
	pages = {578--585},
}

@inproceedings{hoebel_exploration_2020,
	address = {Houston, United States},
	title = {An exploration of uncertainty information for segmentation quality assessment},
	isbn = {978-1-5106-3393-3 978-1-5106-3394-0},
	url = {https://www.spiedigitallibrary.org/conference-proceedings-of-spie/11313/2548722/An-exploration-of-uncertainty-information-for-segmentation-quality-assessment/10.1117/12.2548722.full},
	doi = {10.1117/12.2548722},
	language = {en},
	urldate = {2026-02-04},
	booktitle = {Medical {Imaging} 2020: {Image} {Processing}},
	publisher = {SPIE},
	author = {Hoebel, Katharina and Andrearczyk, Vincent and Beers, Andrew L. and Patel, Jay B. and Chang, Ken and Depeursinge, Adrien and Mueller, Henning and Kalpathy-Cramer, Jayashree},
	editor = {Landman, Bennett A. and Išgum, Ivana},
	month = mar,
	year = {2020},
	pages = {55},
}

@article{mehrtash_confidence_2020,
	title = {Confidence {Calibration} and {Predictive} {Uncertainty} {Estimation} for {Deep} {Medical} {Image} {Segmentation}},
	volume = {39},
	issn = {0278-0062},
	url = {https://pmc.ncbi.nlm.nih.gov/articles/PMC7704933/},
	doi = {10.1109/TMI.2020.3006437},
	number = {12},
	urldate = {2026-02-04},
	journal = {IEEE transactions on medical imaging},
	author = {Mehrtash, Alireza and Wells, William M. and Tempany, Clare M. and Abolmaesumi, Purang and Kapur, Tina},
	month = dec,
	year = {2020},
	pages = {3868--3878},
}

@article{zenk_comparative_2025,
	title = {Comparative benchmarking of failure detection methods in medical image segmentation: {Unveiling} the role of confidence aggregation},
	volume = {101},
	issn = {1361-8415},
	shorttitle = {Comparative benchmarking of failure detection methods in medical image segmentation},
	url = {https://www.sciencedirect.com/science/article/pii/S1361841524003177},
	doi = {10.1016/j.media.2024.103392},
	urldate = {2026-02-04},
	journal = {Medical Image Analysis},
	author = {Zenk, Maximilian and Zimmerer, David and Isensee, Fabian and Traub, Jeremias and Norajitra, Tobias and Jäger, Paul F. and Maier-Hein, Klaus},
	month = apr,
	year = {2025},
	pages = {103392},
}

@article{jungo_analyzing_2020,
	title = {Analyzing the {Quality} and {Challenges} of {Uncertainty} {Estimations} for {Brain} {Tumor} {Segmentation}},
	volume = {14},
	issn = {1662-4548},
	doi = {10.3389/fnins.2020.00282},
	language = {eng},
	journal = {Frontiers in Neuroscience},
	author = {Jungo, Alain and Balsiger, Fabian and Reyes, Mauricio},
	year = {2020},
	pages = {282},
}

@article{wasserthal_totalsegmentator_2023,
	title = {{TotalSegmentator}: robust segmentation of 104 anatomical structures in {CT} images},
	volume = {5},
	issn = {2638-6100},
	shorttitle = {{TotalSegmentator}},
	url = {http://arxiv.org/abs/2208.05868},
	doi = {10.1148/ryai.230024},
	number = {5},
	urldate = {2026-02-26},
	journal = {Radiology: Artificial Intelligence},
	author = {Wasserthal, Jakob and Breit, Hanns-Christian and Meyer, Manfred T. and Pradella, Maurice and Hinck, Daniel and Sauter, Alexander W. and Heye, Tobias and Boll, Daniel and Cyriac, Joshy and Yang, Shan and Bach, Michael and Segeroth, Martin},
	month = sep,
	year = {2023},
	note = {arXiv:2208.05868 [eess]},
	pages = {e230024},
}

@misc{lin_microsoft_2015,
	title = {Microsoft {COCO}: {Common} {Objects} in {Context}},
	shorttitle = {Microsoft {COCO}},
	url = {http://arxiv.org/abs/1405.0312},
	doi = {10.48550/arXiv.1405.0312},
	urldate = {2026-02-26},
	publisher = {arXiv},
	author = {Lin, Tsung-Yi and Maire, Michael and Belongie, Serge and Bourdev, Lubomir and Girshick, Ross and Hays, James and Perona, Pietro and Ramanan, Deva and Zitnick, C. Lawrence and Dollár, Piotr},
	month = feb,
	year = {2015},
	note = {arXiv:1405.0312 [cs]},
}

@article{everingham_pascal_2010,
	title = {The {Pascal} {Visual} {Object} {Classes} ({VOC}) {Challenge}},
	volume = {88},
	issn = {1573-1405},
	url = {https://doi.org/10.1007/s11263-009-0275-4},
	doi = {10.1007/s11263-009-0275-4},
	language = {en},
	number = {2},
	urldate = {2026-02-26},
	journal = {International Journal of Computer Vision},
	author = {Everingham, Mark and Van Gool, Luc and Williams, Christopher K. I. and Winn, John and Zisserman, Andrew},
	month = jun,
	year = {2010},
	pages = {303--338},
}

\appendix

\clearpage
\setcounter{page}{1}

\section*{\centering \textbf{\title{}}}
{\centering Supplementary Material\par}

\section{Datasets}
\label{sec:dataset}

The general experimental setup for source model ranking under transfer to a target dataset is as follows, a set of `source' models $\mathcal{M}=\{M_{i}\}_{i=1}^{N_{m}}$ are trained for the same task (i.e nuclei semantic segmentation). The models can have a range of architectures and source datasets. A single dataset is then selected as `target' and all of the models from $\mathcal{M}$ are applied to the test set of the target data. We then rank the models $M_{i=1\dots N_{m}}$ and compare this ranking to the true transfer performance ranking, obtained using ground-truth. In this section we further detail the employed datasets, and in section \cref{sec:supp_models} the tested models.

We experimented applying our consistency based model ranking (CMR) method across a wide range of datasets and modalities from the microscopy and biomedical domains. We focused on ranking the target performance of semantic and instance segmentation models across several distinct tasks. The tasks covered arguably the most common problems in microscopy image analysis; nuclei, cell and mitochondria segmentation. We additionally investigated a popular biomedical segmentation challenge (i.e, ToothFairy2) for human jaw segmentation. Our wide range of experiments across such a diverse set of datasets and tasks highlights the general applicability of our approach for unsupervised ranking of segmentation models. 

\subsection{Electron Microscopy (EM), Mitochondria}
\label{sec:data_sem_EM}
\smallskip
\noindent \textbf{Semantic Segmentation.}
We use four datasets with semantic segmentation ground-truth, all are neural tissues taken from different species and EM modalities: 
\begin{itemize}
    \item EPFL (rat)~\cite{lucchi_learning_2013}
    \begin{itemize}
        \item The dataset represents a section taken from the CA1 hippocampus region of the brain. The dataset has the rescaled dimensions $330\times480\times640$.
        \item The volume was split with the first 165 z-slices used for training, the next 40 z-slices used for validation and the remaining 125 z-slices used for testing.
    \end{itemize}
    \item Hmito, Rmito~\cite{franco-barranco_current_2023} 
    \begin{itemize}
        \item The dataset comprises two 3D EM image stacks taken from the brain tissue of a human (Hmito) and a rat (Rmito). Each volume has pixel dimensions $500\times4096\times4096$.
        \item The volumes were split with the first 300 z-slices used for training, the next 50 z-slices for validation and the remaining 150 z-slices for testing.
    \end{itemize}
    \item VNC (fruit fly)~\cite{phelps_reconstruction_2021}
    \begin{itemize}
        \item Transmission Electron Microscopy of ventral nerve cord (VNC) of an adult female \textit{Drosophila melanogaster}. Once voxels were rescaled the dataset had pixel dimensions $20\times589\times589$.
        \item The VNC dataset is by far the smallest dataset, therefore, in order to maximise available patches for testing we did not perform a typical test train split. For VNC source models we used the first 18 z-slices for training and the remaining 2 z-slices for validation. We did not transfer VNC source trained models to a VNC test set, removing this transfer from our set of evaluated transfers. For models trained on alternative sources and transferred to VNC we thus tested on the entire 20 z-slices of VNC.
    \end{itemize}
\end{itemize}

\noindent All datasets were resized to match the MitoEM voxel dimensions. For all datasets we split the volumes into 2D $256 \times 256$ image patches for training and evaluation.

\smallskip
\noindent \textbf{Classification.} In addition to direct use for semantic segmentation, we reformulated these datasets as a per-patch binary classification task, labelling a patch as positive if mitochondria are present. The volumes were split into $80\times80$ 2D patches. To remove positive images with very little mitochondria present, we additionally filtered images based on proportion of mitochondria present in each ground-truth patch. For each dataset we used the corresponding foreground ratios; $\text{EPFL} = 0.1$ foreground, $\text{Hmito} = 0.1$ foreground, $\text{Rmito} = 0.1$ foreground, $\text{VNC} = 0.05$ foreground.   

This auxiliary classification task allows us to disentangle the effects of domain shift from the intrinsic suitability of classification-based ranking methods for semantic segmentation.
\subsection{Light Microscopy (LM), Nuclei}

\smallskip
\noindent \textbf{Semantic Segmentation.} We trained nuclei segmentation models on the following fluorescent datasets;
\begin{itemize}
    \item BBBC039~\cite{ljosa_annotated_2012}
    \begin{itemize}
        \item A high-throughput chemical screen on U2OS cells.
        \item The dataset comprises of a set of 2D images. We used 100 images for training, 50 for validation and 48 for testing. (See github repo for exact image id split).
        \item From each image 2D $256\times256$ patches were extracted for training and images were cropped to a standard $512 \times 512$ for evaluation.
    \end{itemize}
    \item Go-Nuclear~\cite{vijayan_deep_2024}
    \begin{itemize}
        \item Fluorescent dataset of \textit{Arabidopsis thaliana}.
        \item The dataset comprises of 5 separate data volumes. Three volumes (`id: 1135' ($263 \times 1024 \times 1024$), `id: 1136' ($268\times1120\times1120$) and `id: 1137' ($262\times1080\times1080$) were used as training data. For each volume the 40-210 z-slices were taken to avoid imaging artefacts at beginning an end of the volumes. One volume `id: 1139' ($280\times1120\times1120$) was used for validation and the last volume `id: 1160' ($241\times753\times1672$) was used for testing.
        \item From each volume 2D $256\times256$ patches were extracted for training and evaluation.
    \end{itemize}
    \item HeLaCytoNuc~\cite{de_helacytonuc_2024}
    \begin{itemize}
        \item A dataset of HeLa cell nuclei.
        \item The dataset consists of 1873 training images, 535 validation images and 267 test images. Each image has pixel dimensions $520\times696$.
        \item From each image overlapping 2D $256\times256$ patches were extracted for training and whole images were taken for evaluation.
    \end{itemize}
    \item Hoechst~\cite{arvidsson_annotated_2023}
    \begin{itemize}
        \item A modified U2OS osteosarcoma cell line.
        \item The dataset consists of 30 training images and 10 test images. Each image has pixel dimensions $1104\times1104$
        \item From each image overlapping 2D $256\times256$ patches were extracted for training and whole images were taken for evaluation.
    \end{itemize}
    \item S\_BIAD895/SB-895~\cite{von_chamier_democratising_2021}
    \begin{itemize}
        \item The dataset consists of 47 images with pixel dimensions $1024\times1024$. We did not split the dataset into test and train splits as we did not apply models trained on S\_BIAD895 to S\_BIAD895 data, removing this transfer from our set of evaluated transfers. We only considered S\_BIAD895 models under transfer to other target datasets. 
        \item Thus for models transferred to S\_BIAD895, the full image set could be used for testing.
        \item From each image 2D $256\times256$ patches were extracted for training and whole images were taken for evaluation.
    \end{itemize}
    \item The SELMA3D 2024 challenge~\cite{leal-taixe_benchmark_2019}.
    \begin{itemize}
        \item The dataset consists of eight $200\times200\times200$ training volumes, one $200\times200\times200$ validation volume and three $200\times200\times200$ test volumes.
        \item We took 2D $200\times200$ patches from each z-slice for training and evaluation.
    \end{itemize}
    \item S\_BIAD1410~\cite{hawkins_rescu-nets_2024}
    \begin{itemize}
        \item A dataset of fruit fly embryonic development.
        \item The dataset comprises of ten training volumes with the following dimensions (($401\times512\times512$), ($240\times512\times512$), ($400\times512\times512$), ($380\times512\times512$) and six times ($121\times512\times512$)).
        \item From each volume 2D $256\times256$ patches were extracted for training and evaluation.
    \end{itemize}
\end{itemize}

The set of source models trained on the seven available datasets were then transferred to four target datasets; BBBC039~\cite{ljosa_annotated_2012}, Hoechst~\cite{arvidsson_annotated_2023}, S\_BIAD895~\cite{von_chamier_democratising_2021} and DSB2018~\cite{caicedo_nucleus_2019} (a fluorescent only subset of the data science bowl 2018 challenge). The target datasets were chosen based on two criteria, firstly, for the quality of their ground-truth annotation, to allow for realistic performance evaluation scores. Secondly, the performance of models transferred to the target dataset should have some differentiation. Model differentiation allows for proper investigation of ranking metrics. If all the models transferred to a dataset perform equally then ranking becomes obsolete and hard to evaluate. 

\smallskip
\noindent \textbf{Instance Segmentation.} We trained nuclei instance segmentation models on five datasets; BBBC039~\cite{ljosa_annotated_2012}, HeLaCytoNuc~\cite{de_helacytonuc_2024}, Hoechst~\cite{arvidsson_annotated_2023}, 
S\_BIAD895/SB-895~\cite{von_chamier_democratising_2021} and SBIAD1410~\cite{hawkins_rescu-nets_2024}. The source models were then transferred to the test sets of four target datasets BBBC039~\cite{ljosa_annotated_2012}, Hoechst~\cite{arvidsson_annotated_2023}, 
S\_BIAD895/SB-895~\cite{von_chamier_democratising_2021} and one additional dataset S\_BIAD634/SB-634~\cite{kromp_annotated_2020} was introduced. S\_BIAD634/SB-634 is a dataset of nuclei images of normal and cancer cells from different tissue origins and sample preparation types. The dataset comprises of 42 $1024\times 1280$ training images and 37 $1024\times 1280$ test images. Once again Target datasets were chosen with the same two criteria for instance ground-truth quality and differentiable model performance under transfer. For details on training of instance segmentation models see \cref{sec:supp_models}.

\subsection{Light Microscopy (LM), Cells}
\smallskip
\noindent \textbf{Instance Segmentation.}
We consider four dense instance segmentation problems for cell datasets: 
\begin{itemize}
    \item FlyWing~\cite{leal-taixe_benchmark_2019}
    \begin{itemize}
        \item Fluorescent volume of a developing fruit fly wing.
        \item The dataset comprises of four training volumes with the following pixel dimensions (($160\times773\times881$), ($160\times773\times881$), ($148\times765\times877$) and ($180\times776\times893$)), two validation volumes both with dimensions ($160\times590\times773$) and two test volumes both with dimensions ($160\times639\times765$).
        \item From each volume 2D $256\times256$ patches were extracted for training and full z-slices taken for evaluation.
    \end{itemize}
    \item Ovules~\cite{wolny_accurate_2020}
    \begin{itemize}
        \item Fluorescent volume of a \textit{Arabidopsis thaliana}
        \item We used 20 volumes (see data for dimensions) for training, a further single volume ($397\times880\times1332$) for validation and a single volume ($320\times960\times1000$) for testing.
        \item From each volume 2D $256\times256$ patches were extracted for training and full z-slices taken for evaluation.
    \end{itemize}
    \item PNAS~\cite {willis_cell_2016}
    \begin{itemize}
        \item Fluorescent volume of the \textit{Arabidopsis thaliana} apical stem cell niche.
        \item For training we used 103 volumes collected at 4 hour time points between 0-84hrs from four plants (see data for dimensions). Volumes from two separate plants were used for validation and testing.
        \item From each volume 2D $256\times256$ images were extracted for training and full z-slices taken for evaluation.
    \end{itemize}
    \item Covid-IF~\cite{pape_microscopy-based_2021}
    \begin{itemize}
        \item Immunofluorescence imaging of human serum.
        \item The dataset comprises of 48 $1024\times1024$ images, which are all used for testing since we didn't train any models on this dataset and only use it to evaluate publicly available pre-trained models.
    \end{itemize}
\end{itemize}

\subsection{Cone-Beam Computed Tomography (CBCT), ToothFairy2}
\smallskip
\noindent {\textbf{Multiclass Semantic Segmentation.}
For multiclass semantic segmentation experiments, we use ToothFairy2 from a MICCAI2024 challenge ~\cite{2024TMI,2025CVPR}. It comprises 530 volumetric scans with voxel-wise annotations of maxillofacial anatomy, of which 480 are publicly available for training, while 50, acquired by a different institution, are held out for evaluation via the grand-challenge platform. The dataset provides annotations for 42 classes, including jawbones, left and right inferior alveolar canals, maxillary sinuses, pharynx, upper and lower teeth, bridges, crowns, and implants. For evaluation, we obtained access to the private test set, collected by a different centre on different acquisition machines. 
}

\section{Models}
\label{sec:supp_models}
We trained a range of model architectures to build a diverse model set for transfer comparison. All of the models trained by us are trained on the training set of a single `source' dataset. We additionally investigated several publicly available models taken from the Toothfairy2 challenge~\cite{2024TMI}, the Bioimage Model Zoo~\cite{ouyang_bioimage_2022} and large generalist models taken from popular repositories (Cellpose-SAM~\cite{pachitariu_cellpose-sam_2025}, $\mu$SAM~\cite{archit_segment_2025}, TotalSegmentator~\cite{wasserthal_totalsegmentator_2023}. Information listing the full set of models considered can be found in \cref{sec:models} in the main paper. In the following section we detail the model training setup.

All the self-trained networks were trained until convergence on the source dataset and performed highly on the held out source test set. The classification networks were trained with a Binary Cross entropy (BCE) loss, while the segmentation networks used a combination of Dice and CE losses. For models trained for the Toothfairy2 challenge dataset, the nnUNet planner was used for preprocessing and hyperparameters selection. The models were trained from scratch on a single 48GB A40 Nvidia GPU using CUDA 11.8 and PyTorch 2.1.2. All other models were trained from scratch on a set of 8 12GB NVIDIA GeForce RTX 3080 Ti GPUs using CUDA12.1 and Pytorch 2.5.1.

\subsection{Instance Segmentation Training}

For the instance segmentation models trained by us we follow a popular instantiation process~\cite{pape_microscopy-based_2021} of predicting cell boundaries and then using seeded watershed and graph partitioning algorithm GASP~\cite{bailoni_gasp_2022} to obtain instances.

\section{Model Perturbation}
\label{sec:perturbations}
\smallskip
\noindent\textbf{Input Perturbations:} We tested a range of different input perturbations by applying several popular image transformations at test-time: additive Gaussian noise, gamma correction, and changes in brightness and contrast. The transformations are defined as follows,
\begin{equation}
    \text{AdditiveGaussianNoise:} \;\;\;\;  x'= x + \mathcal{N}(0, \sigma),
\end{equation}
\begin{equation}
    \text{RandomBrightness:} \;\;\;\; x'= x + \theta_{B},
\end{equation}
\begin{equation}
    \text{RandomContrast:} \;\;\;\; x'= \mu(x) + \theta_{C}(x-\mu(x)),
\end{equation}
\begin{equation}
    \text{RandomGammaCorrection:} \;\;\;\; x'=x^{\gamma},
\end{equation}
where the strength of each perturbation can be controlled by $\sigma, \;\theta_{B}, \;\theta_{C} \; \text{and}\; \gamma$ parameters respectively.

\smallskip
\noindent\textbf{Feature Space Perturbations:} We relied upon test-time application of spatial DropOut~\cite{tompson_efficient_2015} to provide feature space perturbation. We tested several approaches for applying DropOut to our models. Firstly, we tested applying Spatial DropOut only to the bottleneck layer of U-Net style architectures. Our experiments for bottleneck only DropOut are shown on the nuclei datasets both for semantic and instance segmentation (\cref{tab:CMR_EI_nuclei_dropout,tab:CMR_NHD_nuclei_dropout,tab:CMR_ars_nuclei_dropout}). Secondly, in addition to the bottle neck layer we also investigated adding DropOut to all layers that connect the encoder to the decoder, namely skip connections in U-Net style networks. Our experiments for bottleneck + skip connection DropOut are shown on the Toothfairy2 dataset for multiclass semantic segmentation (\cref{tab:CMR_ei_toothfairy}). Lastly, we investigate applying DropOut uniformly to every layer of the network, this represents the most general and architecture agnostic approach considered. Our experiments for all layer DropOut are shown on the mitochondria semantic segmentation datasets along with the cell instance segmentation datasets (\cref{tab:CMR_EI_mito_dropout,tab:CMR_NHD_mito_dropout,tab:CMR_ars_cells_dropout}). 

We found that all the considered feature space perturbation approaches were able to perform effectively, as shown in the tables in the following sections. However, it is worth noting that the first approach of applying DropOut to the bottleneck layer only can be sensitive to residual connections. The second approach of applying DropOut to both the bottleneck layer and skip connections works very well on the model sets tested, but does assume that the models have skip connections in the architecture, otherwise it reduces to the first case (bottleneck only) and could suffer the same sensitivity to residual connections. The last approach of applying DropOut to all layers is the most architecture agnostic and resulted in very strong correlation between the CMR metric and performance score.

\section{Unsupervised Domain Adaptation (UDA) Approaches}

When considering the reuse of pre-trained models users can either directly apply a model with frozen weights, in what we refer to as zero-shot reuse, or after applying some unsupervised domain adaptation (UDA) method to models in order to adapt them to the target data. Application of UDA has the potential to improve model performance on the target data, but requires users to have the resources and know-how to properly make use of existing methods. Furthermore, UDA methods can often be unstable and can even lead to degradation of model performance on the target data~\cite{wang_characterizing_2019}. Hence, even if UDA is applied the ranking task still persists, and now the best adapted model must be selected.

In order to investigate ranking of post-UDA models we utilised two common UDA approaches to adapt our pretrained source networks to target data. Firstly, we used a Mean Teacher~\cite{tarvainen_mean_2018} approach that relies on a `student' `teacher' paired network setting. The Mean Teacher framework for self supervised training initialises both the student and teacher networks using the pre-trained source model weights and then relies on the `teacher' branch to provide pseudo labels via its predictions. The student branch is trained via gradient descent using a consistency loss to the pseudo labels. Finally, the teacher weights are then updated via an exponential moving average of the student weights. We did not employ any additional pseudo-label selection steps as we found `no-selection' was a sufficient strategy in many cases for successful UDA. In addition, in our experiments we are not actually directly interested in improving the performance of UDA, but rather our real question is: can we rank the final post-UDA models regardless of whether UDA successfully improved target performance or indeed led to model degradation?

The second UDA approach we employed was Adaptive Batch Normalization~\cite{li_adaptive_2018} (AdaBN). AdaBN provides a simple and parameter-free approach for UDA and is based on updating the statistics of all Batch Normalization layers in a network to match the target data. This simple approach proved surprisingly effective at improving post-UDA model performance.

For both the Mean Teacher and AdaBN  approaches we were able to successfully rank the post-UDA model performance, outcompeting the state-of-the-art baseline Transfer Score~\cite{yang_can_2024}. In \cref{tab:finetuned_semantic}, we showed the results for single CMR perturbation strengths, in \cref{tab:CMR_post_UDA} we show the CMR correlation to post-UDA F1-score across a range of perturbation strengths.

\section{Semantic Segmentation}
In the following section we investigate ranking of semantic segmentation models, each of the tables \cref{tab:CMR_ei_mito_gauss,tab:CMR_EI_mito_dropout,tab:CMR_ei_nuclei_gauss,tab:CMR_EI_nuclei_dropout,tab:CMR_NHD_mito_gauss,tab:CMR_NHD_mito_dropout,tab:CMR_NHD_nuclei_gauss,tab:CMR_NHD_nuclei_dropout} shows the correlation scores between our CMR metrics (CMR-EI and CMR-NHD) and the performance F1-score for sets of models applied directly to a single target dataset, across a range of perturbation strengths. In the main paper, due to space constraints, we show the correlation scores for a single perturbation strength averaged over the set of target datasets.

We investigate the dependency of CMR on the strength of perturbation applied for both input and feature space perturbations. The tables show that CMR's performance is stable across a wide range of perturbations.

\subsection{Mitochondria Perturbation Sweep}

For mitochondria semantic segmentation we investigated ranking a model set with $|\mathcal{M}|=15$. The models were trained separately using all four Mitochondria datasets (EPFL~\cite{lucchi_learning_2013}, Hmito~\cite{franco-barranco_current_2023}, Rmito~\cite{franco-barranco_current_2023} and VNC~\cite{phelps_reconstruction_2021}) as different sources. The model set comprised of a range of model architectures: 2D U-Net (trained both with and without augmentations), Residual U-Net and UNETR. Hence we built a diverse set of domain-specialist models which we apply to each target dataset in turn and rank the corresponding performance via CMR.

To investigate input space perturbations we applied additive Gaussian noise to the inputs between the strengths $\sigma = 0.01 - 0.2$. We show the results for both CMR-EI (\cref{tab:CMR_ei_mito_gauss}) and CMR-NHD (\cref{tab:CMR_NHD_mito_gauss}).

For feature space perturbation we investigated applying spatial DropOut equally to all layers of a network, where the strength of the perturbation is controlled by the proportion of feature maps dropped at each layer $p_{d}$. We investigated DropOut proportions in the range $p_{d}= 0.001 - 0.1$. We show results for for both CMR-EI (\cref{tab:CMR_EI_mito_dropout}) and CMR-NHD (\cref{tab:CMR_NHD_mito_dropout}). The tables show that CMR's performance is stable across a wide range of perturbations.

\begin{table}[htb]
\centering
\scriptsize
\caption{Per-dataset correlation to F1-score ranking for Semantic Segmentation of Mitochondria~\cite{lucchi_learning_2013,franco-barranco_current_2023,phelps_reconstruction_2021} across a range of Gaussian Input Perturbation Strengths. For each ranking experiment $|\mathcal{M}|=15$. $\sigma$ was randomly sampled from the noted range. \textit{pval.} $<$ 0.05 (*), \textit{pval.} $<$ 0.01 (**).}

\begin{subtable}[t]{0.48\textwidth}
\centering
\tiny
\caption{CMR-EI Correlation scores}
\setlength{\tabcolsep}{0.85pt}
\begin{tabular}{@{}c c
   >{\columncolor{GreyTable}}c
   >{\columncolor{GreyTable}}c
   c c
   >{\columncolor{GreyTable}}c
   >{\columncolor{GreyTable}}c
   c c@{}}
\toprule
\multirow{2}{*}{Metric} & {} & \multicolumn{2}{c}{EPFL} & \multicolumn{2}{c}{Hmito} & \multicolumn{2}{c}{Rmito} & \multicolumn{2}{c}{VNC} \\
&  &  \cellcolor{SecondaryColumnColor} & \cellcolor{SecondaryColumnColor}\textit{pval.} &  & \textit{pval.} &  \cellcolor{SecondaryColumnColor} & \cellcolor{SecondaryColumnColor}\textit{pval.} &  & \textit{pval.} \\
\cmidrule{1-10}
\multirow{3}{*}{\begin{tabular}{@{}c@{}}CMR-EI \\ \textit{Gauss, $\sigma$} \\ \textit{$[0.01,0.03]$}\end{tabular}} & K$\tau$ & 0.73 & (**) & 0.66 & (**) & 0.77 & (**) & 0.46 & (0.06) \\
& S$\rho$ & 0.88 & (**) & 0.79 & (**) & 0.90 & (**) & 0.61 & (0.06) \\
& P$r$ & 0.82 & (**) & 0.75 & (**) & 0.87 & (**) & 0.63 & (*) \\
\cmidrule{1-10}
\multirow{3}{*}{\begin{tabular}{@{}c@{}}CMR-EI \\ \textit{Gauss, $\sigma$} \\ \textit{$[0.03,0.05]$}\end{tabular}} & K$\tau$ & 0.77 & (**) & 0.73 & (**) & 0.79 & (**) & 0.52 & (*) \\
& S$\rho$ & 0.90 & (**) & 0.86 & (**) & 0.90 & (**) & 0.63 & (*) \\
& P$r$ & 0.86 & (**) & 0.82 & (**) & 0.84 & (**) & 0.67 & (*) \\
\cmidrule{1-10}
\multirow{3}{*}{\begin{tabular}{@{}c@{}}CMR-EI \\ \textit{Gauss, $\sigma$} \\ \textit{$[0.05,0.07]$}\end{tabular}} & K$\tau$ & 0.81 & (**) & 0.73 & (**) & 0.82 & (**) & 0.61 & (**) \\
& S$\rho$ & 0.93 & (**) & 0.86 & (**) & 0.92 & (**) & 0.66 & (*) \\
& P$r$ & 0.88 & (**) & 0.83 & (**) & 0.82 & (**) & 0.71 & (*) \\
\cmidrule{1-10}
\multirow{3}{*}{\begin{tabular}{@{}c@{}}CMR-EI \\ \textit{Gauss, $\sigma$} \\ \textit{$[0.07,0.1]$}\end{tabular}} & K$\tau$ & 0.83 & (**) & 0.75 & (**) & 0.90 & (**) & 0.55 & (*) \\
& S$\rho$ & 0.93 & (**) & 0.87 & (**) & 0.95 & (**) & 0.63 & (*) \\
& P$r$ & 0.91 & (**) & 0.82 & (**) & 0.88 & (**) & 0.73 & (**) \\
\cmidrule{1-10}
\multirow{3}{*}{\begin{tabular}{@{}c@{}}CMR-EI \\ \textit{Gauss, $\sigma$} \\ \textit{$[0.1,0.12]$}\end{tabular}} & K$\tau$ & 0.87 & (**) & 0.73 & (**) & 0.88 & (**) & 0.58 & (**) \\
& S$\rho$ & 0.95 & (**) & 0.87 & (**) & 0.95 & (**) & 0.65 & (*) \\
& P$r$ & 0.91 & (**) & 0.80 & (**) & 0.85 & (**) & 0.75 & (**) \\
\cmidrule{1-10}
\multirow{3}{*}{\begin{tabular}{@{}c@{}}CMR-EI \\ \textit{Gauss, $\sigma$} \\ \textit{$[0.12,0.15]$}\end{tabular}} & K$\tau$ & 0.85 & (**) & 0.73 & (**) & 0.84 & (**) & 0.58 & (**) \\
& S$\rho$ & 0.94 & (**) & 0.87 & (**) & 0.94 & (**) & 0.67 & (*) \\
& P$r$ & 0.90 & (**) & 0.81 & (**) & 0.84 & (**) & 0.78 & (**) \\
\cmidrule{1-10}
\multirow{3}{*}{\begin{tabular}{@{}c@{}}CMR-EI \\ \textit{Gauss, $\sigma$} \\ \textit{$[0.15,0.2]$}\end{tabular}} & K$\tau$ & 0.77 & (**) & 0.71 & (**) & 0.84 & (**) & 0.55 & (*) \\
& S$\rho$ & 0.92 & (**) & 0.86 & (**) & 0.94 & (**) & 0.62 & (*) \\
& P$r$ & 0.86 & (**) & 0.84 & (**) & 0.89 & (**) & 0.78 & (**) \\
\bottomrule
\end{tabular}
\label{tab:CMR_ei_mito_gauss}
\end{subtable}%
\hfill
\begin{subtable}[t]{0.48\textwidth}
\centering
\tiny
\caption{CMR-NHD Correlation scores}
\setlength{\tabcolsep}{0.85pt}
\begin{tabular}{@{}c c
   >{\columncolor{GreyTable}}c
   >{\columncolor{GreyTable}}c
   c c
   >{\columncolor{GreyTable}}c
   >{\columncolor{GreyTable}}c
   c c@{}}
\toprule
\multirow{2}{*}{Metric} & {} & \multicolumn{2}{c}{EPFL} & \multicolumn{2}{c}{Hmito} & \multicolumn{2}{c}{Rmito} & \multicolumn{2}{c}{VNC} \\
&  &  \cellcolor{SecondaryColumnColor} & \cellcolor{SecondaryColumnColor}\textit{pval.} &  & \textit{pval.} &  \cellcolor{SecondaryColumnColor} & \cellcolor{SecondaryColumnColor}\textit{pval.} &  & \textit{pval.} \\
\cmidrule{1-10}
\multirow{3}{*}{\begin{tabular}{@{}c@{}}CMR-NHD \\ \textit{Gauss, $\sigma$} \\ \textit{$[0.01,0.03]$}\end{tabular}} & K$\tau$ & 0.73 & (**) & 0.66 & (**) & 0.71 & (**) & 0.52 & (*) \\
& S$\rho$ & 0.89 & (**) & 0.82 & (**) & 0.90 & (**) & 0.63 & (0.05) \\
& P$r$ & 0.83 & (**) & 0.75 & (**) & 0.54 & (*) & 0.72 & (**) \\
\cmidrule{1-10}
\multirow{3}{*}{\begin{tabular}{@{}c@{}}CMR-NHD \\ \textit{Gauss, $\sigma$} \\ \textit{$[0.03,0.05]$}\end{tabular}} & K$\tau$ & 0.77 & (**) & 0.67 & (**) & 0.73 & (**) & 0.48 & (*) \\
& S$\rho$ & 0.93 & (**) & 0.83 & (**) & 0.91 & (**) & 0.59 & (0.05) \\
& P$r$ & 0.83 & (**) & 0.74 & (**) & 0.61 & (*) & 0.74 & (**) \\
\cmidrule{1-10}
\multirow{3}{*}{\begin{tabular}{@{}c@{}}CMR-NHD \\ \textit{Gauss, $\sigma$} \\ \textit{$[0.05,0.07]$}\end{tabular}} & K$\tau$ & 0.79 & (**) & 0.68 & (**) & 0.71 & (**) & 0.48 & (*) \\
& S$\rho$ & 0.94 & (**) & 0.83 & (**) & 0.90 & (**) & 0.59 & (0.05) \\
& P$r$ & 0.84 & (**) & 0.76 & (**) & 0.59 & (*) & 0.69 & (*) \\
\cmidrule{1-10}
\multirow{3}{*}{\begin{tabular}{@{}c@{}}CMR-NHD \\ \textit{Gauss, $\sigma$} \\ \textit{$[0.07,0.1]$}\end{tabular}} & K$\tau$ & 0.73 & (**) & 0.69 & (**) & 0.75 & (**) & 0.55 & (*) \\
& S$\rho$ & 0.91 & (**) & 0.85 & (**) & 0.92 & (**) & 0.64 & (*) \\
& P$r$ & 0.90 & (**) & 0.74 & (**) & 0.68 & (**) & 0.72 & (**) \\
\cmidrule{1-10}
\multirow{3}{*}{\begin{tabular}{@{}c@{}}CMR-NHD \\ \textit{Gauss, $\sigma$} \\ \textit{$[0.1,0.12]$}\end{tabular}} & K$\tau$ & 0.77 & (**) & 0.71 & (**) & 0.77 & (**) & 0.55 & (*) \\
& S$\rho$ & 0.92 & (**) & 0.85 & (**) & 0.92 & (**) & 0.64 & (*) \\
& P$r$ & 0.92 & (**) & 0.73 & (**) & 0.69 & (**) & 0.75 & (**) \\
\cmidrule{1-10}
\multirow{3}{*}{\begin{tabular}{@{}c@{}}CMR-NHD \\ \textit{Gauss, $\sigma$} \\ \textit{$[0.12,0.15]$}\end{tabular}} & K$\tau$ & 0.81 & (**) & 0.69 & (**) & 0.77 & (**) & 0.55 & (*) \\
& S$\rho$ & 0.93 & (**) & 0.82 & (**) & 0.92 & (**) & 0.64 & (*) \\
& P$r$ & 0.91 & (**) & 0.75 & (**) & 0.71 & (**) & 0.77 & (**) \\
\cmidrule{1-10}
\multirow{3}{*}{\begin{tabular}{@{}c@{}}CMR-NHD \\ \textit{Gauss, $\sigma$} \\ \textit{$[0.15,0.2]$}\end{tabular}} & K$\tau$ & 0.75 & (**) & 0.69 & (**) & 0.75 & (**) & 0.55 & (*) \\
& S$\rho$ & 0.91 & (**) & 0.82 & (**) & 0.91 & (**) & 0.65 & (*) \\
& P$r$ & 0.88 & (**) & 0.79 & (**) & 0.78 & (**) & 0.77 & (**) \\
\bottomrule
\end{tabular}
\label{tab:CMR_NHD_mito_gauss}
\end{subtable}

\end{table}
\begin{table}[htb]
\centering
\scriptsize
\caption{Per-dataset correlation to F1-score ranking for Semantic Segmentation of Mitochondria~\cite{lucchi_learning_2013,franco-barranco_current_2023,phelps_reconstruction_2021} across a range of DropOut Perturbation Strengths. DropOut was applied equally to all layers of the networks. For each ranking experiment $|\mathcal{M}|=15$. \textit{pval.} $<$ 0.05 (*), \textit{pval.} $<$ 0.01 (**).}

\begin{subtable}[t]{0.48\textwidth}
\centering
\tiny
\caption{CMR-EI Correlation scores}
\setlength{\tabcolsep}{0.85pt}
\begin{tabular}{@{}c c
   >{\columncolor{GreyTable}}c
   >{\columncolor{GreyTable}}c
   c c
   >{\columncolor{GreyTable}}c
   >{\columncolor{GreyTable}}c
   c c@{}}
\toprule
\multirow{2}{*}{Metric} & {} & \multicolumn{2}{c}{EPFL} & \multicolumn{2}{c}{Hmito} & \multicolumn{2}{c}{Rmito} & \multicolumn{2}{c}{VNC} \\
&  &  \cellcolor{SecondaryColumnColor} & \cellcolor{SecondaryColumnColor}\textit{pval.} &  & \textit{pval.} &  \cellcolor{SecondaryColumnColor} & \cellcolor{SecondaryColumnColor}\textit{pval.} &  & \textit{pval.} \\
\cmidrule{1-10}
\multirow{3}{*}{\begin{tabular}{@{}c@{}}CMR-EI \\ \textit{DropOut} \\ \textit{$p_{d}=$ 0.001}\end{tabular}} & K$\tau$ & 0.75 & (**) & 0.56 & (**) & 0.73 & (**) & 0.48 & (*) \\
& S$\rho$ & 0.93 & (**) & 0.78 & (**) & 0.90 & (**) & 0.60 & (*) \\
& P$r$ & 0.91 & (**) & 0.76 & (**) & 0.90 & (**) & 0.78 & (**) \\
\cmidrule{1-10}
\multirow{3}{*}{\begin{tabular}{@{}c@{}}CMR-EI \\ \textit{DropOut} \\ \textit{$p_{d}=$ 0.005}\end{tabular}} & K$\tau$ & 0.75 & (**) & 0.58 & (**) & 0.77 & (**) & 0.48 & (*) \\
& S$\rho$ & 0.93 & (**) & 0.79 & (**) & 0.91 & (**) & 0.64 & (*) \\
& P$r$ & 0.91 & (**) & 0.76 & (**) & 0.92 & (**) & 0.79 & (**) \\
\cmidrule{1-10}
\multirow{3}{*}{\begin{tabular}{@{}c@{}}CMR-EI \\ \textit{DropOut} \\ \textit{$p_{d}=$ 0.01}\end{tabular}} & K$\tau$ & 0.83 & (**) & 0.58 & (**) & 0.75 & (**) & 0.55 & (*) \\
& S$\rho$ & 0.95 & (**) & 0.78 & (**) & 0.90 & (**) & 0.68 & (*) \\
& P$r$ & 0.94 & (**) & 0.77 & (**) & 0.92 & (**) & 0.79 & (**) \\
\cmidrule{1-10}
\multirow{3}{*}{\begin{tabular}{@{}c@{}}CMR-EI \\ \textit{DropOut} \\ \textit{$p_{d}=$ 0.02}\end{tabular}} & K$\tau$ & 0.87 & (**) & 0.68 & (**) & 0.77 & (**) & 0.52 & (*) \\
& S$\rho$ & 0.96 & (**) & 0.83 & (**) & 0.90 & (**) & 0.66 & (*) \\
& P$r$ & 0.95 & (**) & 0.79 & (**) & 0.94 & (**) & 0.79 & (**) \\
\cmidrule{1-10}
\multirow{3}{*}{\begin{tabular}{@{}c@{}}CMR-EI \\ \textit{DropOut} \\ \textit{$p_{d}=$ 0.03}\end{tabular}} & K$\tau$ & 0.85 & (**) & 0.68 & (**) & 0.77 & (**) & 0.61 & (**) \\
& S$\rho$ & 0.96 & (**) & 0.83 & (**) & 0.90 & (**) & 0.76 & (**) \\
& P$r$ & 0.95 & (**) & 0.79 & (**) & 0.94 & (**) & 0.78 & (**) \\
\cmidrule{1-10}
\multirow{3}{*}{\begin{tabular}{@{}c@{}}CMR-EI \\ \textit{DropOut} \\ \textit{$p_{d}=$ 0.04}\end{tabular}} & K$\tau$ & 0.83 & (**) & 0.66 & (**) & 0.79 & (**) & 0.52 & (*) \\
& S$\rho$ & 0.95 & (**) & 0.83 & (**) & 0.91 & (**) & 0.64 & (*) \\
& P$r$ & 0.94 & (**) & 0.77 & (**) & 0.91 & (**) & 0.73 & (**) \\
\cmidrule{1-10}
\multirow{3}{*}{\begin{tabular}{@{}c@{}}CMR-EI \\ \textit{DropOut} \\ \textit{$p_{d}=$ 0.05}\end{tabular}} & K$\tau$ & 0.85 & (**) & 0.66 & (**) & 0.79 & (**) & 0.61 & (**) \\
& S$\rho$ & 0.96 & (**) & 0.83 & (**) & 0.91 & (**) & 0.73 & (*) \\
& P$r$ & 0.95 & (**) & 0.78 & (**) & 0.90 & (**) & 0.71 & (**) \\
\cmidrule{1-10}
\multirow{3}{*}{\begin{tabular}{@{}c@{}}CMR-EI \\ \textit{DropOut} \\ \textit{$p_{d}=$ 0.1}\end{tabular}} & K$\tau$ & 0.75 & (**) & 0.49 & (*) & 0.57 & (**) & 0.52 & (*) \\
& S$\rho$ & 0.90 & (**) & 0.65 & (*) & 0.71 & (**) & 0.69 & (**) \\
& P$r$ & 0.87 & (**) & 0.68 & (**) & 0.66 & (**) & 0.65 & (*) \\
\bottomrule
\end{tabular}
\label{tab:CMR_EI_mito_dropout}
\end{subtable}%
\hfill
\begin{subtable}[t]{0.48\textwidth}
\centering
\tiny
\caption{CMR-NHD Correlation scores}
\setlength{\tabcolsep}{0.85pt}
\begin{tabular}{@{}c c
   >{\columncolor{GreyTable}}c
   >{\columncolor{GreyTable}}c
   c c
   >{\columncolor{GreyTable}}c
   >{\columncolor{GreyTable}}c
   c c@{}}
\toprule
\multirow{2}{*}{Metric} & {} & \multicolumn{2}{c}{EPFL} & \multicolumn{2}{c}{Hmito} & \multicolumn{2}{c}{Rmito} & \multicolumn{2}{c}{VNC} \\
&  &  \cellcolor{SecondaryColumnColor} & \cellcolor{SecondaryColumnColor}\textit{pval.} &  & \textit{pval.} &  \cellcolor{SecondaryColumnColor} & \cellcolor{SecondaryColumnColor}\textit{pval.} &  & \textit{pval.} \\
\cmidrule{1-10}
\multirow{3}{*}{\begin{tabular}{@{}c@{}}CMR-NHD \\ \textit{DropOut} \\ \textit{$p_{d}=$ 0.001}\end{tabular}} & K$\tau$ & 0.69 & (**) & 0.58 & (**) & 0.61 & (**) & 0.58 & (**) \\
& S$\rho$ & 0.89 & (**) & 0.76 & (**) & 0.80 & (**) & 0.72 & (*) \\
& P$r$ & 0.90 & (**) & 0.69 & (**) & 0.78 & (**) & 0.80 & (**) \\
\cmidrule{1-10}
\multirow{3}{*}{\begin{tabular}{@{}c@{}}CMR-NHD \\ \textit{DropOut} \\ \textit{$p_{d}=$ 0.005}\end{tabular}} & K$\tau$ & 0.71 & (**) & 0.62 & (**) & 0.63 & (**) & 0.61 & (**) \\
& S$\rho$ & 0.90 & (**) & 0.80 & (**) & 0.81 & (**) & 0.76 & (**) \\
& P$r$ & 0.90 & (**) & 0.70 & (**) & 0.81 & (**) & 0.82 & (**) \\
\cmidrule{1-10}
\multirow{3}{*}{\begin{tabular}{@{}c@{}}CMR-NHD \\ \textit{DropOut} \\ \textit{$p_{d}=$ 0.01}\end{tabular}} & K$\tau$ & 0.77 & (**) & 0.66 & (**) & 0.69 & (**) & 0.64 & (**) \\
& S$\rho$ & 0.93 & (**) & 0.82 & (**) & 0.88 & (**) & 0.76 & (*) \\
& P$r$ & 0.94 & (**) & 0.72 & (**) & 0.83 & (**) & 0.81 & (**) \\
\cmidrule{1-10}
\multirow{3}{*}{\begin{tabular}{@{}c@{}}CMR-NHD \\ \textit{DropOut} \\ \textit{$p_{d}=$ 0.02}\end{tabular}} & K$\tau$ & 0.81 & (**) & 0.66 & (**) & 0.69 & (**) & 0.61 & (*) \\
& S$\rho$ & 0.94 & (**) & 0.81 & (**) & 0.88 & (**) & 0.77 & (**) \\
& P$r$ & 0.97 & (**) & 0.73 & (**) & 0.86 & (**) & 0.81 & (**) \\
\cmidrule{1-10}
\multirow{3}{*}{\begin{tabular}{@{}c@{}}CMR-NHD \\ \textit{DropOut} \\ \textit{$p_{d}=$ 0.03}\end{tabular}} & K$\tau$ & 0.81 & (**) & 0.62 & (**) & 0.69 & (**) & 0.64 & (**) \\
& S$\rho$ & 0.94 & (**) & 0.78 & (**) & 0.88 & (**) & 0.82 & (**) \\
& P$r$ & 0.96 & (**) & 0.74 & (**) & 0.87 & (**) & 0.79 & (**) \\
\cmidrule{1-10}
\multirow{3}{*}{\begin{tabular}{@{}c@{}}CMR-NHD \\ \textit{DropOut} \\ \textit{$p_{d}=$ 0.04}\end{tabular}} & K$\tau$ & 0.81 & (**) & 0.60 & (**) & 0.67 & (**) & 0.48 & (*) \\
& S$\rho$ & 0.94 & (**) & 0.76 & (**) & 0.83 & (**) & 0.61 & (0.06) \\
& P$r$ & 0.95 & (**) & 0.73 & (**) & 0.85 & (**) & 0.73 & (**) \\
\cmidrule{1-10}
\multirow{3}{*}{\begin{tabular}{@{}c@{}}CMR-NHD \\ \textit{DropOut} \\ \textit{$p_{d}=$ 0.05}\end{tabular}} & K$\tau$ & 0.87 & (**) & 0.62 & (**) & 0.71 & (**) & 0.52 & (*) \\
& S$\rho$ & 0.96 & (**) & 0.76 & (**) & 0.84 & (**) & 0.68 & (*) \\
& P$r$ & 0.96 & (**) & 0.74 & (**) & 0.85 & (**) & 0.71 & (*) \\
\cmidrule{1-10}
\multirow{3}{*}{\begin{tabular}{@{}c@{}}CMR-NHD \\ \textit{DropOut} \\ \textit{$p_{d}=$ 0.1}\end{tabular}} & K$\tau$ & 0.73 & (**) & 0.47 & (*) & 0.52 & (**) & 0.52 & (*) \\
& S$\rho$ & 0.89 & (**) & 0.65 & (**) & 0.65 & (**) & 0.71 & (*) \\
& P$r$ & 0.88 & (**) & 0.65 & (**) & 0.64 & (*) & 0.64 & (*) \\
\bottomrule
\end{tabular}
\label{tab:CMR_NHD_mito_dropout}
\end{subtable}

\end{table}

\FloatBarrier

\subsection{Nuclei Perturbation Sweep}

For nuclei semantic segmentation we investigated ranking a model set with $|\mathcal{M}|=7$. We built the model set by training a set of specialist 2D U-Net models on seven distinct source datasets: BBBC039~\cite{ljosa_annotated_2012}, Go-Nuclear~\cite{vijayan_deep_2024}, HeLaCytoNuc~\cite{de_helacytonuc_2024}, Hoechst~\cite{arvidsson_annotated_2023}, S\_BIAD895/SB-895~\cite{von_chamier_democratising_2021}, SELMA3D 2024 challenge~\cite{leal-taixe_benchmark_2019} and S\_BIAD1410~\cite{hawkins_rescu-nets_2024}. 

We then investigated applying a range of input augmentations; additive Gaussian noise, Gamma correction, Brightness adjustment and Contrast adjustment (as defined in \cref{sec:perturbations}). We investigated additive Gaussian noise with strengths in the range $\sigma = 0.0 - 0.2$, Gamma correction with strengths in the range $\gamma = 0.8 - 1.2$, where $\gamma = 1.0$ equals no adjustment, Brightness adjustment with strengths in the range $\theta_{B}=0.0-0.2$ and Contrast adjustment with strengths in the range $\theta_{C} = 0.8 -1.2$, where $\theta_{C} = 1.0$ equals no adjustment. We show the results for both CMR-EI (\cref{tab:CMR_ei_nuclei_gauss,tab:CMR_ei_nuclei_brt,tab:CMR_ei_nuclei_ctr,tab:CMR_ei_nuclei_gamma}) and CMR-NHD (\cref{tab:CMR_NHD_nuclei_gauss,tab:CMR_nhd_nuclei_brt,tab:CMR_nhd_nuclei_ctr,tab:CMR_nhd_nuclei_gamma}).

For feature space perturbation we investigated applying spatial DropOut only to the bottle neck layer of networks, where the strength of the perturbation is controlled by the proportion of feature maps dropped at each layer $p_{d}$. We investigated DropOut proportions in the range $p_{d}= 0.05 - 0.5$. We show results for for both CMR-EI (\cref{tab:CMR_EI_nuclei_dropout}) and CMR-NHD (\cref{tab:CMR_NHD_nuclei_dropout}). Again over a wide range of perturbation strengths across all considered input and feature space perturbations the correlation scores achieved by our CMR metrics remain largely stable. Although, as discussed in the main paper, \cref{sec:model_perturbation}, care should be taken to ensure perturbations remain `tolerable' in the extreme perturbation case, too weak perturbation can lead to a loss of correlation between consistency based metrics and transfer performance. For example, the first row of \cref{tab:CMR_ei_nuclei_gauss} for the DSB2018 target dataset shows the weakest level of Gaussian input noise ($\sigma \;\text{sampled from} \; [0.0,0.05]$) can cause the CMR-NHD metric to become unstable on DSB2018.

\begin{table}[htb]
\centering
\scriptsize
\caption{Per-dataset correlation to F1-score ranking for Semantic Segmentation of Nuclei~\cite{von_chamier_democratising_2021,caicedo_nucleus_2019,ljosa_annotated_2012,arvidsson_annotated_2023} across a range of Gaussian Input Perturbation Strengths. For each ranking experiment $|\mathcal{M}|=7$. $\sigma$ is randomly sampled from the noted range. \textit{pval.} $<$ 0.05 (*), \textit{pval.} $<$ 0.01 (**).}

\begin{subtable}[t]{0.48\textwidth}
\centering
\tiny
\caption{CMR-EI Correlation scores}
\setlength{\tabcolsep}{0.7pt}
\begin{tabular}{@{}c c
   >{\columncolor{GreyTable}}c
   >{\columncolor{GreyTable}}c
   c c
   >{\columncolor{GreyTable}}c
   >{\columncolor{GreyTable}}c
   c c@{}}
\toprule
\multirow{2}{*}{Metric} & {} & \multicolumn{2}{c}{BBBC039} & \multicolumn{2}{c}{DSB2018} & \multicolumn{2}{c}{Hoechst} & \multicolumn{2}{c}{SB-895} \\
&  &  \cellcolor{SecondaryColumnColor} & \cellcolor{SecondaryColumnColor}\textit{pval.} &  & \textit{pval.} &  \cellcolor{SecondaryColumnColor} & \cellcolor{SecondaryColumnColor}\textit{pval.} &  & \textit{pval.} \\
\cmidrule{1-10}
\multirow{3}{*}{\begin{tabular}{@{}c@{}}CMR-EI \\ \textit{Gauss, $\sigma$} \\ \textit{$[0.0,0.05]$}\end{tabular}} & K$\tau$ & 0.62 & (0.08) & 0.52 & (0.12) & 0.62 & (0.06) & 0.60 & (0.19) \\
& S$\rho$ & 0.71 & (0.09) & 0.68 & (0.14) & 0.75 & (0.07) & 0.77 & (0.11) \\
& P$r$ & 0.99 & (**) & 0.98 & (**) & 0.95 & (**) & 0.98 & (**) \\
\cmidrule{1-10}
\multirow{3}{*}{\begin{tabular}{@{}c@{}}CMR-EI \\ \textit{Gauss, $\sigma$} \\ \textit{$[0.05,0.1]$}\end{tabular}} & K$\tau$ & 0.62 & (0.07) & 0.52 & (0.16) & 0.52 & (0.14) & 0.60 & (0.15) \\
& S$\rho$ & 0.79 & (*) & 0.68 & (0.11) & 0.71 & (0.10) & 0.77 & (0.10) \\
& P$r$ & 1.00 & (**) & 0.99 & (**) & 0.94 & (**) & 0.98 & (**) \\
\cmidrule{1-10}
\multirow{3}{*}{\begin{tabular}{@{}c@{}}CMR-EI \\ \textit{Gauss, $\sigma$} \\ \textit{$[0.1,0.2]$}\end{tabular}} & K$\tau$ & 0.71 & (*) & 0.52 & (0.16) & 0.52 & (0.13) & 0.73 & (0.06) \\
& S$\rho$ & 0.86 & (*) & 0.68 & (0.10) & 0.71 & (0.11) & 0.83 & (0.09) \\
& P$r$ & 1.00 & (**) & 0.99 & (**) & 0.93 & (**) & 0.99 & (**) \\
\bottomrule
\end{tabular}
\label{tab:CMR_ei_nuclei_gauss}
\end{subtable}%
\hfill
\begin{subtable}[t]{0.48\textwidth}
\centering
\tiny
\caption{CMR-NHD Correlation scores}
\setlength{\tabcolsep}{0.7pt}
\begin{tabular}{@{}c c
   >{\columncolor{GreyTable}}c
   >{\columncolor{GreyTable}}c
   c c
   >{\columncolor{GreyTable}}c
   >{\columncolor{GreyTable}}c
   c c@{}}
\toprule
\multirow{2}{*}{Metric} & {} & \multicolumn{2}{c}{BBBC039} & \multicolumn{2}{c}{DSB2018} & \multicolumn{2}{c}{Hoechst} & \multicolumn{2}{c}{SB-895} \\
&  &  \cellcolor{SecondaryColumnColor} & \cellcolor{SecondaryColumnColor}\textit{pval.} &  & \textit{pval.} &  \cellcolor{SecondaryColumnColor} & \cellcolor{SecondaryColumnColor}\textit{pval.} &  & \textit{pval.} \\
\cmidrule{1-10}
\multirow{3}{*}{\begin{tabular}{@{}c@{}}CMR-NHD \\ \textit{Gauss, $\sigma$} \\ \textit{$[0.0,0.05]$}\end{tabular}} & K$\tau$ & 0.43 & (0.21) & 0.14 & (0.77) & 0.71 & (*) & 0.73 & (0.07) \\
& S$\rho$ & 0.64 & (0.12) & 0.18 & (0.69) & 0.86 & (*) & 0.83 & (0.06) \\
& P$r$ & 1.00 & (**) & 0.98 & (**) & 0.99 & (**) & 0.99 & (**) \\
\cmidrule{1-10}
\multirow{3}{*}{\begin{tabular}{@{}c@{}}CMR-NHD \\ \textit{Gauss, $\sigma$} \\ \textit{$[0.05,0.1]$}\end{tabular}} & K$\tau$ & 0.52 & (0.13) & 0.71 & (*) & 0.71 & (*) & 0.47 & (0.27) \\
& S$\rho$ & 0.71 & (0.08) & 0.79 & (*) & 0.86 & (*) & 0.54 & (0.30) \\
& P$r$ & 1.00 & (**) & 1.00 & (**) & 0.93 & (**) & 0.98 & (**) \\
\cmidrule{1-10}
\multirow{3}{*}{\begin{tabular}{@{}c@{}}CMR-NHD \\ \textit{Gauss, $\sigma$} \\ \textit{$[0.1,0.2]$}\end{tabular}} & K$\tau$ & 0.71 & (*) & 0.81 & (*) & 0.62 & (0.07) & 0.60 & (0.14) \\
& S$\rho$ & 0.86 & (*) & 0.89 & (*) & 0.82 & (*) & 0.71 & (0.12) \\
& P$r$ & 1.00 & (**) & 1.00 & (**) & 0.92 & (**) & 0.98 & (**) \\
\bottomrule
\end{tabular}
\label{tab:CMR_NHD_nuclei_gauss}
\end{subtable}

\end{table}
\begin{table}[ht]
\centering
\scriptsize
\caption{Per-dataset correlation to F1-score ranking for Semantic Segmentation of Nuclei~\cite{von_chamier_democratising_2021,caicedo_nucleus_2019,ljosa_annotated_2012,arvidsson_annotated_2023} across a range of Brightness Input Perturbation Strengths. For each ranking experiment $|\mathcal{M}|=7$. $\theta_B$ is randomly sampled from the noted range. \textit{pval.} $<$ 0.05 (*), \textit{pval.} $<$ 0.01 (**).}

\begin{subtable}[t]{0.48\textwidth}
\centering
\tiny
\caption{CMR-EI Correlation scores}
\setlength{\tabcolsep}{0.6pt}
\begin{tabular}{@{}c c
   >{\columncolor{GreyTable}}c
   >{\columncolor{GreyTable}}c
   c c
   >{\columncolor{GreyTable}}c
   >{\columncolor{GreyTable}}c
   c c@{}}
\toprule
\multirow{2}{*}{Metric} & {} & \multicolumn{2}{c}{BBBC039} & \multicolumn{2}{c}{DSB2018} & \multicolumn{2}{c}{Hoechst} & \multicolumn{2}{c}{SB-895} \\
&  &  \cellcolor{SecondaryColumnColor} & \cellcolor{SecondaryColumnColor}\textit{pval.} &  & \textit{pval.} &  \cellcolor{SecondaryColumnColor} & \cellcolor{SecondaryColumnColor}\textit{pval.} &  & \textit{pval.} \\
\cmidrule{1-10}
\multirow{3}{*}{\begin{tabular}{@{}c@{}}CMR-EI \\ \textit{Brt, $\theta_{B}$} \\ \textit{$[0.0,0.05]$}\end{tabular}} & K$\tau$ & 0.71 & (*) & 0.43 & (0.25) & 0.52 & (0.16) & 0.60 & (0.12) \\
& S$\rho$ & 0.86 & (*) & 0.64 & (0.15) & 0.71 & (0.08) & 0.77 & (0.08) \\
& P$r$ & 0.99 & (**) & 0.91 & (**) & 0.94 & (**) & 0.96 & (**) \\
\cmidrule{1-10}
\multirow{3}{*}{\begin{tabular}{@{}c@{}}CMR-EI \\ \textit{Brt, $\theta_{B}$} \\ \textit{$[0.05,0.1]$}\end{tabular}} & K$\tau$ & 0.81 & (*) & 0.62 & (0.08) & 0.81 & (*) & 0.73 & (0.07) \\
& S$\rho$ & 0.93 & (*) & 0.75 & (0.07) & 0.93 & (**) & 0.83 & (0.05) \\
& P$r$ & 0.99 & (**) & 0.99 & (**) & 0.94 & (**) & 0.99 & (**) \\
\cmidrule{1-10}
\multirow{3}{*}{\begin{tabular}{@{}c@{}}CMR-EI \\ \textit{Brt, $\theta_{B}$} \\ \textit{$[0.1,0.2]$}\end{tabular}} & K$\tau$ & 0.71 & (*) & 0.62 & (0.06) & 0.81 & (*) & 0.73 & (0.06) \\
& S$\rho$ & 0.86 & (*) & 0.75 & (0.07) & 0.93 & (*) & 0.83 & (0.07) \\
& P$r$ & 0.99 & (**) & 0.99 & (**) & 0.94 & (**) & 0.99 & (**) \\
\bottomrule
\end{tabular}
\label{tab:CMR_ei_nuclei_brt}
\end{subtable}%
\hfill
\begin{subtable}[t]{0.48\textwidth}
\centering
\tiny
\caption{CMR-NHD Correlation scores}
\setlength{\tabcolsep}{0.6pt}
\begin{tabular}{@{}c c
   >{\columncolor{GreyTable}}c
   >{\columncolor{GreyTable}}c
   c c
   >{\columncolor{GreyTable}}c
   >{\columncolor{GreyTable}}c
   c c@{}}
\toprule
\multirow{2}{*}{Metric} & {} & \multicolumn{2}{c}{BBBC039} & \multicolumn{2}{c}{DSB2018} & \multicolumn{2}{c}{Hoechst} & \multicolumn{2}{c}{SB-895} \\
&  &  \cellcolor{SecondaryColumnColor} & \cellcolor{SecondaryColumnColor}\textit{pval.} &  & \textit{pval.} &  \cellcolor{SecondaryColumnColor} & \cellcolor{SecondaryColumnColor}\textit{pval.} &  & \textit{pval.} \\
\cmidrule{1-10}
\multirow{3}{*}{\begin{tabular}{@{}c@{}}CMR-NHD \\ \textit{Brt, $\theta_{B}$} \\ \textit{$[0.0,0.05]$}\end{tabular}} & K$\tau$ & 0.52 & (0.12) & 0.71 & (*) & 0.71 & (*) & 0.47 & (0.25) \\
& S$\rho$ & 0.71 & (0.08) & 0.79 & (0.05) & 0.86 & (*) & 0.54 & (0.29) \\
& P$r$ & 1.00 & (**) & 0.82 & (*) & 0.92 & (**) & 0.82 & (0.05) \\
\cmidrule{1-10}
\multirow{3}{*}{\begin{tabular}{@{}c@{}}CMR-NHD \\ \textit{Brt, $\theta_{B}$} \\ \textit{$[0.05,0.1]$}\end{tabular}} & K$\tau$ & 0.71 & (*) & 0.62 & (0.07) & 0.71 & (*) & 0.73 & (0.05) \\
& S$\rho$ & 0.86 & (*) & 0.71 & (0.08) & 0.86 & (*) & 0.83 & (0.06) \\
& P$r$ & 1.00 & (**) & 0.98 & (**) & 0.90 & (*) & 0.99 & (**) \\
\cmidrule{1-10}
\multirow{3}{*}{\begin{tabular}{@{}c@{}}CMR-NHD \\ \textit{Brt, $\theta_{B}$} \\ \textit{$[0.1,0.2]$}\end{tabular}} & K$\tau$ & 0.71 & (*) & 0.71 & (*) & 0.90 & (*) & 0.73 & (0.07) \\
& S$\rho$ & 0.86 & (*) & 0.82 & (*) & 0.96 & (**) & 0.83 & (0.07) \\
& P$r$ & 1.00 & (**) & 0.99 & (**) & 0.93 & (**) & 0.99 & (**) \\
\bottomrule
\end{tabular}
\label{tab:CMR_nhd_nuclei_brt}
\end{subtable}

\end{table}
\begin{table}[htb]
\centering
\scriptsize
\caption{Per-dataset correlation to F1-score ranking for Semantic Segmentation of Nuclei~\cite{von_chamier_democratising_2021,caicedo_nucleus_2019,ljosa_annotated_2012,arvidsson_annotated_2023} across a range of Contrast Input Perturbation Strengths. For each ranking experiment $|\mathcal{M}|=7$. $\theta_C$ is randomly sampled from the noted range. \textit{pval.} $<$ 0.05 (*), \textit{pval.} $<$ 0.01 (**).}

\begin{subtable}[t]{0.48\textwidth}
\centering
\tiny
\caption{CMR-EI Correlation scores}
\setlength{\tabcolsep}{0.5pt}
\begin{tabular}{@{}c c
   >{\columncolor{GreyTable}}c
   >{\columncolor{GreyTable}}c
   c c
   >{\columncolor{GreyTable}}c
   >{\columncolor{GreyTable}}c
   c c@{}}
\toprule
\multirow{2}{*}{Metric} & {} & \multicolumn{2}{c}{BBBC039} & \multicolumn{2}{c}{DSB2018} & \multicolumn{2}{c}{Hoechst} & \multicolumn{2}{c}{SB-895} \\
&  &  \cellcolor{SecondaryColumnColor} & \cellcolor{SecondaryColumnColor}\textit{pval.} &  & \textit{pval.} &  \cellcolor{SecondaryColumnColor} & \cellcolor{SecondaryColumnColor}\textit{pval.} &  & \textit{pval.} \\
\cmidrule{1-10}
\multirow{3}{*}{\begin{tabular}{@{}c@{}}CMR-EI \\ \textit{Ctr, $\theta_{C}$} \\ \textit{$[0.8,0.9]$}\end{tabular}} & K$\tau$ & 0.81 & (*) & 0.33 & (0.35) & 0.52 & (0.16) & 0.47 & (0.32) \\
& S$\rho$ & 0.93 & (*) & 0.50 & (0.26) & 0.68 & (0.12) & 0.71 & (0.11) \\
& P$r$ & 0.95 & (**) & 0.91 & (**) & 0.94 & (**) & 0.98 & (**) \\
\cmidrule{1-10}
\multirow{3}{*}{\begin{tabular}{@{}c@{}}CMR-EI \\ \textit{Ctr, $\theta_{C}$} \\ \textit{$[0.9,0.95]$}\end{tabular}} & K$\tau$ & 0.71 & (*) & 0.33 & (0.36) & 0.52 & (0.15) & 0.47 & (0.26) \\
& S$\rho$ & 0.86 & (*) & 0.50 & (0.27) & 0.68 & (0.12) & 0.71 & (0.12) \\
& P$r$ & 0.98 & (**) & 0.95 & (**) & 0.95 & (**) & 0.98 & (**) \\
\cmidrule{1-10}
\multirow{3}{*}{\begin{tabular}{@{}c@{}}CMR-EI \\ \textit{Ctr, $\theta_{C}$} \\ \textit{$[1.05,1.1]$}\end{tabular}} & K$\tau$ & 0.62 & (0.09) & 0.43 & (0.28) & 0.62 & (0.08) & 0.47 & (0.27) \\
& S$\rho$ & 0.71 & (0.10) & 0.64 & (0.13) & 0.75 & (0.09) & 0.71 & (0.13) \\
& P$r$ & 0.98 & (**) & 0.97 & (**) & 0.94 & (**) & 0.98 & (**) \\
\cmidrule{1-10}
\multirow{3}{*}{\begin{tabular}{@{}c@{}}CMR-EI \\ \textit{Ctr, $\theta_{C}$} \\ \textit{$[1.1,1.2]$}\end{tabular}} & K$\tau$ & 0.62 & (0.07) & 0.43 & (0.26) & 0.62 & (0.09) & 0.47 & (0.29) \\
& S$\rho$ & 0.71 & (0.11) & 0.64 & (0.15) & 0.75 & (0.05) & 0.71 & (0.15) \\
& P$r$ & 0.99 & (**) & 0.97 & (**) & 0.95 & (**) & 0.98 & (**) \\
\bottomrule
\end{tabular}
\label{tab:CMR_ei_nuclei_ctr}
\end{subtable}%
\hfill
\begin{subtable}[t]{0.48\textwidth}
\centering
\tiny
\caption{CMR-NHD Correlation scores}
\setlength{\tabcolsep}{0.5pt}
\begin{tabular}{@{}c c
   >{\columncolor{GreyTable}}c
   >{\columncolor{GreyTable}}c
   c c
   >{\columncolor{GreyTable}}c
   >{\columncolor{GreyTable}}c
   c c@{}}
\toprule
\multirow{2}{*}{Metric} & {} & \multicolumn{2}{c}{BBBC039} & \multicolumn{2}{c}{DSB2018} & \multicolumn{2}{c}{Hoechst} & \multicolumn{2}{c}{SB-895} \\
&  &  \cellcolor{SecondaryColumnColor} & \cellcolor{SecondaryColumnColor}\textit{pval.} &  & \textit{pval.} &  \cellcolor{SecondaryColumnColor} & \cellcolor{SecondaryColumnColor}\textit{pval.} &  & \textit{pval.} \\
\cmidrule{1-10}
\multirow{3}{*}{\begin{tabular}{@{}c@{}}CMR-NHD \\ \textit{Ctr, $\theta_{C}$} \\ \textit{$[0.8,0.9]$}\end{tabular}} & K$\tau$ & 0.81 & (*) & 0.43 & (0.23) & 0.52 & (0.12) & 0.73 & (0.05) \\
& S$\rho$ & 0.93 & (*) & 0.54 & (0.20) & 0.71 & (0.09) & 0.83 & (0.05) \\
& P$r$ & 0.94 & (**) & 0.91 & (**) & 0.92 & (**) & 0.99 & (**) \\
\cmidrule{1-10}
\multirow{3}{*}{\begin{tabular}{@{}c@{}}CMR-NHD \\ \textit{Ctr, $\theta_{C}$} \\ \textit{$[0.9,0.95]$}\end{tabular}} & K$\tau$ & 0.81 & (*) & 0.43 & (0.25) & 0.81 & (*) & 0.60 & (0.12) \\
& S$\rho$ & 0.93 & (*) & 0.54 & (0.24) & 0.93 & (*) & 0.77 & (0.09) \\
& P$r$ & 0.98 & (**) & 0.96 & (**) & 0.97 & (**) & 0.98 & (**) \\
\cmidrule{1-10}
\multirow{3}{*}{\begin{tabular}{@{}c@{}}CMR-NHD \\ \textit{Ctr, $\theta_{C}$} \\ \textit{$[1.05,1.1]$}\end{tabular}} & K$\tau$ & 0.81 & (*) & 0.43 & (0.27) & 0.71 & (*) & 0.73 & (0.07) \\
& S$\rho$ & 0.93 & (**) & 0.54 & (0.21) & 0.89 & (*) & 0.83 & (0.07) \\
& P$r$ & 0.99 & (**) & 0.97 & (**) & 0.92 & (**) & 0.99 & (**) \\
\cmidrule{1-10}
\multirow{3}{*}{\begin{tabular}{@{}c@{}}CMR-NHD \\ \textit{Ctr, $\theta_{C}$} \\ \textit{$[1.1,1.2]$}\end{tabular}} & K$\tau$ & 0.81 & (*) & 0.43 & (0.24) & 0.71 & (0.05) & 0.60 & (0.14) \\
& S$\rho$ & 0.93 & (*) & 0.64 & (0.12) & 0.89 & (*) & 0.77 & (0.08) \\
& P$r$ & 0.99 & (**) & 0.99 & (**) & 0.91 & (**) & 0.99 & (**) \\
\bottomrule
\end{tabular}
\label{tab:CMR_nhd_nuclei_ctr}
\end{subtable}

\end{table}
\begin{table}[htb]
\centering
\caption{Per-dataset correlation to F1-score ranking for Semantic Segmentation of Nuclei~\cite{von_chamier_democratising_2021,caicedo_nucleus_2019,ljosa_annotated_2012,arvidsson_annotated_2023} across a range of Gamma Input Perturbation Strengths. For each ranking experiment $|\mathcal{M}|=7$. $\gamma$ is randomly sampled from the noted range. \textit{pval.} $<$ 0.05 (*), \textit{pval.} $<$ 0.01 (**).}

\begin{subtable}[t]{0.48\textwidth}
\centering
\tiny
\caption{CMR-EI Correlation scores}
\setlength{\tabcolsep}{0.4pt}
\begin{tabular}{@{}c c
   >{\columncolor{GreyTable}}c
   >{\columncolor{GreyTable}}c
   c c
   >{\columncolor{GreyTable}}c
   >{\columncolor{GreyTable}}c
   c c@{}}
\toprule
\multirow{2}{*}{Metric} & {} & \multicolumn{2}{c}{BBBC039} & \multicolumn{2}{c}{DSB2018} & \multicolumn{2}{c}{Hoechst} & \multicolumn{2}{c}{SB-895} \\
&  &  \cellcolor{SecondaryColumnColor} & \cellcolor{SecondaryColumnColor}\textit{pval.} &  & \textit{pval.} &  \cellcolor{SecondaryColumnColor} & \cellcolor{SecondaryColumnColor}\textit{pval.} &  & \textit{pval.} \\
\cmidrule{1-10}
\multirow{3}{*}{\begin{tabular}{@{}c@{}}CMR-EI \\ \textit{Gamma, $\gamma$} \\ \textit{$[0.8,0.9]$}\end{tabular}} & K$\tau$ & 0.71 & (*) & 0.52 & (0.12) & 0.81 & (**) & 0.60 & (0.12) \\
& S$\rho$ & 0.86 & (*) & 0.68 & (0.13) & 0.89 & (*) & 0.77 & (0.09) \\
& P$r$ & 0.98 & (**) & 0.96 & (**) & 0.98 & (**) & 0.97 & (**) \\
\cmidrule{1-10}
\multirow{3}{*}{\begin{tabular}{@{}c@{}}CMR-EI \\ \textit{Gamma, $\gamma$} \\ \textit{$[0.9,0.95]$}\end{tabular}} & K$\tau$ & 0.71 & (*) & 0.52 & (0.14) & 0.71 & (*) & 0.47 & (0.26) \\
& S$\rho$ & 0.86 & (*) & 0.68 & (0.13) & 0.82 & (*) & 0.71 & (0.16) \\
& P$r$ & 0.99 & (**) & 0.96 & (**) & 0.98 & (**) & 0.96 & (**) \\
\cmidrule{1-10}
\multirow{3}{*}{\begin{tabular}{@{}c@{}}CMR-EI \\ \textit{Gamma, $\gamma$} \\ \textit{$[1.05,1.1]$}\end{tabular}} & K$\tau$ & 0.71 & (*) & 0.52 & (0.13) & 0.62 & (0.06) & 0.47 & (0.26) \\
& S$\rho$ & 0.86 & (*) & 0.68 & (0.12) & 0.75 & (0.08) & 0.71 & (0.14) \\
& P$r$ & 0.98 & (**) & 0.94 & (**) & 0.97 & (**) & 0.97 & (**) \\
\cmidrule{1-10}
\multirow{3}{*}{\begin{tabular}{@{}c@{}}CMR-EI \\ \textit{Gamma, $\gamma$} \\ \textit{$[1.1,1.2]$}\end{tabular}} & K$\tau$ & 0.81 & (*) & 0.52 & (0.16) & 0.81 & (*) & 0.73 & (0.06) \\
& S$\rho$ & 0.93 & (**) & 0.68 & (0.14) & 0.89 & (*) & 0.83 & (0.07) \\
& P$r$ & 0.96 & (**) & 0.94 & (**) & 0.98 & (**) & 0.96 & (**) \\
\bottomrule
\end{tabular}
\label{tab:CMR_ei_nuclei_gamma}
\end{subtable}%
\hfill
\begin{subtable}[t]{0.48\textwidth}
\centering
\tiny
\caption{CMR-NHD Correlation scores}
\setlength{\tabcolsep}{0.5pt}
\begin{tabular}{@{}c c
   >{\columncolor{GreyTable}}c
   >{\columncolor{GreyTable}}c
   c c
   >{\columncolor{GreyTable}}c
   >{\columncolor{GreyTable}}c
   c c@{}}
\toprule
\multirow{2}{*}{Metric} & {} & \multicolumn{2}{c}{BBBC039} & \multicolumn{2}{c}{DSB2018} & \multicolumn{2}{c}{Hoechst} & \multicolumn{2}{c}{SB-895} \\
&  &  \cellcolor{SecondaryColumnColor} & \cellcolor{SecondaryColumnColor}\textit{pval.} &  & \textit{pval.} &  \cellcolor{SecondaryColumnColor} & \cellcolor{SecondaryColumnColor}\textit{pval.} &  & \textit{pval.} \\
\cmidrule{1-10}
\multirow{3}{*}{\begin{tabular}{@{}c@{}}CMR-NHD \\ \textit{Gamma, $\gamma$} \\ \textit{$[0.8,0.9]$}\end{tabular}} & K$\tau$ & 0.71 & (*) & 0.62 & (0.05) & 0.90 & (*) & 0.87 & (*) \\
& S$\rho$ & 0.86 & (*) & 0.75 & (0.06) & 0.96 & (**) & 0.94 & (*) \\
& P$r$ & 0.97 & (**) & 0.91 & (**) & 0.99 & (**) & 0.95 & (**) \\
\cmidrule{1-10}
\multirow{3}{*}{\begin{tabular}{@{}c@{}}CMR-NHD \\ \textit{Gamma, $\gamma$} \\ \textit{$[0.9,0.95]$}\end{tabular}} & K$\tau$ & 0.62 & (0.07) & 0.52 & (0.14) & 0.81 & (*) & 0.87 & (*) \\
& S$\rho$ & 0.75 & (0.06) & 0.68 & (0.13) & 0.93 & (*) & 0.94 & (*) \\
& P$r$ & 0.97 & (**) & 0.92 & (**) & 1.00 & (**) & 0.95 & (**) \\
\cmidrule{1-10}
\multirow{3}{*}{\begin{tabular}{@{}c@{}}CMR-NHD \\ \textit{Gamma, $\gamma$} \\ \textit{$[1.05,1.1]$}\end{tabular}} & K$\tau$ & 0.62 & (0.08) & 0.43 & (0.26) & 1.00 & (**) & 0.73 & (0.05) \\
& S$\rho$ & 0.82 & (0.05) & 0.57 & (0.19) & 1.00 & (**) & 0.83 & (0.08) \\
& P$r$ & 0.93 & (**) & 0.88 & (*) & 1.00 & (**) & 0.83 & (*) \\
\cmidrule{1-10}
\multirow{3}{*}{\begin{tabular}{@{}c@{}}CMR-NHD \\ \textit{Gamma, $\gamma$} \\ \textit{$[1.1,1.2]$}\end{tabular}} & K$\tau$ & 0.71 & (*) & 0.62 & (0.06) & 0.81 & (*) & 1.00 & (**) \\
& S$\rho$ & 0.79 & (*) & 0.75 & (0.06) & 0.93 & (*) & 1.00 & (*) \\
& P$r$ & 0.84 & (*) & 0.89 & (*) & 1.00 & (**) & 0.80 & (0.06) \\
\bottomrule
\end{tabular}
\label{tab:CMR_nhd_nuclei_gamma}
\end{subtable}

\end{table}
\begin{table}[htb]
\centering
\caption{Per-dataset correlation to F1-score ranking for Semantic Segmentation of Nuclei~\cite{von_chamier_democratising_2021,caicedo_nucleus_2019,ljosa_annotated_2012,arvidsson_annotated_2023} across a range of DropOut Perturbation Strengths. DropOut was applied only to the bottleneck layer of networks. For each ranking experiment $|\mathcal{M}|=7$. \textit{pval.} $<$ 0.05 (*), \textit{pval.} $<$ 0.01 (**).}

\begin{subtable}[t]{0.48\textwidth}
\centering
\tiny
\caption{CMR-EI Correlation scores}
\setlength{\tabcolsep}{0.5pt}
\begin{tabular}{@{}c c
   >{\columncolor{GreyTable}}c
   >{\columncolor{GreyTable}}c
   c c
   >{\columncolor{GreyTable}}c
   >{\columncolor{GreyTable}}c
   c c@{}}
\toprule
\multirow{2}{*}{Metric} & {} & \multicolumn{2}{c}{BBBC039} & \multicolumn{2}{c}{DSB2018} & \multicolumn{2}{c}{Hoechst} & \multicolumn{2}{c}{SB-895} \\
&  &  \cellcolor{SecondaryColumnColor} & \cellcolor{SecondaryColumnColor}\textit{pval.} &  & \textit{pval.} &  \cellcolor{SecondaryColumnColor} & \cellcolor{SecondaryColumnColor}\textit{pval.} &  & \textit{pval.} \\
\cmidrule{1-10}
\multirow{3}{*}{\begin{tabular}{@{}c@{}}CMR-EI \\ \textit{DropOut} \\ \textit{$p_{d}=$ 0.05}\end{tabular}} & K$\tau$ & 0.71 & (*) & 0.43 & (0.24) & 0.52 & (0.18) & 0.60 & (0.13) \\
& S$\rho$ & 0.86 & (*) & 0.64 & (0.15) & 0.71 & (0.09) & 0.77 & (0.11) \\
& P$r$ & 0.99 & (**) & 0.92 & (**) & 0.90 & (*) & 0.97 & (**) \\
\cmidrule{1-10}
\multirow{3}{*}{\begin{tabular}{@{}c@{}}CMR-EI \\ \textit{DropOut} \\ \textit{$p_{d}=$ 0.1}\end{tabular}} & K$\tau$ & 0.71 & (*) & 0.43 & (0.25) & 0.52 & (0.15) & 0.60 & (0.14) \\
& S$\rho$ & 0.86 & (*) & 0.64 & (0.14) & 0.71 & (0.12) & 0.77 & (0.12) \\
& P$r$ & 0.99 & (**) & 0.86 & (*) & 0.86 & (*) & 0.96 & (**) \\
\cmidrule{1-10}
\multirow{3}{*}{\begin{tabular}{@{}c@{}}CMR-EI \\ \textit{DropOut} \\ \textit{$p_{d}=$ 0.2}\end{tabular}} & K$\tau$ & 0.71 & (*) & 0.43 & (0.24) & 0.52 & (0.12) & 0.60 & (0.12) \\
& S$\rho$ & 0.86 & (*) & 0.64 & (0.13) & 0.71 & (0.09) & 0.77 & (0.08) \\
& P$r$ & 0.99 & (**) & 0.86 & (*) & 0.85 & (*) & 0.96 & (**) \\
\cmidrule{1-10}
\multirow{3}{*}{\begin{tabular}{@{}c@{}}CMR-EI \\ \textit{DropOut} \\ \textit{$p_{d}=$ 0.3}\end{tabular}} & K$\tau$ & 0.71 & (*) & 0.52 & (0.12) & 0.52 & (0.14) & 0.60 & (0.15) \\
& S$\rho$ & 0.86 & (*) & 0.68 & (0.13) & 0.71 & (0.11) & 0.77 & (0.13) \\
& P$r$ & 0.99 & (**) & 0.79 & (*) & 0.85 & (*) & 0.96 & (**) \\
\cmidrule{1-10}
\multirow{3}{*}{\begin{tabular}{@{}c@{}}CMR-EI \\ \textit{DropOut} \\ \textit{$p_{d}=$ 0.4}\end{tabular}} & K$\tau$ & 0.81 & (*) & 0.71 & (*) & 0.52 & (0.15) & 0.73 & (0.06) \\
& S$\rho$ & 0.93 & (**) & 0.82 & (*) & 0.71 & (0.09) & 0.83 & (0.06) \\
& P$r$ & 0.98 & (**) & 0.83 & (*) & 0.80 & (*) & 0.95 & (**) \\
\cmidrule{1-10}
\multirow{3}{*}{\begin{tabular}{@{}c@{}}CMR-EI \\ \textit{DropOut} \\ \textit{$p_{d}=$ 0.5}\end{tabular}} & K$\tau$ & 0.90 & (*) & 0.81 & (*) & 0.52 & (0.13) & 0.73 & (0.05) \\
& S$\rho$ & 0.96 & (*) & 0.89 & (*) & 0.71 & (0.07) & 0.83 & (0.05) \\
& P$r$ & 0.98 & (**) & 0.77 & (*) & 0.89 & (*) & 0.95 & (**) \\
\bottomrule
\end{tabular}
\label{tab:CMR_EI_nuclei_dropout}
\end{subtable}%
\hfill
\begin{subtable}[t]{0.48\textwidth}
\centering
\tiny
\caption{CMR-NHD Correlation scores}
\setlength{\tabcolsep}{0.5pt}
\begin{tabular}{@{}c c
   >{\columncolor{GreyTable}}c
   >{\columncolor{GreyTable}}c
   c c
   >{\columncolor{GreyTable}}c
   >{\columncolor{GreyTable}}c
   c c@{}}
\toprule
\multirow{2}{*}{Metric} & {} & \multicolumn{2}{c}{BBBC039} & \multicolumn{2}{c}{DSB2018} & \multicolumn{2}{c}{Hoechst} & \multicolumn{2}{c}{SB-895} \\
&  &  \cellcolor{SecondaryColumnColor} & \cellcolor{SecondaryColumnColor}\textit{pval.} &  & \textit{pval.} &  \cellcolor{SecondaryColumnColor} & \cellcolor{SecondaryColumnColor}\textit{pval.} &  & \textit{pval.} \\
\cmidrule{1-10}
\multirow{3}{*}{\begin{tabular}{@{}c@{}}CMR-NHD \\ \textit{DropOut} \\ \textit{$p_{d}=$ 0.05}\end{tabular}} & K$\tau$ & 0.71 & (*) & 0.43 & (0.23) & 0.52 & (0.15) & 0.87 & (*) \\
& S$\rho$ & 0.86 & (*) & 0.64 & (0.16) & 0.68 & (0.11) & 0.94 & (*) \\
& P$r$ & 0.97 & (**) & 0.01 & (0.98) & 0.65 & (0.11) & 0.81 & (0.05) \\
\cmidrule{1-10}
\multirow{3}{*}{\begin{tabular}{@{}c@{}}CMR-NHD \\ \textit{DropOut} \\ \textit{$p_{d}=$ 0.1}\end{tabular}} & K$\tau$ & 0.71 & (*) & 0.71 & (0.05) & 0.62 & (0.06) & 0.73 & (0.05) \\
& S$\rho$ & 0.86 & (*) & 0.86 & (*) & 0.82 & (*) & 0.83 & (0.07) \\
& P$r$ & 0.97 & (**) & -0.03 & (0.95) & 0.61 & (0.15) & 0.68 & (0.13) \\
\cmidrule{1-10}
\multirow{3}{*}{\begin{tabular}{@{}c@{}}CMR-NHD \\ \textit{DropOut} \\ \textit{$p_{d}=$ 0.2}\end{tabular}} & K$\tau$ & 0.90 & (*) & 0.52 & (0.12) & 0.71 & (*) & 0.73 & (0.07) \\
& S$\rho$ & 0.96 & (*) & 0.68 & (0.13) & 0.89 & (*) & 0.83 & (0.05) \\
& P$r$ & 0.97 & (**) & 0.10 & (0.82) & 0.67 & (0.10) & 0.74 & (0.10) \\
\cmidrule{1-10}
\multirow{3}{*}{\begin{tabular}{@{}c@{}}CMR-NHD \\ \textit{DropOut} \\ \textit{$p_{d}=$ 0.3}\end{tabular}} & K$\tau$ & 0.90 & (*) & 0.62 & (0.07) & 0.81 & (*) & 0.73 & (0.05) \\
& S$\rho$ & 0.96 & (**) & 0.71 & (0.10) & 0.93 & (*) & 0.83 & (0.06) \\
& P$r$ & 0.97 & (**) & -0.01 & (0.99) & 0.63 & (0.13) & 0.74 & (0.09) \\
\cmidrule{1-10}
\multirow{3}{*}{\begin{tabular}{@{}c@{}}CMR-NHD \\ \textit{DropOut} \\ \textit{$p_{d}=$ 0.4}\end{tabular}} & K$\tau$ & 0.81 & (*) & 0.71 & (0.05) & 0.71 & (*) & 0.60 & (0.13) \\
& S$\rho$ & 0.89 & (*) & 0.89 & (*) & 0.89 & (*) & 0.71 & (0.14) \\
& P$r$ & 0.97 & (**) & 0.18 & (0.70) & 0.63 & (0.13) & 0.71 & (0.11) \\
\cmidrule{1-10}
\multirow{3}{*}{\begin{tabular}{@{}c@{}}CMR-NHD \\ \textit{DropOut} \\ \textit{$p_{d}=$ 0.5}\end{tabular}} & K$\tau$ & 0.81 & (*) & 0.71 & (*) & 0.52 & (0.13) & 0.73 & (0.05) \\
& S$\rho$ & 0.89 & (*) & 0.89 & (*) & 0.68 & (0.13) & 0.83 & (0.05) \\
& P$r$ & 0.96 & (**) & 0.11 & (0.81) & 0.84 & (*) & 0.68 & (0.14) \\
\bottomrule
\end{tabular}
\label{tab:CMR_NHD_nuclei_dropout}
\end{subtable}

\end{table}

\FloatBarrier

\subsection{Baseline Transferability Metrics}

We also show the correlation scores per dataset for all the baseline transferability metrics. \cref{tab:transfer_metric_mitochondria} shows the results for the four mitochondria datasets (EPFL~\cite{lucchi_learning_2013}, Hmito~\cite{franco-barranco_current_2023}, Rmito~\cite{franco-barranco_current_2023} and VNC~\cite{phelps_reconstruction_2021}) and \cref{tab:transfer_metric_nuclei} shows the results for the four nuclei datasets (BBBC039~\cite{ljosa_annotated_2012}, DSB2018~\cite{caicedo_nucleus_2019}, Hoechst~\cite{arvidsson_annotated_2023}, S\_BIAD895~\cite{von_chamier_democratising_2021}). In the main paper due to space constraints we showed the correlation scores averaged within the dataset groups.
\begin{table}[h]
\centering
\tiny
\caption{Per-dataset baseline correlations to F1-score ranking for Semantic Segmentation of Mitochondria. For each ranking experiment $|\mathcal{M}|=15$. \textit{pval.} $<$ 0.05 (*), \textit{pval.} $<$ 0.01 (**).}
\setlength{\tabcolsep}{3pt}
\begin{tabular}{@{}c c
   >{\columncolor{GreyTable}}c
   >{\columncolor{GreyTable}}c
   c c
   >{\columncolor{GreyTable}}c
   >{\columncolor{GreyTable}}c
   c c@{}}
\toprule
\multirow{2}{*}{Metric} & {} & \multicolumn{2}{c}{EPFL} & \multicolumn{2}{c}{Hmito} & \multicolumn{2}{c}{Rmito} & \multicolumn{2}{c}{VNC} \\
&  &  \cellcolor{SecondaryColumnColor} & \cellcolor{SecondaryColumnColor}\textit{pval.} &  & \textit{pval.} &  \cellcolor{SecondaryColumnColor} & \cellcolor{SecondaryColumnColor}\textit{pval.} &  & \textit{pval.} \\
\cmidrule{1-10}
\multirow{3}{*}{CCFV} & K$\tau$ & 0.03 & (0.89) & -0.09 & (0.70) & -0.21 & (0.30) & -0.18 & (0.45) \\
& S$\rho$ & 0.02 & (0.96) & -0.15 & (0.56) & -0.29 & (0.30) & -0.22 & (0.51) \\
& P$r$ & -0.03 & (0.92) & -0.17 & (0.54) & -0.23 & (0.40) & -0.29 & (0.36) \\
\cmidrule{1-10}
\multirow{3}{*}{GBC} & K$\tau$ & 0.45 & (*) & 0.43 & (*) & 0.34 & (0.09) & 0.33 & (0.15) \\
& S$\rho$ & 0.60 & (*) & 0.54 & (*) & 0.51 & (0.06) & 0.41 & (0.17) \\
& P$r$ & 0.65 & (*) & 0.54 & (*) & 0.47 & (0.08) & 0.23 & (0.48) \\
\cmidrule{1-10}
\multirow{3}{*}{LEEP} & K$\tau$ & 0.79 & (**) & 0.90 & (**) & 0.94 & (**) & 0.91 & (**) \\
& S$\rho$ & 0.89 & (**) & 0.98 & (**) & 0.98 & (**) & 0.98 & (**) \\
& P$r$ & 0.96 & (**) & 0.96 & (**) & 0.98 & (**) & 0.98 & (**) \\
\cmidrule{1-10}
\multirow{3}{*}{NLEEP} & K$\tau$ & 0.26 & (0.21) & 0.77 & (**) & 0.65 & (**) & -0.06 & (0.86) \\
& S$\rho$ & 0.32 & (0.23) & 0.90 & (**) & 0.82 & (**) & -0.06 & (0.92) \\
& P$r$ & 0.26 & (0.34) & 0.66 & (*) & 0.63 & (*) & -0.01 & (0.97) \\
\cmidrule{1-10}
\multirow{3}{*}{Hscore} & K$\tau$ & -0.12 & (0.58) & 0.33 & (0.10) & 0.36 & (0.06) & 0.36 & (0.11) \\
& S$\rho$ & -0.21 & (0.46) & 0.50 & (0.05) & 0.53 & (*) & 0.50 & (0.09) \\
& P$r$ & 0.00 & (1.00) & 0.49 & (0.07) & 0.46 & (0.08) & 0.54 & (0.07) \\
\cmidrule{1-10}
\multirow{3}{*}{RegHscore} & K$\tau$ & -0.12 & (0.52) & 0.31 & (0.12) & 0.40 & (*) & 0.36 & (0.10) \\
& S$\rho$ & -0.21 & (0.44) & 0.49 & (0.06) & 0.58 & (*) & 0.50 & (0.11) \\
& P$r$ & 0.00 & (1.00) & 0.48 & (0.07) & 0.46 & (0.08) & 0.54 & (0.07) \\
\cmidrule{1-10}
\multirow{3}{*}{LogME} & K$\tau$ & -0.24 & (0.23) & 0.10 & (0.61) & 0.25 & (0.22) & 0.21 & (0.39) \\
& S$\rho$ & -0.33 & (0.23) & 0.17 & (0.56) & 0.32 & (0.24) & 0.25 & (0.43) \\
& P$r$ & -0.30 & (0.27) & 0.05 & (0.85) & 0.09 & (0.75) & 0.41 & (0.18) \\
\cmidrule{1-10}
\multirow{3}{*}{NCTI} & K$\tau$ & -0.05 & (0.85) & 0.30 & (0.14) & 0.27 & (0.16) & 0.33 & (0.13) \\
& S$\rho$ & -0.07 & (0.75) & 0.42 & (0.12) & 0.35 & (0.23) & 0.44 & (0.17) \\
& P$r$ & 0.07 & (0.81) & 0.42 & (0.12) & 0.42 & (0.12) & 0.58 & (0.05) \\
\cmidrule{1-10}
\multirow{3}{*}{TS} & K$\tau$ & 0.54 & (*) & 0.28 & (0.14) & 0.08 & (0.77) & 0.09 & (0.74) \\
& S$\rho$ & 0.72 & (*) & 0.33 & (0.23) & 0.02 & (0.94) & 0.11 & (0.70) \\
& P$r$ & 0.60 & (*) & 0.19 & (0.50) & 0.14 & (0.62) & 0.00 & (0.99) \\
\cmidrule{1-10}
\multirow{3}{*}{NuNo} & K$\tau$ & 0.49 & (*) & 0.2 & (0.28) & 0.06 & (0.85) & -0.06 & (0.87) \\
& S$\rho$ & 0.68 & (*) & 0.20 & (0.51) & -0.03 & (0.92) & -0.04 & (0.90) \\
& P$r$ & 0.52 & (*) & 0.10 & (0.72) & -0.05 & (0.86) & -0.04 & (0.90) \\
\cmidrule{1-10}
\multirow{3}{*}{Dispersion} & K$\tau$ & 0.03 & (0.94) & -0.07 & (0.78) & -0.06 & (0.73) & 0.00 & (1.00) \\
& S$\rho$ & 0.05 & (0.87) & -0.13 & (0.66) & -0.12 & (0.66) & -0.08 & (0.81) \\
& P$r$ & -0.07 & (0.81) & -0.25 & (0.36) & -0.20 & (0.49) & -0.17 & (0.60) \\
\bottomrule
\end{tabular}
\label{tab:transfer_metric_mitochondria}
\end{table}
\begin{table}[tb]
\centering
\tiny
\caption{Per-dataset baseline correlations to F1-score ranking for Semantic Segmentation of Nuclei~\cite{von_chamier_democratising_2021,caicedo_nucleus_2019,ljosa_annotated_2012,arvidsson_annotated_2023}. For each ranking experiment $|\mathcal{M}|=12$. \textit{pval.} $<$ 0.05 (*), \textit{pval.} $<$ 0.01 (**).}
\setlength{\tabcolsep}{3pt}
\begin{tabular}{@{}c c
   >{\columncolor{GreyTable}}c
   >{\columncolor{GreyTable}}c
   c c
   >{\columncolor{GreyTable}}c
   >{\columncolor{GreyTable}}c
   c c@{}}
\toprule
\multirow{2}{*}{Metric} & {} & \multicolumn{2}{c}{BBBC039} & \multicolumn{2}{c}{DSB2018} & \multicolumn{2}{c}{Hoechst} & \multicolumn{2}{c}{S\_BIAD895} \\
&  &  \cellcolor{SecondaryColumnColor} & \cellcolor{SecondaryColumnColor}\textit{pval.} &  & \textit{pval.} &  \cellcolor{SecondaryColumnColor} & \cellcolor{SecondaryColumnColor}\textit{pval.} &  & \textit{pval.} \\
\cmidrule{1-10}
\multirow{3}{*}{CCFV} & K$\tau$ & 0.05 & (1.00) & 0.05 & (1.00) & 0.24 & (0.54) & -0.47 & (0.29) \\
& S$\rho$ & 0.14 & (0.77) & 0.04 & (0.98) & 0.21 & (0.61) & -0.60 & (0.25) \\
& P$r$ & -0.31 & (0.51) & -0.01 & (0.99) & 0.45 & (0.31) & -0.78 & (0.07) \\
\cmidrule{1-10}
\multirow{3}{*}{GBC} & K$\tau$ & -0.14 & (0.80) & -0.14 & (0.82) & 0.24 & (0.54) & -0.47 & (0.29) \\
& S$\rho$ & -0.21 & (0.67) & -0.21 & (0.65) & 0.36 & (0.45) & -0.60 & (0.21) \\
& P$r$ & -0.16 & (0.72) & 0.11 & (0.81) & 0.32 & (0.48) & -0.59 & (0.22) \\
\cmidrule{1-10}
\multirow{3}{*}{LEEP} & K$\tau$ & 0.81 & (*) & 0.71 & (*) & 0.14 & (0.77) & 1.00 & (**) \\
& S$\rho$ & 0.89 & (*) & 0.86 & (*) & 0.32 & (0.51) & 1.00 & (*) \\
& P$r$ & 0.95 & (**) & 0.92 & (**) & 0.94 & (**) & 0.94 & (*) \\
\cmidrule{1-10}
\multirow{3}{*}{NLEEP} & K$\tau$ & 0.14 & (0.77) & -0.05 & (1.00) & 0.14 & (0.77) & 0.20 & (0.79) \\
& S$\rho$ & 0.18 & (0.72) & -0.04 & (1.00) & 0.14 & (0.76) & 0.20 & (0.70) \\
& P$r$ & 0.62 & (0.14) & 0.30 & (0.52) & 0.09 & (0.85) & 0.26 & (0.62) \\
\cmidrule{1-10}
\multirow{3}{*}{Hscore} & K$\tau$ & 0.33 & (0.33) & -0.14 & (0.78) & -0.05 & (1.00) & 1.00 & (*) \\
& S$\rho$ & 0.36 & (0.46) & -0.36 & (0.45) & 0.00 & (1.00) & 1.00 & (*) \\
& P$r$ & 0.36 & (0.43) & 0.13 & (0.77) & -0.04 & (0.94) & 0.78 & (0.07) \\
\cmidrule{1-10}
\multirow{3}{*}{RegHscore} & K$\tau$ & 0.33 & (0.41) & 0.05 & (0.97) & 0.14 & (0.72) & 0.87 & (*) \\
& S$\rho$ & 0.36 & (0.45) & -0.07 & (0.92) & 0.21 & (0.66) & 0.94 & (*) \\
& P$r$ & 0.29 & (0.52) & 0.18 & (0.70) & 0.07 & (0.88) & 0.84 & (*) \\
\cmidrule{1-10}
\multirow{3}{*}{LogME} & K$\tau$ & 0.14 & (0.82) & -0.14 & (0.78) & -0.33 & (0.41) & -0.47 & (0.25) \\
& S$\rho$ & 0.18 & (0.67) & -0.21 & (0.68) & -0.50 & (0.24) & -0.60 & (0.23) \\
& P$r$ & -0.36 & (0.43) & -0.54 & (0.21) & -0.12 & (0.80) & -0.94 & (*) \\
\cmidrule{1-10}
\multirow{3}{*}{NCTI} & K$\tau$ & 0.33 & (0.39) & -0.14 & (0.76) & -0.05 & (1.00) & 1.00 & (**) \\
& S$\rho$ & 0.36 & (0.45) & -0.36 & (0.45) & 0.00 & (1.00) & 1.00 & (*) \\
& P$r$ & 0.37 & (0.41) & 0.14 & (0.76) & -0.02 & (0.96) & 0.72 & (0.10) \\
\cmidrule{1-10}
\multirow{3}{*}{TS} & K$\tau$ & 0.14 & (0.84) & -0.33 & (0.41) & -0.05 & (1.00) & 0.33 & (0.45) \\
& S$\rho$ & 0.25 & (0.64) & -0.43 & (0.35) & 0.11 & (0.77) & 0.49 & (0.35) \\
& P$r$ & -0.02 & (0.97) & -0.56 & (0.20) & -0.13 & (0.79) & 0.37 & (0.48) \\
\cmidrule{1-10}
\multirow{3}{*}{NuNo} & K$\tau$ & 0.33 & (0.36) & -0.43 & (0.23) & -0.14 & (0.82) & 0.60 & (0.13) \\
& S$\rho$ & 0.57 & (0.16) & -0.43 & (0.37) & -0.11 & (0.85) & 0.66 & (0.19) \\
& P$r$ & 0.15 & (0.75) & -0.48 & (0.28) & -0.04 & (0.93) & 0.68 & (0.14) \\
\cmidrule{1-10}
\multirow{3}{*}{Dispersion} & K$\tau$ & 0.24 & (0.54) & -0.24 & (0.56) & -0.24 & (0.53) & -0.47 & (0.29) \\
& S$\rho$ & 0.25 & (0.58) & -0.43 & (0.36) & -0.39 & (0.44) & -0.60 & (0.26) \\
& P$r$ & 0.43 & (0.33) & -0.02 & (0.97) & -0.62 & (0.14) & -0.56 & (0.25) \\
\bottomrule
\end{tabular}
\label{tab:transfer_metric_nuclei}
\end{table}
\FloatBarrier

\subsection{Multiclass Toothfairy2 Perturbation Sweep}

For multiclass semantic segmentation we built a diverse set of models with $|\mathcal{M}|=8$. All but one of the models were trained on the Toothfairy2 datasets taken from the MICCAI2024 challenge~\cite{2024TMI,2025CVPR}, the exception being the generalist TotalSegmentator~\cite{wasserthal_totalsegmentator_2023} model. The model set comprised of a wide range of architectures: 2D nnU-Net, residual-encoder 3D nnU-Net, SwinUNETR, TotalSegmentator, UMamba and VMamba.

We investigated applying both additive Gaussian noise (controlled by $\sigma$) and Gamma Correction (controlled by $\gamma$) as input perturbations. For feature space perturbation we investigated applying spatial DropOut to the bottleneck and skip connection layers of the network, where the strength of the perturbation is controlled by the proportion of feature maps dropped at each layer $p_{d}$. We report the performance of both CMR-EI (\cref{tab:CMR_ei_toothfairy}) and CMR-NHD (\cref{tab:CMR_nhd_toothfairy}).

\begin{table}[ht]
\centering
\tiny
\caption{CMR-EI Correlation scores for multiclass Semantic Segmentation of Toothfairy2 human jaw dataset across a range of Perturbations. $|\mathcal{M}|=8$. \textit{pval.} $<$ 0.05 (*), \textit{pval.} $<$ 0.01 (**).}
\setlength{\tabcolsep}{3pt}
\begin{tabular}{@{}c c
   >{\columncolor{GreyTable}}c
   >{\columncolor{GreyTable}}c
   c c
   >{\columncolor{GreyTable}}c
   >{\columncolor{GreyTable}}c
   c c
   >{\columncolor{GreyTable}}c
   >{\columncolor{GreyTable}}c@{}}
\toprule
\multirow{2}{*}{Metric} & {} & \multicolumn{2}{c}{IACs} & \multicolumn{2}{c}{Teeth} & \multicolumn{2}{c}{Mand.} & \multicolumn{2}{c}{Sinus.} & \multicolumn{2}{c}{Overall} \\
&  &  \cellcolor{SecondaryColumnColor} & \cellcolor{SecondaryColumnColor}\textit{pval.} &  & \textit{pval.} &  \cellcolor{SecondaryColumnColor} & \cellcolor{SecondaryColumnColor}\textit{pval.} &  & \textit{pval.} &  \cellcolor{SecondaryColumnColor} & \cellcolor{SecondaryColumnColor}\textit{pval.}\\
\cmidrule{1-12}
\multirow{3}{*}{\begin{tabular}{@{}c@{}}CMR-EI \\ \textit{Gauss} \\ \textit{($\sigma=$ 2.0x)}\end{tabular}} & K$\tau$ & 0.90 & (**) & 0.81 & (**) & 0.62 & (*) & 0.43 & (0.3) & 0.71 & (*)\\
& S$\rho$ & 0.96 & (**) & 0.89 & (*) & 0.82 & (*) & 0.50 & (0.3) & 0.86 & (*)\\
& P$r$ & 0.99 & (**) & 0.95 & (**) & 0.73 & (0.1) & 0.38 & (0.4) & 0.93 & (**)\\
\cmidrule{1-12}
\multirow{3}{*}{\begin{tabular}{@{}c@{}}CMR-EI \\ \textit{Gauss} \\ \textit{($\sigma=$ 2.5x)}\end{tabular}} & K$\tau$ & 0.93 & (**) & 0.86 & (*) & 0.65 & (*) & 0.40 & (0.2) & 0.71 & (*)\\
& S$\rho$ & 0.97 & (**) & 0.85 & (0.1) & 0.85 & (**) & 0.44 & (0.1) & 0.84 & (*)\\
& P$r$ & 0.96 & (*) & 0.94 & (**) & 0.73 & (*) & 0.39 & (0.3) & 0.90 & (*)\\
\cmidrule{1-12}
\multirow{3}{*}{\begin{tabular}{@{}c@{}}CMR-EI \\ \textit{Gauss} \\ \textit{($\sigma=$ 3.0x)}\end{tabular}} & K$\tau$ & 0.90 & (**) & 0.84 & (**) & 0.65 & (*) & 0.43 & (0.2) & 0.71 & (*)\\
& S$\rho$ & 0.97 & (**) & 0.89 & (**) & 0.85 & (**) & 0.44 & (0.2) & 0.86 & (*)\\
& P$r$ & 0.96 & (*) & 0.92 & (**) & 0.59 & (0.2) & 0.42 & (0.2) & 0.90 & (*)\\
\cmidrule{1-12}
\multirow{3}{*}{\begin{tabular}{@{}c@{}}CMR-EI \\ \textit{Gamma} \\ \textit{($\gamma=$ 0.2)}\end{tabular}} & K$\tau$ & 0.93 & (**) & 0.77 & (*) & 0.65 & (0.1) & 0.57 & (0.1) & 0.78 & (*)\\
& S$\rho$ & 0.96 & (**) & 0.88 & (**) & 0.74 & (0.2) & 0.61 & (0.1) & 0.88 & (0.1)\\
& P$r$ & 0.95 & (*) & 0.71 & (*) & 0.66 & (0.1) & 0.50 & (0.1) & 0.70 & (0.2)\\
\cmidrule{1-12}
\multirow{3}{*}{\begin{tabular}{@{}c@{}}CMR-EI \\ \textit{Gamma} \\ \textit{($\gamma=$ 0.5)}\end{tabular}} & K$\tau$ & 0.97 & (**) & 0.77 & (**) & 0.68 & (0.1) & 0.48 & (0.2) & 0.81 & (**)\\
& S$\rho$ & 0.98 & (**) & 0.90 & (**) & 0.80 & (*) & 0.51 & (0.2) & 0.89 & (**)\\
& P$r$ & 0.94 & (*) & 0.83 & (**) & 0.75 & (*) & 0.42 & (0.3) & 0.89 & (**)\\
\cmidrule{1-12}
\multirow{3}{*}{\begin{tabular}{@{}c@{}}CMR-EI \\ \textit{Gamma} \\ \textit{($\gamma=$ 2.0)}\end{tabular}} & K$\tau$ & 1.00 & (**) & 0.71 & (*) & 0.43 & (0.3) & 0.43 & (0.2) & 0.81 & (**)\\
& S$\rho$ & 1.00 & (**) & 0.86 & (*) & 0.54 & (0.2) & 0.57 & (0.2) & 0.89 & (*)\\
& P$r$ & 0.99 & (**) & 0.89 & (**) & 0.74 & (0.1) & 0.38 & (0.4) & 0.90 & (**)\\
\cmidrule{1-12}
\multirow{3}{*}{\begin{tabular}{@{}c@{}}CMR-EI \\ \textit{DropOut} \\ \textit{($p_{d}=$ 0.3)}\end{tabular}} &  K$\tau$ & 1.00 & (**) & 0.60 & (0.2) & 0.83 & (**) & 0.44 & (0.2) & 0.58 & (0.2)\\
& S$\rho$ & 1.00 & (**) & 0.77 & (0.1) & 0.82 & (*) & 0.52 & (0.2) & 0.72 & (0.1)\\
& P$r$ & 0.95 & (**) & 0.94 & (*) & 0.87 & (*) & 0.49 & (0.1) & 0.86 & (*)\\
\cmidrule{1-12}
\multirow{3}{*}{\begin{tabular}{@{}c@{}}CMR-EI \\ \textit{DropOut} \\ \textit{($p_{d}=$ 0.5)}\end{tabular}} & K$\tau$ & 0.87 & (*) & 0.67 & (0.1) & 0.87 & (*) & 0.87 & (*) & 0.93 & (*)\\
& S$\rho$ & 0.94 & (**) & 0.77 & (0.1) & 0.87 & (*) & 0.84 & (*) & 0.95 & (*)\\
& P$r$ & 0.83 & (*) & 0.85 & (**) & 0.89 & (**) & 0.88 & (**) & 0.93 & (**)\\
\cmidrule{1-12}
\multirow{3}{*}{\begin{tabular}{@{}c@{}}CMR-EI \\ \textit{DropOut} \\ \textit{($p_{d}=$ 0.7)}\end{tabular}} & K$\tau$ & 0.65 & (*) & 0.58 & (0.2) & 0.68 & (0.1) & 0.59 & (0.1) & 0.60 & (0.2)\\
& S$\rho$ & 0.79 & (*) & 0.61 & (0.2) & 0.75 & (0.1) & 0.58 & (0.2) & 0.57 & (0.4)\\
& P$r$ & 0.70 & (*) & 0.82 & (*) & 0.86 & (0.1) & 0.55 & (0.2) & 0.79 & (**)\\
\bottomrule
\end{tabular}
\label{tab:CMR_ei_toothfairy}
\end{table}
\begin{table}[tb]
\centering
\tiny
\caption{CMR-NHD Correlation scores for multiclass Semantic Segmentation of Toothfairy2 human jaw dataset across a range of Perturbations. $|\mathcal{M}|=8$. \textit{pval.} $<$ 0.05 (*), \textit{pval.} $<$ 0.01 (**).}
\setlength{\tabcolsep}{3pt}
\begin{tabular}{@{}c c
   >{\columncolor{GreyTable}}c
   >{\columncolor{GreyTable}}c
   c c
   >{\columncolor{GreyTable}}c
   >{\columncolor{GreyTable}}c
   c c
   >{\columncolor{GreyTable}}c
   >{\columncolor{GreyTable}}c@{}}
\toprule
\multirow{2}{*}{Metric} & {} & \multicolumn{2}{c}{IACs} & \multicolumn{2}{c}{Teeth} & \multicolumn{2}{c}{Mand.} & \multicolumn{2}{c}{Sinus.} & \multicolumn{2}{c}{Overall} \\
&  &  \cellcolor{SecondaryColumnColor} & \cellcolor{SecondaryColumnColor}\textit{pval.} &  & \textit{pval.} &  \cellcolor{SecondaryColumnColor} & \cellcolor{SecondaryColumnColor}\textit{pval.} &  & \textit{pval.} &  \cellcolor{SecondaryColumnColor} & \cellcolor{SecondaryColumnColor}\textit{pval.}\\
\cmidrule{1-12}
\multirow{3}{*}{\begin{tabular}{@{}c@{}}CMR-NHD \\ \textit{Gauss} \\ \textit{($\sigma=$ 2.0x)}\end{tabular}} & K$\tau$ & 0.81 & (**) & 1.00 & (**) & 0.52 & (0.2) & 0.81 &(*) & 0.90 & (**)\\
& S$\rho$ & 0.93 & (*) & 1.00 & (**) & 0.68 & (0.1) & 0.89 & (*) & 0.96 & (**)\\
& P$r$ & 0.98 & (**) & 0.98 & (**) & 0.65 & (0.1) & 0.73 & (0.1) & 0.98 & (**) \\
\cmidrule{1-12}
\multirow{3}{*}{\begin{tabular}{@{}c@{}}CMR-NHD \\ \textit{Gauss} \\ \textit{($\sigma=$ 2.5x)}\end{tabular}} & K$\tau$ & 0.81 & (**) & 1.00 & (**) & 0.43 & (0.2) & 0.81 & (**) & 0.90 & (**)\\
& S$\rho$ & 0.93 & (**) & 1.00 & (**) & 0.57 & (0.2) & 0.89 & (*) & 0.96 & (**)\\
& P$r$ & 0.99 & (**) & 0.96 & (**) & 0.60 & (0.2) & 0.67 & (**) & 0.96 & (**) \\
\cmidrule{1-12}
\multirow{3}{*}{\begin{tabular}{@{}c@{}}CMR-NHD \\ \textit{Gauss} \\ \textit{($\sigma=$ 3.0x)}\end{tabular}} & K$\tau$ & 0.81 & (**) & 1.00 & (**) & 0.52 & (0.2) & 0.81 & (**) & 0.90 & (**)\\
& S$\rho$ & 0.93 & (**) & 1.00 & (**) & 0.68 & (0.1) & 0.89 & (*) & 0.96 & (**)\\
& P$r$ & 0.99 & (**) & 0.94 & (**) & 0.50 & (0.3) & 0.64 & (0.1) & 0.94 & (**) \\
\cmidrule{1-12}
\multirow{3}{*}{\begin{tabular}{@{}c@{}}CMR-NHD \\ \textit{Gamma} \\ \textit{($\gamma=$ 0.2)}\end{tabular}} & K$\tau$ & 0.81 & (*) & 0.71 & (*) & 0.43 & (0.3) & 0.62 & (0.1) & 0.81 & (**)\\
& S$\rho$ & 0.93 & (**) & 0.86 & (*) & 0.49 & (0.1) & 0.75 & (0.1) & 0.89 & (**)\\
& P$r$ & 0.99 & (**) & 0.67 & (0.1) & 0.42 & (0.2) & 0.64 & (0.1) & 0.72 & (0.1) \\
\cmidrule{1-12}
\multirow{3}{*}{\begin{tabular}{@{}c@{}}CMR-NHD \\ \textit{Gamma} \\ \textit{($\gamma=$ 0.5)}\end{tabular}} & K$\tau$ & 0.90 & (**) & 0.81 & (*) & 0.90 & (**) & 0.81 & (*) & 0.81 & (**)\\
& S$\rho$ & 0.96 & (**) & 0.93 & (**) & 0.96 & (**) & 0.89 & (**) & 0.89 & (**)\\
& P$r$ & 0.96 & (**) & 0.94 & (**) & 0.79 & (*) & 0.97 & (**) & 0.94 & (**) \\
\cmidrule{1-12}
\multirow{3}{*}{\begin{tabular}{@{}c@{}}CMR-NHD \\ \textit{Gamma} \\ \textit{($\gamma=$ 2.0)}\end{tabular}} & K$\tau$ & 0.90 & (*) & 0.90 & (**) & 0.81 & (**) & 0.90 & (**) & 1.00 & (**)\\
& S$\rho$ & 0.96 & (**) & 0.96 & (**) & 0.89 & (**) & 0.96 & (**) & 1.00 & (**)\\
& P$r$ & 0.98 & (**) & 0.95 & (**) & 0.80 & (*) & 0.98 & (**) & 0.96 & (**) \\
\cmidrule{1-12}
\multirow{3}{*}{\begin{tabular}{@{}c@{}}CMR-NHD \\ \textit{DropOut} \\ \textit{($p_{d}=$ 0.3)}\end{tabular}} & K$\tau$ & 0.90 & (**) & 0.71 & (*) & 0.81 & (**) & 0.43 & (0.2) & 0.81 & (**) \\
& S$\rho$ & 0.96 & (**) & 0.86 & (*) & 0.89 & (*) & 0.43 & (0.4) & 0.89 & (**) \\
& P$r$ & 0.86 & (**) & 0.95 & (**) & 0.83 & (*) & 0.77 & (*) & 0.95 & (**)\\
\cmidrule{1-12}
\multirow{3}{*}{\begin{tabular}{@{}c@{}}CMR-NHD \\ \textit{DropOut} \\ \textit{($p_{d}=$ 0.5)}\end{tabular}} & K$\tau$ & 0.87 & (*) & 0.75 & (0.1) & 0.87 & (**) & 0.83 & (*) & 0.73 & (0.1) \\
& S$\rho$ & 0.94 & (**) & 0.77 & (0.1) & 0.94 & (*) & 0.94 & (*) & 0.83 & (0.1) \\
& P$r$ & 0.80 & (0.1) & 0.94 & (**) & 0.96 & (**) & 0.90 & (**) & 0.93 & (**)\\
\cmidrule{1-12}
\multirow{3}{*}{\begin{tabular}{@{}c@{}}CMR-NHD \\ \textit{DropOut} \\ \textit{($p_{d}=$ 0.7)}\end{tabular}} & K$\tau$ & 0.62 & (0.1) & 0.43 & (0.2) & 0.81 & (*) & 0.44 & (0.2) & 0.53 & (0.3) \\
& S$\rho$ & 0.71 & (0.1) & 0.50 & (0.3) & 0.93 & (**) & 0.43 & (0.3) & 0.49 & (0.2) \\
& P$r$ & 0.69 & (0.1) & 0.85 & (*) & 0.86 & (**) & 0.57 & (0.2) & 0.71 & (*)\\
\bottomrule
\end{tabular}
\label{tab:CMR_nhd_toothfairy}
\end{table}

\FloatBarrier

\subsection{Post-UDA Perturbation Sweep}

In this section we investigate the performance of the CMR metric for predicting the model ttarget performance ranking after the application of two unsupervised domain adaptation (UDA) approaches, Mean Teacher~\cite{tarvainen_mean_2018} (\cref{tab:CMR_ei_mitochondria_UDA_gauss}) and adaptive batch normalisation (AdaBN)~\cite{li_adaptive_2018} (\cref{tab:CMR_ei_mitochondria_UDA_gauss_adabn}). The tables show a sweep over a range of perturbation strengths and report the correlation scores per target dataset. In the main paper due to space limitations we reported the correlation scores averaged over the target datasets for a single perturbation strength. We investigated applying additive Gaussian noise to the input of the models with strengths varying in the range $\sigma = 0.01 - 0.2$. 

\begin{table}[htb]
\centering
\caption{Per-dataset post-UDA CMR-EI Correlation scores to F1-score ranking for Semantic Segmentation of Mitochondria across a range of Gaussian Input Perturbation Strengths. For each ranking experiment $|\mathcal{M}|=12$. $\sigma$ is randomly sampled from the noted range. \textit{pval.} $<$ 0.05 (*), \textit{pval.} $<$ 0.01 (**).}

\begin{subtable}[t]{0.48\textwidth}
\centering
\tiny
\caption{Mean Teacher}
\setlength{\tabcolsep}{0.7pt}
\begin{tabular}{@{}c c
   >{\columncolor{GreyTable}}c
   >{\columncolor{GreyTable}}c
   c c
   >{\columncolor{GreyTable}}c
   >{\columncolor{GreyTable}}c
   c c@{}}
\toprule
\multirow{2}{*}{Metric} & {} & \multicolumn{2}{c}{EPFL} & \multicolumn{2}{c}{Hmito} & \multicolumn{2}{c}{Rmito} & \multicolumn{2}{c}{VNC} \\
&  &  \cellcolor{SecondaryColumnColor} & \cellcolor{SecondaryColumnColor}\textit{pval.} &  & \textit{pval.} &  \cellcolor{SecondaryColumnColor} & \cellcolor{SecondaryColumnColor}\textit{pval.} &  & \textit{pval.} \\
\cmidrule{1-10}
\multirow{3}{*}{\begin{tabular}{@{}c@{}}CMR-EI \\ \textit{Gauss, $\sigma$} \\ \textit{$[0.01,0.03]$}\end{tabular}} & K$\tau$ & 0.71 & (**) & 0.64 & (**) & 0.56 & (*) & 0.15 & (0.57) \\
& S$\rho$ & 0.85 & (**) & 0.78 & (*) & 0.75 & (*) & 0.22 & (0.53) \\
& P$r$ & 0.93 & (**) & 0.71 & (*) & 0.52 & (0.10) & 0.11 & (0.73) \\
\cmidrule{1-10}
\multirow{3}{*}{\begin{tabular}{@{}c@{}}CMR-EI \\ \textit{Gauss, $\sigma$} \\ \textit{$[0.03,0.05]$}\end{tabular}} & K$\tau$ & 0.82 & (**) & 0.67 & (**) & 0.60 & (*) & 0.09 & (0.71) \\
& S$\rho$ & 0.90 & (**) & 0.79 & (**) & 0.78 & (*) & 0.15 & (0.65) \\
& P$r$ & 0.95 & (**) & 0.75 & (**) & 0.59 & (0.05) & 0.16 & (0.61) \\
\cmidrule{1-10}
\multirow{3}{*}{\begin{tabular}{@{}c@{}}CMR-EI \\ \textit{Gauss, $\sigma$} \\ \textit{$[0.05,0.07]$}\end{tabular}} & K$\tau$ & 0.82 & (**) & 0.71 & (**) & 0.64 & (**) & 0.09 & (0.75) \\
& S$\rho$ & 0.90 & (**) & 0.81 & (**) & 0.80 & (*) & 0.17 & (0.61) \\
& P$r$ & 0.96 & (**) & 0.76 & (**) & 0.64 & (*) & 0.19 & (0.55) \\
\cmidrule{1-10}
\multirow{3}{*}{\begin{tabular}{@{}c@{}}CMR-EI \\ \textit{Gauss, $\sigma$} \\ \textit{$[0.07,0.1]$}\end{tabular}} & K$\tau$ & 0.78 & (**) & 0.67 & (**) & 0.67 & (**) & 0.12 & (0.65) \\
& S$\rho$ & 0.89 & (**) & 0.78 & (*) & 0.85 & (**) & 0.16 & (0.63) \\
& P$r$ & 0.96 & (**) & 0.77 & (**) & 0.67 & (*) & 0.22 & (0.49) \\
\cmidrule{1-10}
\multirow{3}{*}{\begin{tabular}{@{}c@{}}CMR-EI \\ \textit{Gauss, $\sigma$} \\ \textit{$[0.1,0.12]$}\end{tabular}} & K$\tau$ & 0.82 & (**) & 0.75 & (**) & 0.64 & (*) & 0.12 & (0.60) \\
& S$\rho$ & 0.92 & (**) & 0.83 & (**) & 0.81 & (**) & 0.16 & (0.65) \\
& P$r$ & 0.96 & (**) & 0.77 & (**) & 0.69 & (*) & 0.27 & (0.39) \\
\cmidrule{1-10}
\multirow{3}{*}{\begin{tabular}{@{}c@{}}CMR-EI \\ \textit{Gauss, $\sigma$} \\ \textit{$[0.12,0.15]$}\end{tabular}} & K$\tau$ & 0.78 & (**) & 0.75 & (**) & 0.64 & (*) & 0.21 & (0.38) \\
& S$\rho$ & 0.91 & (**) & 0.83 & (**) & 0.81 & (**) & 0.24 & (0.45) \\
& P$r$ & 0.93 & (**) & 0.76 & (**) & 0.70 & (*) & 0.28 & (0.38) \\
\cmidrule{1-10}
\multirow{3}{*}{\begin{tabular}{@{}c@{}}CMR-EI \\ \textit{Gauss, $\sigma$} \\ \textit{$[0.15,0.2]$}\end{tabular}} & K$\tau$ & 0.67 & (**) & 0.71 & (**) & 0.64 & (*) & 0.18 & (0.45) \\
& S$\rho$ & 0.85 & (**) & 0.82 & (**) & 0.81 & (**) & 0.22 & (0.47) \\
& P$r$ & 0.89 & (**) & 0.72 & (*) & 0.71 & (*) & 0.28 & (0.37) \\
\bottomrule
\end{tabular}
\label{tab:CMR_ei_mitochondria_UDA_gauss}
\end{subtable}%
\hfill
\begin{subtable}[t]{0.48\textwidth}
\centering
\tiny
\caption{AdaBN}
\setlength{\tabcolsep}{0.7pt}
\begin{tabular}{@{}c c
   >{\columncolor{GreyTable}}c
   >{\columncolor{GreyTable}}c
   c c
   >{\columncolor{GreyTable}}c
   >{\columncolor{GreyTable}}c
   c c@{}}
\toprule
\multirow{2}{*}{Metric} & {} & \multicolumn{2}{c}{EPFL} & \multicolumn{2}{c}{Hmito} & \multicolumn{2}{c}{Rmito} & \multicolumn{2}{c}{VNC} \\
&  &  \cellcolor{SecondaryColumnColor} & \cellcolor{SecondaryColumnColor}\textit{pval.} &  & \textit{pval.} &  \cellcolor{SecondaryColumnColor} & \cellcolor{SecondaryColumnColor}\textit{pval.} &  & \textit{pval.} \\
\cmidrule{1-10}
\multirow{3}{*}{\begin{tabular}{@{}c@{}}CMR-EI \\ \textit{Gauss, $\sigma$} \\ \textit{$[0.01,0.03]$}\end{tabular}} & K$\tau$ & 0.67 & (**) & 0.73 & (**) & 0.55 & (*) & 0.33 & (0.30) \\
& S$\rho$ & 0.80 & (**) & 0.85 & (**) & 0.71 & (*) & 0.45 & (0.25) \\
& P$r$ & 0.93 & (**) & 0.93 & (**) & 0.92 & (**) & 0.57 & (0.11) \\
\cmidrule{1-10}
\multirow{3}{*}{\begin{tabular}{@{}c@{}}CMR-EI \\ \textit{Gauss, $\sigma$} \\ \textit{$[0.03,0.05]$}\end{tabular}} & K$\tau$ & 0.70 & (**) & 0.73 & (**) & 0.61 & (**) & 0.33 & (0.21) \\
& S$\rho$ & 0.87 & (**) & 0.88 & (**) & 0.76 & (*) & 0.42 & (0.29) \\
& P$r$ & 0.89 & (**) & 0.92 & (**) & 0.89 & (**) & 0.62 & (0.07) \\
\cmidrule{1-10}
\multirow{3}{*}{\begin{tabular}{@{}c@{}}CMR-EI \\ \textit{Gauss, $\sigma$} \\ \textit{$[0.05,0.07]$}\end{tabular}} & K$\tau$ & 0.73 & (**) & 0.73 & (**) & 0.58 & (*) & 0.33 & (0.25) \\
& S$\rho$ & 0.88 & (**) & 0.88 & (**) & 0.74 & (**) & 0.42 & (0.26) \\
& P$r$ & 0.85 & (**) & 0.91 & (**) & 0.88 & (**) & 0.71 & (*) \\
\cmidrule{1-10}
\multirow{3}{*}{\begin{tabular}{@{}c@{}}CMR-EI \\ \textit{Gauss, $\sigma$} \\ \textit{$[0.07,0.1]$}\end{tabular}} & K$\tau$ & 0.70 & (**) & 0.73 & (**) & 0.55 & (*) & 0.39 & (0.16) \\
& S$\rho$ & 0.86 & (**) & 0.88 & (**) & 0.73 & (*) & 0.47 & (0.21) \\
& P$r$ & 0.76 & (**) & 0.91 & (**) & 0.87 & (**) & 0.71 & (*) \\
\cmidrule{1-10}
\multirow{3}{*}{\begin{tabular}{@{}c@{}}CMR-EI \\ \textit{Gauss, $\sigma$} \\ \textit{$[0.1,0.12]$}\end{tabular}} & K$\tau$ & 0.67 & (**) & 0.73 & (**) & 0.58 & (**) & 0.33 & (0.28) \\
& S$\rho$ & 0.83 & (**) & 0.88 & (**) & 0.76 & (**) & 0.42 & (0.27) \\
& P$r$ & 0.59 & (*) & 0.90 & (**) & 0.86 & (**) & 0.70 & (*) \\
\cmidrule{1-10}
\multirow{3}{*}{\begin{tabular}{@{}c@{}}CMR-EI \\ \textit{Gauss, $\sigma$} \\ \textit{$[0.12,0.15]$}\end{tabular}} & K$\tau$ & 0.67 & (**) & 0.73 & (**) & 0.61 & (**) & 0.39 & (0.16) \\
& S$\rho$ & 0.81 & (**) & 0.90 & (**) & 0.77 & (**) & 0.52 & (0.15) \\
& P$r$ & 0.44 & (0.15) & 0.89 & (**) & 0.85 & (**) & 0.71 & (*) \\
\cmidrule{1-10}
\multirow{3}{*}{\begin{tabular}{@{}c@{}}CMR-EI \\ \textit{Gauss, $\sigma$} \\ \textit{$[0.15,0.2]$}\end{tabular}} & K$\tau$ & 0.64 & (**) & 0.67 & (**) & 0.58 & (*) & 0.39 & (0.17) \\
& S$\rho$ & 0.76 & (*) & 0.85 & (**) & 0.76 & (**) & 0.52 & (0.15) \\
& P$r$ & 0.38 & (0.22) & 0.86 & (**) & 0.82 & (**) & 0.74 & (*) \\
\bottomrule
\end{tabular}
\label{tab:CMR_ei_mitochondria_UDA_gauss_adabn}
\end{subtable}
\label{tab:CMR_post_UDA}
\end{table}

\cref{tab:CMR_post_UDA} shows that the correlation score of our CMR approach remains very stable over a wide range of perturbation strengths. As noted in the main paper CMR ranking strongly correlates to F1-score ranking across three out of the four Mitochondria datasets, but struggles with the VNC dataset (but still beats the Transfer 
Score baseline see \cref{tab:finetuned_semantic}). We wanted to investigate this apparent underperformance on VNC further. 

Firstly, it is important to note that VNC is by far the smallest dataset thus reducing its statistical robustness. The entire VNC dataset only 20 slices of $589 \times 589$ pixels. As noted in \cref{sec:data_sem_EM}, due to the small dataset size in direct zero-shot application we purposely disregarded the source-to-source VNC model transfer and instead used the full 20 slices as test data for models trained on other sources. However, in post-UDA ranking analysis this is no longer possible as we must split the data into train/test splits so that models can be adapted on the target training set and then tested on the test set. Hence, for the post-UDA ranking the models transferred to VNC are assessed on only 2 $589 \times 589$ slices.

\begin{figure}[h]
    \centering
    \includegraphics[width=0.9\linewidth]{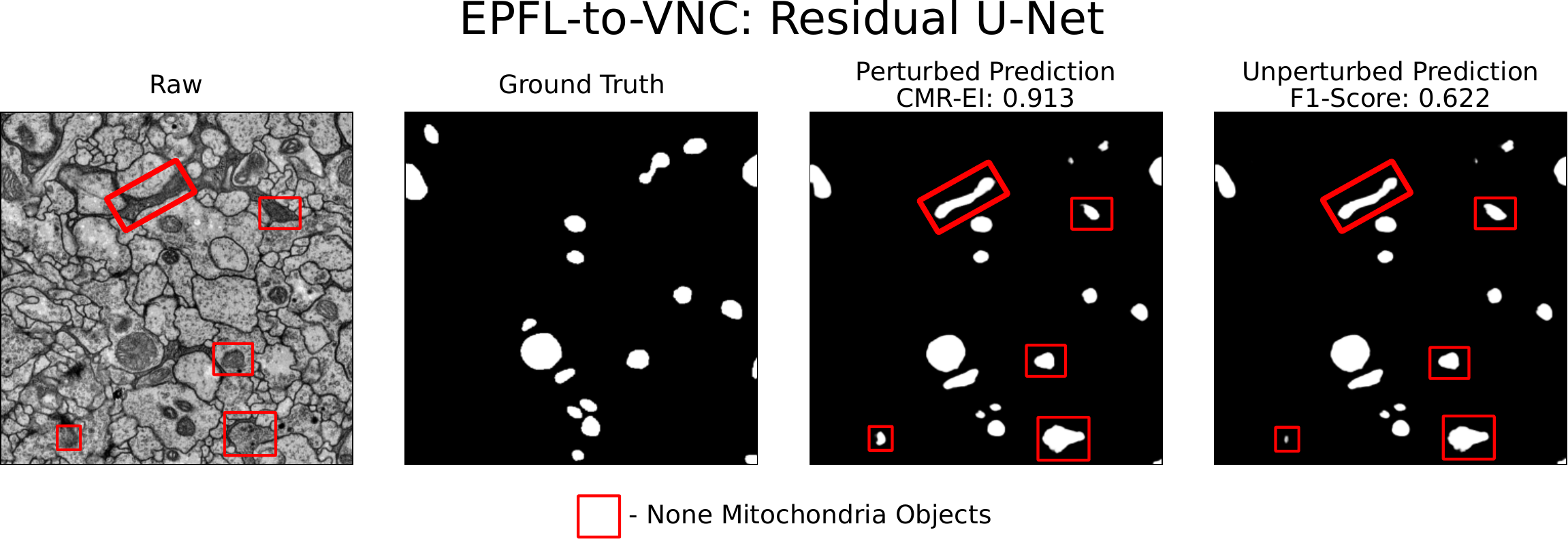}
    \caption{EPFL-to-VNC Mean Teacher UDA Residual U-Net class confusion.}
    \label{fig:post_UDA_class_confusion}
\end{figure}

However, when we look into the errors made by the CMR ranking for post-UDA transfer to the VNC dataset we see that models trained on the EPFL dataset seem to be outliers with unexpectedly high consistency scores despite low performance. When investigating this further it seemed there was class confusion, see \cref{fig:post_UDA_class_confusion}, in the predictions of the EPFL models, where `none mitochondria' objects are consistently but erroneously segmented by the transferred EPFL source models. This results in low performance F1-score, but high consistency score.

As noted in the main paper consistency regularisation in Mean Teacher training likely reinforces this class confusion between visually similar, but distinct structures present in EPFL and VNC, violating the assumption that source and target models address the same task. Hence, it is not surprising that CMR ranking after AdaBN UDA, which lacks a consistency regularisation component, performs better on the VNC target dataset, as shown in \cref{tab:CMR_ei_mitochondria_UDA_gauss_adabn}.

\FloatBarrier

\subsection{Transferability Metrics: Final-Finetuned Correlation}

For the supervised baseline methods compared against~\cite{nguyen_leep_2020,bao_information-theoretic_2022,yang_pick_2023,pandy_transferability_2021,ibrahim_newer_2023,li_ranking_2020,wang_how_2023,you_logme_2021} the goal of the transferability metrics is to predict the final supervised finetuned transfer performance of a model based on its initial pre-trained state. In the main paper we investigate the real world scenario of unsupervised model transfer, hence we are limited to assessing direct zero-shot application of pre-trained models to target data or assessing models after application of UDA approaches. We showed already in \cref{tab:semantic_seg_correlations} that, even with supervision, existing transfer metrics fail at direct zero-shot target performance ranking in the semantic setting. How about in the setting of predicting the final supervised finetuned performance of models? 

To investigate this we further trained a set of source models on the supervised training set of each of the mitochondria datasets in turn and then used the available ground truth to evaluate the performance of the fine-tuned models on the test set. Then using the pre-finetuning weights of the models we calculated all the base line transferability metrics and measured the correlation scores between pre-finetuned transferability metric and post-finetuned performance score. \cref{tab:transfer_metric_mitochondria_final_finetuned} shows the correlation scores split by target dataset, we can see that all the metrics fail to meaningfully correlate with final-finetuned performance. Further highlighting that existing transferability metrics cannot be reliably used for predicting the transfer of semantic segmentation models applied to biomedical data.

\begin{table}[tb]
\centering
\tiny
\caption{Transferability Metric correlation scores to final-finetuned F1-score performance ranking, Semantic Segmentation of Mitochondria with a range of Gaussian Input Perturbation Strengths. $|\mathcal{M}|=11$. \textit{pval.} $<$ 0.05 (*), \textit{pval.} $<$ 0.01 (**).}
\setlength{\tabcolsep}{3pt}
\begin{tabular}{@{}c c
   >{\columncolor{GreyTable}}c
   >{\columncolor{GreyTable}}c
   c c
   >{\columncolor{GreyTable}}c
   >{\columncolor{GreyTable}}c
   c c@{}}
\toprule
\multirow{2}{*}{Metric} & {} & \multicolumn{2}{c}{EPFL} & \multicolumn{2}{c}{Hmito} & \multicolumn{2}{c}{Rmito} & \multicolumn{2}{c}{VNC} \\
&  &  \cellcolor{SecondaryColumnColor} & \cellcolor{SecondaryColumnColor}\textit{pval.} &  & \textit{pval.} &  \cellcolor{SecondaryColumnColor} & \cellcolor{SecondaryColumnColor}\textit{pval.} &  & \textit{pval.} \\
\cmidrule{1-10}
\multirow{3}{*}{CCFV} & K$\tau$ & -0.16 & (0.57) & -0.42 & (0.10) & -0.53 & (*) & -0.27 & (0.25) \\
& S$\rho$ & -0.24 & (0.48) & -0.57 & (0.05) & -0.69 & (*) & -0.37 & (0.24) \\
& P$r$ & -0.25 & (0.46) & -0.52 & (0.10) & -0.50 & (0.12) & -0.40 & (0.19) \\
\cmidrule{1-10}
\multirow{3}{*}{GBC} & K$\tau$ & 0.31 & (0.21) & 0.20 & (0.47) & 0.45 & (0.07) & 0.24 & (0.34) \\
& S$\rho$ & 0.47 & (0.14) & 0.25 & (0.45) & 0.48 & (0.16) & 0.28 & (0.36) \\
& P$r$ & 0.24 & (0.47) & 0.53 & (0.10) & 0.41 & (0.22) & 0.45 & (0.14) \\
\cmidrule{1-10}
\multirow{3}{*}{LEEP} & K$\tau$ & 0.56 & (*) & -0.09 & (0.77) & 0.16 & (0.55) & 0.27 & (0.28) \\
& S$\rho$ & 0.72 & (*) & -0.04 & (0.92) & 0.36 & (0.31) & 0.41 & (0.20) \\
& P$r$ & 0.44 & (0.17) & 0.10 & (0.78) & 0.39 & (0.23) & 0.37 & (0.23) \\
\cmidrule{1-10}
\multirow{3}{*}{NLEEP} & K$\tau$ & 0.09 & (0.78) & -0.20 & (0.41) & 0.38 & (0.09) & 0.03 & (0.92) \\
& S$\rho$ & 0.15 & (0.66) & -0.21 & (0.52) & 0.45 & (0.17) & -0.03 & (0.92) \\
& P$r$ & 0.11 & (0.75) & 0.10 & (0.76) & 0.45 & (0.17) & -0.02 & (0.95) \\
\cmidrule{1-10}

\multirow{3}{*}{Hscore} & K$\tau$ & -0.02 & (1.00) & 0.27 & (0.26) & 0.45 & (0.06) & 0.15 & (0.49) \\
& S$\rho$ & 0.09 & (0.81) & 0.42 & (0.24) & 0.53 & (0.12) & 0.28 & (0.35) \\
& P$r$ & 0.20 & (0.56) & 0.21 & (0.54) & 0.41 & (0.20) & 0.17 & (0.60) \\
\cmidrule{1-10}
\multirow{3}{*}{RegHscore} & K$\tau$ & -0.02 & (1.00) & 0.24 & (0.37) & 0.45 & (0.06) & 0.15 & (0.58) \\
& S$\rho$ & 0.09 & (0.78) & 0.40 & (0.22) & 0.53 & (0.09) & 0.28 & (0.40) \\
& P$r$ & 0.20 & (0.56) & 0.21 & (0.54) & 0.41 & (0.21) & 0.17 & (0.60) \\
\cmidrule{1-10}
\multirow{3}{*}{LogME} & K$\tau$ & -0.13 & (0.59) & -0.02 & (0.98) & 0.35 & (0.17) & 0.00 & (1.00) \\
& S$\rho$ & 0.05 & (0.84) & -0.01 & (0.97) & 0.41 & (0.22) & 0.13 & (0.66) \\
& P$r$ & 0.16 & (0.64) & -0.08 & (0.82) & 0.03 & (0.94) & 0.29 & (0.36) \\
\cmidrule{1-10}
\multirow{3}{*}{NCTI} & K$\tau$ & -0.16 & (0.54) & -0.09 & (0.75) & 0.16 & (0.57) & 0.00 & (1.00) \\
& S$\rho$ & -0.21 & (0.54) & -0.24 & (0.51) & 0.26 & (0.45) & -0.01 & (0.99) \\
& P$r$ & 0.08 & (0.82) & -0.24 & (0.49) & 0.37 & (0.26) & 0.04 & (0.89) \\
\bottomrule
\end{tabular}
\vspace{-12pt}
\label{tab:transfer_metric_mitochondria_final_finetuned}
\end{table}
\FloatBarrier

\section{Instance Segmentation}

In the following section we investigate ranking of instance segmentation models, each of the tables \cref{tab:CMR_ars_cells_gauss,tab:CMR_ars_cells_dropout,tab:CMR_ars_nuclei_gauss,tab:CMR_ars_nuclei_gamma,tab:CMR_ars_nuclei_brt,tab:CMR_ars_nuclei_ctr,tab:CMR_ars_nuclei_dropout,tab:CMR_ars_covid_if_gauss} shows the correlation scores between our CMR-ARS metric and instance segmentation performance metrics  for sets of models applied directly to a single target dataset, across a range of perturbation strengths. In the main paper, due to space constraints, we show the correlation scores for a single perturbation strength averaged over the set of target datasets. We investigate the dependency of CMR on the strength of perturbation applied for both input and feature space perturbations. The tables show that CMR's performance is stable across a wide range of perturbations.

\subsection{Cells Perturbation Sweep}

For cell instance segmentation we investigated ranking a diverse set of domain specialist models with $|\mathcal{M}|=8$. The models were trained on three different source datasets FlyWing~\cite{leal-taixe_benchmark_2019}, Ovules~\cite{wolny_accurate_2020} and PNAS~\cite{willis_cell_2016} and include 2D U-Nets (trained both with and without augmentations) and Residual 2D U-Nets.

We investigated applying additive Gaussian noise as an input perturbation between the strengths $\sigma = 0.01 - 0.2$. We show the results for CMR-ARS in \cref{tab:CMR_ars_cells_gauss}.

For feature space perturbation we investigated applying spatial Test-time DropOut (TTD) equally to all layers of a network, where the strength of the perturbation is controlled by the proportion of feature maps dropped at each layer $p_{d}$. We investigated DropOut proportions in the range $p_{d}= 0.001 - 0.1$. We show results for for both CMR-ARS in \cref{tab:CMR_ars_cells_dropout}.

\begin{table}[h]
\centering
\tiny
\caption{Per-dataset CMR-ARS correlation scores to mAP@[0.5:0.95] ranking for Instance Segmentation of Cells across a range of Gaussian Input Perturbation Strengths. $|\mathcal{M}|=8$. \textit{pval.} $<$ 0.05 (*), \textit{pval.} $<$ 0.01 (**).}
\setlength{\tabcolsep}{3pt}
\begin{tabular}{@{}c c
   >{\columncolor{GreyTable}}c
   >{\columncolor{GreyTable}}c
   c c
   >{\columncolor{GreyTable}}c
   >{\columncolor{GreyTable}}c@{}}
\toprule
\multirow{2}{*}{Metric} & {} & \multicolumn{2}{c}{FlyWing} & \multicolumn{2}{c}{Ovules} & \multicolumn{2}{c}{PNAS} \\
&  &  \cellcolor{SecondaryColumnColor} & \cellcolor{SecondaryColumnColor}\textit{pval.} &  & \textit{pval.} &  \cellcolor{SecondaryColumnColor} & \cellcolor{SecondaryColumnColor}\textit{pval.} \\
\cmidrule{1-8}
\multirow{3}{*}{\begin{tabular}{@{}c@{}}CMR-ARS \\ \textit{Gauss, $\sigma$} \\ \textit{$[0.01, 0.05]$}\end{tabular}} & K$\tau$ & 0.64 & (*) & 0.43 & (0.16) & 0.93 & (**) \\
& S$\rho$ & 0.79 & (*) & 0.60 & (0.16) & 0.98 & (**) \\
& P$r$ & 0.79 & (*) & 0.95 & (**) & 0.83 & (*) \\
\cmidrule{1-8}
\multirow{3}{*}{\begin{tabular}{@{}c@{}}CMR-ARS \\ \textit{Gauss, $\sigma$} \\ \textit{$[0.05, 0.1]$}\end{tabular}} & K$\tau$ & 0.57 & (0.07) & 0.43 & (0.21) & 0.71 & (*) \\
& S$\rho$ & 0.76 & (*) & 0.57 & (0.15) & 0.83 & (*) \\
& P$r$ & 0.91 & (**) & 0.87 & (**) & 0.91 & (**) \\
\cmidrule{1-8}
\multirow{3}{*}{\begin{tabular}{@{}c@{}}CMR-ARS \\ \textit{Gauss, $\sigma$} \\ \textit{$[0.1, 0.15]$}\end{tabular}} & K$\tau$ & 0.57 & (0.07) & 0.64 & (*) & 0.57 & (0.07) \\
& S$\rho$ & 0.76 & (*) & 0.79 & (*) & 0.76 & (*) \\
& P$r$ & 0.94 & (**) & 0.88 & (**) & 0.89 & (**) \\
\cmidrule{1-8}
\multirow{3}{*}{\begin{tabular}{@{}c@{}}CMR-ARS \\ \textit{Gauss, $\sigma$} \\ \textit{$[0.15, 0.2]$}\end{tabular}} & K$\tau$ & 0.50 & (0.11) & 0.57 & (0.06) & 0.57 & (0.06) \\
& S$\rho$ & 0.74 & (0.05) & 0.74 & (0.05) & 0.76 & (*) \\
& P$r$ & 0.94 & (**) & 0.85 & (*) & 0.90 & (**) \\
\cmidrule{1-8}
\multirow{3}{*}{\begin{tabular}{@{}c@{}}CMR-ARS \\ \textit{Gauss, $\sigma$} \\ \textit{$[0.2, 0.25]$}\end{tabular}} & K$\tau$ & 0.57 & (0.07) & 0.64 & (*) & 0.86 & (**) \\
& S$\rho$ & 0.76 & (*) & 0.79 & (*) & 0.93 & (**) \\
& P$r$ & 0.94 & (**) & 0.88 & (**) & 0.87 & (**) \\
\bottomrule
\end{tabular}
\label{tab:CMR_ars_cells_gauss}
\end{table}
\begin{table}[thb]
\centering
\tiny
\caption{CMR-ARS Correlation scores for Instance Segmentation of Cells across a range of DropOut Perturbation Strengths. $|\mathcal{M}|=8$. \textit{pval.} $<$ 0.05 (*), \textit{pval.} $<$ 0.01 (**).}
\setlength{\tabcolsep}{3pt}
\begin{tabular}{@{}c c
   >{\columncolor{GreyTable}}c
   >{\columncolor{GreyTable}}c
   c c
   >{\columncolor{GreyTable}}c
   >{\columncolor{GreyTable}}c@{}}
\toprule
\multirow{2}{*}{Metric} & {} & \multicolumn{2}{c}{FlyWing} & \multicolumn{2}{c}{Ovules} & \multicolumn{2}{c}{PNAS} \\
&  &  \cellcolor{SecondaryColumnColor} & \cellcolor{SecondaryColumnColor}\textit{pval.} &  & \textit{pval.} &  \cellcolor{SecondaryColumnColor} & \cellcolor{SecondaryColumnColor}\textit{pval.} \\
\cmidrule{1-8}
\multirow{3}{*}{\begin{tabular}{@{}c@{}}CMR-ARS \\ \textit{DropOut} \\ \textit{($p_{d}=$ 0.001)}\end{tabular}} & K$\tau$ & 0.64 & (*) & 0.50 & (0.11) & 0.71 & (*) \\
& S$\rho$ & 0.81 & (*) & 0.74 & (0.06) & 0.83 & (*) \\
& P$r$ & 0.93 & (**) & 0.82 & (*) & 0.81 & (*) \\
\cmidrule{1-8}
\multirow{3}{*}{\begin{tabular}{@{}c@{}}CMR-ARS \\ \textit{DropOut} \\ \textit{($p_{d}=$ 0.005)}\end{tabular}} & K$\tau$ & 0.71 & (*) & 0.50 & (0.14) & 0.71 & (*) \\
& S$\rho$ & 0.83 & (*) & 0.74 & (*) & 0.83 & (*) \\
& P$r$ & 0.93 & (**) & 0.83 & (*) & 0.86 & (*) \\
\cmidrule{1-8}
\multirow{3}{*}{\begin{tabular}{@{}c@{}}CMR-ARS \\ \textit{DropOut} \\ \textit{($p_{d}=$ 0.01)}\end{tabular}} & K$\tau$ & 0.57 & (0.07) & 0.50 & (0.11) & 0.71 & (*) \\
& S$\rho$ & 0.76 & (*) & 0.74 & (0.06) & 0.83 & (*) \\
& P$r$ & 0.94 & (**) & 0.84 & (*) & 0.88 & (**) \\
\cmidrule{1-8}
\multirow{3}{*}{\begin{tabular}{@{}c@{}}CMR-ARS \\ \textit{DropOut} \\ \textit{($p_{d}=$ 0.02)}\end{tabular}} & K$\tau$ & 0.64 & (*) & 0.64 & (*) & 0.79 & (*) \\
& S$\rho$ & 0.79 & (*) & 0.79 & (*) & 0.90 & (*) \\
& P$r$ & 0.94 & (**) & 0.87 & (**) & 0.92 & (**) \\
\cmidrule{1-8}
\multirow{3}{*}{\begin{tabular}{@{}c@{}}CMR-ARS \\ \textit{DropOut} \\ \textit{($p_{d}=$ 0.03)}\end{tabular}} & K$\tau$ & 0.50 & (0.11) & 0.79 & (*) & 0.64 & (*) \\
& S$\rho$ & 0.74 & (0.07) & 0.90 & (*) & 0.81 & (*) \\
& P$r$ & 0.95 & (**) & 0.89 & (**) & 0.91 & (**) \\
\cmidrule{1-8}
\multirow{3}{*}{\begin{tabular}{@{}c@{}}CMR-ARS \\ \textit{DropOut} \\ \textit{($p_{d}=$ 0.04)}\end{tabular}} & K$\tau$ & 0.50 & (0.12) & 0.57 & (0.07) & 0.64 & (*) \\
& S$\rho$ & 0.74 & (*) & 0.76 & (*) & 0.81 & (*) \\
& P$r$ & 0.95 & (**) & 0.79 & (*) & 0.86 & (*) \\
\cmidrule{1-8}
\multirow{3}{*}{\begin{tabular}{@{}c@{}}CMR-ARS \\ \textit{DropOut} \\ \textit{($p_{d}=$ 0.05)}\end{tabular}} & K$\tau$ & 0.50 & (0.10) & 0.57 & (0.05) & 0.71 & (*) \\
& S$\rho$ & 0.74 & (0.06) & 0.71 & (0.05) & 0.83 & (*) \\
& P$r$ & 0.95 & (**) & 0.83 & (*) & 0.88 & (**) \\
\cmidrule{1-8}
\multirow{3}{*}{\begin{tabular}{@{}c@{}}CMR-ARS \\ \textit{DropOut} \\ \textit{($p_{d}=$ 0.1)}\end{tabular}} & K$\tau$ & 0.57 & (0.06) & 0.43 & (0.17) & 0.57 & (*) \\
& S$\rho$ & 0.76 & (*) & 0.60 & (0.14) & 0.71 & (0.07) \\
& P$r$ & 0.95 & (**) & 0.78 & (*) & 0.88 & (**) \\
\bottomrule
\end{tabular}
\label{tab:CMR_ars_cells_dropout}
\end{table}
\FloatBarrier

\subsection{Nuclei Perturbation Sweep}

For nuclei instance segmentation we trained a set of domain specialist models with $|\mathcal{M}|=5$. The models all have a 2D U-Net architecture and were trained on five datasets; BBBC039~\cite{ljosa_annotated_2012}, HeLaCytoNuc~\cite{de_helacytonuc_2024}, Hoechst~\cite{arvidsson_annotated_2023}, 
S\_BIAD895/SB-895~\cite{von_chamier_democratising_2021} and SBIAD1410~\cite{hawkins_rescu-nets_2024}. The source models were then transferred to the test sets of four target datasets BBBC039~\cite{ljosa_annotated_2012}, Hoechst~\cite{arvidsson_annotated_2023}, 
S\_BIAD895~\cite{von_chamier_democratising_2021} and S\_BIAD634/SB-634~\cite{kromp_annotated_2020}.

We investigated applying a range of input augmentations; additive Gaussian noise, Gamma correction, Brightness adjustment and Contrast adjustment (as defined in \cref{sec:perturbations}). We investigated additive Gaussian noise with strengths in the range $\sigma = 0.0 - 0.2$, Gamma correction with strengths in the range $\gamma = 0.8 - 1.2$, where $\gamma = 1.0$ equals no adjustment, Brightness adjustment with strengths in the range $\theta_{B}=0.0-0.2$ and Contrast adjustment with strengths in the range $\theta_{C} = 0.8 -1.2$, where $\theta_{C} = 1.0$ equals no adjustment. We show the results for both CMR-ARS (\cref{tab:CMR_ars_nuclei_gauss,tab:CMR_ars_nuclei_gamma,tab:CMR_ars_nuclei_ctr,tab:CMR_ars_nuclei_brt})

For feature space perturbation we investigated applying spatial DropOut only to the bottle neck layer of networks, where the strength of the perturbation is controlled by the proportion of feature maps dropped at each layer $p_{d}$. We investigated DropOut proportions in the range $p_{d}= 0.05 - 0.5$. We show results for for CMR-ARS in \cref{tab:CMR_ars_nuclei_dropout}. 

We again observe stable correlation performance across a wide range of perturbation strengths. Although, as discussed in the main paper, \cref{sec:model_perturbation}, care should be taken to ensure perturbations remain `tolerable' in the extreme perturbation case, too strong of a perturbation can lead to a loss of correlation between consistency based metrics and transfer performance. For example, the last row of \cref{tab:CMR_ars_nuclei_dropout} shows the strongest level of DropOut ($p_{d}=0.5$) applied can cause the CMR-ARS metric to become unstable on S\_BIAD634.

\begin{table}[htb]
\centering
\tiny
\caption{CMR-ARS Correlation scores to mAP@[0.5:0.95] ranking for Instance Segmentation of Nuclei across a range of Input Perturbation Strengths. For each ranking experiment $|\mathcal{M}|=5$. Perturbation strength are controlled by sampling $\sigma,\; \theta_B,\; \theta_C \;\text{and} \; \gamma$ from the noted ranges. \textit{pval.} $<$ 0.05 (*), \textit{pval.} $<$ 0.01 (**).}

\begin{subtable}[t]{0.48\textwidth}
\centering
\tiny
\caption{Gaussian Input Perturbation}
\setlength{\tabcolsep}{0.5pt}
\begin{tabular}{@{}c c
   >{\columncolor{GreyTable}}c
   >{\columncolor{GreyTable}}c
   c c
   >{\columncolor{GreyTable}}c
   >{\columncolor{GreyTable}}c
   c c@{}}
\toprule
\multirow{2}{*}{Metric} & {} & \multicolumn{2}{c}{BBBC039} & \multicolumn{2}{c}{Hoechst} & \multicolumn{2}{c}{SB-895} & \multicolumn{2}{c}{SB-634} \\
&  &  \cellcolor{SecondaryColumnColor} & \cellcolor{SecondaryColumnColor}\textit{pval.} &  & \textit{pval.} &  \cellcolor{SecondaryColumnColor} & \cellcolor{SecondaryColumnColor}\textit{pval.} &  & \textit{pval.} \\
\cmidrule{1-10}
\multirow{3}{*}{\begin{tabular}{@{}c@{}}CMR-ARS \\ \textit{Gauss, $\sigma$} \\ \textit{$[0.0, 0.05]$}\end{tabular}} & K$\tau$ & 0.80 & (0.08) & 0.80 & (0.11) & 0.33 & (0.75) & 0.80 & (0.10) \\
& S$\rho$ & 0.90 & (0.10) & 0.90 & (0.08) & 0.60 & (0.42) & 0.90 & (0.08) \\
& P$r$ & 0.99 & (**) & 0.85 & (0.07) & 0.47 & (0.53) & 0.93 & (*) \\
\cmidrule{1-10}
\multirow{3}{*}{\begin{tabular}{@{}c@{}}CMR-ARS \\ \textit{Gauss, $\sigma$} \\ \textit{$[0.05, 0.1]$}\end{tabular}} & K$\tau$ & 0.20 & (0.80) & 0.80 & (0.09) & 0.33 & (0.75) & 1.00 & (*) \\
& S$\rho$ & 0.50 & (0.42) & 0.90 & (0.08) & 0.40 & (0.75) & 1.00 & (*) \\
& P$r$ & 0.72 & (0.17) & 0.89 & (*) & 0.99 & (*) & 0.80 & (0.10) \\
\cmidrule{1-10}
\multirow{3}{*}{\begin{tabular}{@{}c@{}}CMR-ARS \\ \textit{Gauss, $\sigma$} \\ \textit{$[0.1, 0.2]$}\end{tabular}} & K$\tau$ & 0.40 & (0.54) & 1.00 & (*) & 0.67 & (0.33) & 0.80 & (0.09) \\
& S$\rho$ & 0.60 & (0.36) & 1.00 & (*) & 0.80 & (0.33) & 0.90 & (0.08) \\
& P$r$ & 0.50 & (0.39) & 0.93 & (*) & 0.94 & (0.06) & 0.79 & (0.11) \\
\bottomrule
\end{tabular}
\label{tab:CMR_ars_nuclei_gauss}
\end{subtable}%
\hfill
\begin{subtable}[t]{0.48\textwidth}
\centering
\tiny
\caption{Brightness Input Perturbation}
\setlength{\tabcolsep}{0.5pt}
\begin{tabular}{@{}c c
   >{\columncolor{GreyTable}}c
   >{\columncolor{GreyTable}}c
   c c
   >{\columncolor{GreyTable}}c
   >{\columncolor{GreyTable}}c
   c c@{}}
\toprule
\multirow{2}{*}{Metric} & {} & \multicolumn{2}{c}{BBBC039} & \multicolumn{2}{c}{Hoechst} & \multicolumn{2}{c}{SB-895} & \multicolumn{2}{c}{SB-634} \\
&  &  \cellcolor{SecondaryColumnColor} & \cellcolor{SecondaryColumnColor}\textit{pval.} &  & \textit{pval.} &  \cellcolor{SecondaryColumnColor} & \cellcolor{SecondaryColumnColor}\textit{pval.} &  & \textit{pval.} \\
\cmidrule{1-10}
\multirow{3}{*}{\begin{tabular}{@{}c@{}}CMR-ARS \\ \textit{Brt, $\theta_{B}$} \\ \textit{$[0.0, 0.05]$}\end{tabular}} & K$\tau$ & 0.60 & (0.22) & 0.60 & (0.26) & 0.00 & (1.00) & 0.80 & (0.10) \\
& S$\rho$ & 0.70 & (0.25) & 0.70 & (0.21) & 0.20 & (0.92) & 0.90 & (0.09) \\
& P$r$ & 0.80 & (0.10) & 0.83 & (0.08) & 0.77 & (0.23) & 0.72 & (0.17) \\
\cmidrule{1-10}
\multirow{3}{*}{\begin{tabular}{@{}c@{}}CMR-ARS \\ \textit{Brt, $\theta_{B}$} \\ \textit{$[0.05, 0.1]$}\end{tabular}} & K$\tau$ & 0.40 & (0.46) & 0.80 & (0.11) & 0.33 & (0.75) & 0.80 & (0.07) \\
& S$\rho$ & 0.60 & (0.32) & 0.90 & (0.10) & 0.40 & (0.75) & 0.90 & (0.10) \\
& P$r$ & 0.69 & (0.19) & 0.92 & (*) & 0.80 & (0.20) & 0.84 & (0.08) \\
\cmidrule{1-10}
\multirow{3}{*}{\begin{tabular}{@{}c@{}}CMR-ARS \\ \textit{Brt, $\theta_{B}$} \\ \textit{$[0.1, 0.2]$}\end{tabular}} & K$\tau$ & 0.60 & (0.23) & 0.80 & (0.09) & 0.67 & (0.33) & 0.40 & (0.49) \\
& S$\rho$ & 0.70 & (0.23) & 0.90 & (0.10) & 0.80 & (0.33) & 0.50 & (0.40) \\
& P$r$ & 0.81 & (0.10) & 0.85 & (0.07) & 0.99 & (*) & 0.43 & (0.47) \\
\bottomrule
\end{tabular}
\label{tab:CMR_ars_nuclei_brt}
\end{subtable}

\vspace{1em}

\begin{subtable}[t]{0.48\textwidth}
\centering
\tiny
\caption{Contrast Input Perturbation}
\setlength{\tabcolsep}{0.5pt}
\begin{tabular}{@{}c c
   >{\columncolor{GreyTable}}c
   >{\columncolor{GreyTable}}c
   c c
   >{\columncolor{GreyTable}}c
   >{\columncolor{GreyTable}}c
   c c@{}}
\toprule
\multirow{2}{*}{Metric} & {} & \multicolumn{2}{c}{BBBC039} & \multicolumn{2}{c}{Hoechst} & \multicolumn{2}{c}{SB-895} & \multicolumn{2}{c}{SB-634} \\
&  &  \cellcolor{SecondaryColumnColor} & \cellcolor{SecondaryColumnColor}\textit{pval.} &  & \textit{pval.} &  \cellcolor{SecondaryColumnColor} & \cellcolor{SecondaryColumnColor}\textit{pval.} &  & \textit{pval.} \\
\cmidrule{1-10}
\multirow{3}{*}{\begin{tabular}{@{}c@{}}CMR-ARS \\ \textit{Ctr, $\theta_{C}$} \\ \textit{$[0.8, 0.9]$}\end{tabular}} & K$\tau$ & 0.80 & (0.09) & 1.00 & (*) & 0.00 & (1.00) & 0.80 & (0.08) \\
& S$\rho$ & 0.90 & (0.09) & 1.00 & (*) & -0.20 & (0.92) & 0.90 & (0.07) \\
& P$r$ & 0.85 & (0.07) & 0.94 & (*) & 0.18 & (0.82) & 0.83 & (0.08) \\
\cmidrule{1-10}
\multirow{3}{*}{\begin{tabular}{@{}c@{}}CMR-ARS \\ \textit{Ctr, $\theta_{C}$} \\ \textit{$[0.9, 0.95]$}\end{tabular}} & K$\tau$ & 0.80 & (0.09) & 1.00 & (*) & 0.67 & (0.33) & 0.80 & (0.10) \\
& S$\rho$ & 0.90 & (0.09) & 1.00 & (*) & 0.80 & (0.33) & 0.90 & (0.09) \\
& P$r$ & 0.96 & (*) & 0.99 & (**) & 0.52 & (0.48) & 0.87 & (0.05) \\
\cmidrule{1-10}
\multirow{3}{*}{\begin{tabular}{@{}c@{}}CMR-ARS \\ \textit{Ctr, $\theta_{C}$} \\ \textit{$[1.05, 1.1]$}\end{tabular}} & K$\tau$ & 0.74 & (0.13) & 1.00 & (*) & 0.67 & (0.33) & 0.60 & (0.25) \\
& S$\rho$ & 0.82 & (0.13) & 1.00 & (*) & 0.80 & (0.33) & 0.80 & (0.13) \\
& P$r$ & 0.99 & (**) & 0.95 & (*) & 0.58 & (0.42) & 0.80 & (0.10) \\
\cmidrule{1-10}
\multirow{3}{*}{\begin{tabular}{@{}c@{}}CMR-ARS \\ \textit{Ctr, $\theta_{C}$} \\ \textit{$[1.1, 1.2]$}\end{tabular}} & K$\tau$ & 0.80 & (0.11) & 0.60 & (0.23) & 0.33 & (0.75) & 0.80 & (0.07) \\
& S$\rho$ & 0.90 & (0.10) & 0.70 & (0.20) & 0.40 & (0.75) & 0.90 & (0.09) \\
& P$r$ & 0.83 & (0.08) & 0.81 & (0.10) & 0.38 & (0.62) & 0.90 & (*) \\
\bottomrule
\end{tabular}
\label{tab:CMR_ars_nuclei_ctr}
\end{subtable}%
\hfill
\begin{subtable}[t]{0.48\textwidth}
\centering
\tiny
\caption{Gamma Input Perturbation}
\setlength{\tabcolsep}{0.5pt}
\begin{tabular}{@{}c c
   >{\columncolor{GreyTable}}c
   >{\columncolor{GreyTable}}c
   c c
   >{\columncolor{GreyTable}}c
   >{\columncolor{GreyTable}}c
   c c@{}}
\toprule
\multirow{2}{*}{Metric} & {} & \multicolumn{2}{c}{BBBC039} & \multicolumn{2}{c}{Hoechst} & \multicolumn{2}{c}{SB-895} & \multicolumn{2}{c}{SB-634} \\
&  &  \cellcolor{SecondaryColumnColor} & \cellcolor{SecondaryColumnColor}\textit{pval.} &  & \textit{pval.} &  \cellcolor{SecondaryColumnColor} & \cellcolor{SecondaryColumnColor}\textit{pval.} &  & \textit{pval.} \\
\cmidrule{1-10}
\multirow{3}{*}{\begin{tabular}{@{}c@{}}CMR-ARS \\ \textit{Gamma, $\gamma$} \\ \textit{$[0.8, 0.9]$}\end{tabular}} & K$\tau$ & 0.40 & (0.42) & 0.80 & (0.08) & -0.33 & (0.75) & 0.40 & (0.49) \\
& S$\rho$ & 0.60 & (0.37) & 0.90 & (0.10) & -0.40 & (0.75) & 0.60 & (0.36) \\
& P$r$ & 0.62 & (0.26) & 0.82 & (0.09) & 0.06 & (0.94) & 0.64 & (0.25) \\
\cmidrule{1-10}
\multirow{3}{*}{\begin{tabular}{@{}c@{}}CMR-ARS \\ \textit{Gamma, $\gamma$} \\ \textit{$[0.9, 0.95]$}\end{tabular}} & K$\tau$ & 0.40 & (0.50) & 0.80 & (0.08) & 0.33 & (0.75) & 0.80 & (0.08) \\
& S$\rho$ & 0.60 & (0.37) & 0.90 & (0.10) & 0.40 & (0.75) & 0.90 & (0.07) \\
& P$r$ & 0.54 & (0.35) & 0.89 & (*) & 0.58 & (0.42) & 0.76 & (0.14) \\
\cmidrule{1-10}
\multirow{3}{*}{\begin{tabular}{@{}c@{}}CMR-ARS \\ \textit{Gamma, $\gamma$} \\ \textit{$[1.05, 1.1]$}\end{tabular}} & K$\tau$ & 0.60 & (0.22) & 0.80 & (0.09) & 0.00 & (1.00) & 0.80 & (0.08) \\
& S$\rho$ & 0.70 & (0.22) & 0.90 & (0.08) & -0.20 & (0.92) & 0.90 & (0.09) \\
& P$r$ & 0.94 & (*) & 0.84 & (0.07) & 0.44 & (0.56) & 0.82 & (0.09) \\
\cmidrule{1-10}
\multirow{3}{*}{\begin{tabular}{@{}c@{}}CMR-ARS \\ \textit{Gamma, $\gamma$} \\ \textit{$[1.1, 1.2]$}\end{tabular}} & K$\tau$ & 0.40 & (0.53) & 1.00 & (*) & 0.33 & (0.75) & 0.40 & (0.45) \\
& S$\rho$ & 0.60 & (0.37) & 1.00 & (*) & 0.40 & (0.75) & 0.60 & (0.34) \\
& P$r$ & 0.71 & (0.18) & 0.87 & (0.05) & 0.68 & (0.32) & 0.76 & (0.14) \\
\bottomrule
\end{tabular}
\label{tab:CMR_ars_nuclei_gamma}
\end{subtable}

\end{table}
\begin{table}[th]
\centering
\tiny
\caption{CMR-ARS Correlation scores to mAP@[0.5:0.95] ranking for Instance Segmentation of Nuclei across a range of DropOut Perturbation Strengths, controlled by $p_d$. For each ranking experiment $|\mathcal{M}|=5$. \textit{pval.} $<$ 0.05 (*), \textit{pval.} $<$ 0.01 (**).}
\setlength{\tabcolsep}{3pt}
\begin{tabular}{@{}c c
   >{\columncolor{GreyTable}}c
   >{\columncolor{GreyTable}}c
   c c
   >{\columncolor{GreyTable}}c
   >{\columncolor{GreyTable}}c
   c c@{}}
\toprule
\multirow{2}{*}{Metric} & {} & \multicolumn{2}{c}{BBBC039} & \multicolumn{2}{c}{Hoechst} & \multicolumn{2}{c}{S\_BIAD895} & \multicolumn{2}{c}{S\_BIAD634} \\
&  &  \cellcolor{SecondaryColumnColor} & \cellcolor{SecondaryColumnColor}\textit{pval.} &  & \textit{pval.} &  \cellcolor{SecondaryColumnColor} & \cellcolor{SecondaryColumnColor}\textit{pval.} &  & \textit{pval.} \\
\cmidrule{1-10}
\multirow{3}{*}{\begin{tabular}{@{}c@{}}CMR-ARS \\ \textit{DropOut} \\ \textit{($p_{d}=$ 0.05)}\end{tabular}} & K$\tau$ & 0.40 & (0.43) & 0.80 & (0.08) & 1.00 & (0.08) & 0.80 & (0.12) \\
& S$\rho$ & 0.60 & (0.32) & 0.90 & (0.08) & 1.00 & (0.08) & 0.90 & (0.08) \\
& P$r$ & 0.83 & (0.08) & 0.90 & (*) & 0.93 & (0.07) & 0.88 & (0.05) \\
\cmidrule{1-10}
\multirow{3}{*}{\begin{tabular}{@{}c@{}}CMR-ARS \\ \textit{DropOut} \\ \textit{($p_{d}=$ 0.1)}\end{tabular}} & K$\tau$ & 0.40 & (0.51) & 1.00 & (*) & 1.00 & (0.08) & 0.80 & (0.08) \\
& S$\rho$ & 0.50 & (0.44) & 1.00 & (*) & 1.00 & (0.08) & 0.90 & (0.11) \\
& P$r$ & 0.87 & (0.05) & 0.86 & (0.06) & 0.93 & (0.07) & 0.91 & (*) \\
\cmidrule{1-10}
\multirow{3}{*}{\begin{tabular}{@{}c@{}}CMR-ARS \\ \textit{DropOut} \\ \textit{($p_{d}=$ 0.2)}\end{tabular}} & K$\tau$ & 0.40 & (0.54) & 1.00 & (*) & 1.00 & (0.08) & 0.80 & (0.09) \\
& S$\rho$ & 0.50 & (0.45) & 1.00 & (*) & 1.00 & (0.08) & 0.90 & (0.07) \\
& P$r$ & 0.95 & (*) & 0.93 & (*) & 0.72 & (0.28) & 0.78 & (0.12) \\
\cmidrule{1-10}
\multirow{3}{*}{\begin{tabular}{@{}c@{}}CMR-ARS \\ \textit{DropOut} \\ \textit{($p_{d}=$ 0.3)}\end{tabular}} & K$\tau$ & 0.60 & (0.25) & 1.00 & (*) & 1.00 & (0.08) & 0.80 & (0.08) \\
& S$\rho$ & 0.70 & (0.23) & 1.00 & (*) & 1.00 & (0.08) & 0.90 & (0.10) \\
& P$r$ & 0.87 & (0.05) & 0.89 & (*) & 0.76 & (0.24) & 0.82 & (0.09) \\
\cmidrule{1-10}
\multirow{3}{*}{\begin{tabular}{@{}c@{}}CMR-ARS \\ \textit{DropOut} \\ \textit{($p_{d}=$ 0.4)}\end{tabular}} & K$\tau$ & 0.60 & (0.22) & 1.00 & (*) & 0.67 & (0.33) & 1.00 & (*) \\
& S$\rho$ & 0.70 & (0.24) & 1.00 & (*) & 0.80 & (0.33) & 1.00 & (*) \\
& P$r$ & 0.95 & (*) & 0.91 & (*) & 0.91 & (0.09) & 0.96 & (*) \\
\cmidrule{1-10}
\multirow{3}{*}{\begin{tabular}{@{}c@{}}CMR-ARS \\ \textit{DropOut} \\ \textit{($p_{d}=$ 0.5)}\end{tabular}} & K$\tau$ & 0.60 & (0.22) & 0.80 & (0.10) & 0.67 & (0.33) & 0.00 & (1.00) \\
& S$\rho$ & 0.70 & (0.25) & 0.90 & (0.07) & 0.80 & (0.33) & 0.00 & (1.00) \\
& P$r$ & 0.98 & (**) & 0.80 & (0.10) & 0.89 & (0.11) & 0.49 & (0.40) \\
\bottomrule
\end{tabular}
\label{tab:CMR_ars_nuclei_dropout}
\end{table}
\hfill

\FloatBarrier

\subsection{Covid\_IF External Models Perturbation Sweep}

In the main paper in \cref{fig:Instance_seg_covid} we also ranked the predicted target performance of a set of publicly available models~\cite{pachitariu_cellpose-sam_2025,archit_segment_2025,pape_microscopy-based_2021} for instance cell segmentation on the Covid\_IF~\cite{pape_microscopy-based_2021} dataset. The models all have very different instantiation processes, but through calculating the ranking metric in output segmentation space we are able to directly compare the varied set of models. The Micro-SAM models~\cite{archit_segment_2025} perform Automatic Instance Segmentation (AIS), which utilises an additionally trained decoder to predict a three channel output: the distance to the object centre, the distance to the object boundary and foreground probabilities. The predictions are then instantiated using seeded watershed. Cellpose-SAM~\cite{pachitariu_cellpose-sam_2025} use a custom decoder to directly predict vector flow fields, which can then be converted to instance segmentations. `Powerful-chipmunk' is a specialist U-Net model taken from the Bioimage model zoo~\cite{ouyang_bioimage_2022}, the model is trained specifically on the Covid\_IF data and predicts object foreground and boundary predictions, which are then converted to instances using watershed and GASP.

In \cref{fig:Instance_seg_covid} from the main paper we show the correlation between CMR-ARS and mSA performance for a single additive Gaussian noise perturbation strength. In \cref{tab:CMR_ars_covid_if_gauss} we investigate a range of Gaussian noise strengths $\sigma=0.01-0.05$.

\begin{table}[th]
\centering
\tiny
\caption{CMR-ARS Correlation scores to mSA for Instance Segmentation of Covid\_IF across a range of Gaussian Input Perturbation Strengths, controlled by $\sigma$ sampled from the noted range. For each ranking experiment $|\mathcal{M}|=5$. \textit{pval.} $<$ 0.05 (*), \textit{pval.} $<$ 0.01 (**).}
\setlength{\tabcolsep}{3pt}
\begin{tabular}{@{}c c
   >{\columncolor{GreyTable}}c
   >{\columncolor{GreyTable}}c@{}}
\toprule
\multirow{2}{*}{Metric} & {} & \multicolumn{2}{c}{Covid\_IF} \\
&  &  \cellcolor{SecondaryColumnColor} & \cellcolor{SecondaryColumnColor}\textit{pval.} \\
\cmidrule{1-4}
\multirow{3}{*}{\begin{tabular}{@{}c@{}}CMR-ARS \\ \textit{Gauss, $\sigma$} \\ \textit{$[0.01, 0.02]$}\end{tabular}} & K$\tau$ & 0.80 & (0.09) \\
& S$\rho$ & 0.90 & (0.08) \\
& P$r$ & 0.86 & (0.06) \\
\cmidrule{1-4}
\multirow{3}{*}{\begin{tabular}{@{}c@{}}CMR-ARS \\ \textit{Gauss, $\sigma$} \\ \textit{$[0.02, 0.03]$}\end{tabular}} & K$\tau$ & 1.00 & (*) \\
& S$\rho$ & 1.00 & (*) \\
& P$r$ & 0.93 & (*) \\
\cmidrule{1-4}
\multirow{3}{*}{\begin{tabular}{@{}c@{}}CMR-ARS \\ \textit{Gauss, $\sigma$} \\ \textit{$[0.03, 0.04]$}\end{tabular}} & K$\tau$ & 1.00 & (*) \\
& S$\rho$ & 1.00 & (*) \\
& P$r$ & 0.91 & (*) \\
\cmidrule{1-4}
\multirow{3}{*}{\begin{tabular}{@{}c@{}}CMR-ARS \\ \textit{Gauss, $\sigma$} \\ \textit{$[0.04, 0.05]$}\end{tabular}} & K$\tau$ & 1.00 & (*) \\
& S$\rho$ & 1.00 & (*) \\
& P$r$ & 0.93 & (*) \\
\bottomrule
\end{tabular}
\label{tab:CMR_ars_covid_if_gauss}
\end{table}

\subsection{Generalist Models: Cells vs Nuclei Prediction}

\begin{figure}[t]
    \centering
    \begin{subfigure}[b]{0.48\linewidth}
        \centering
        \includegraphics[width=\linewidth]{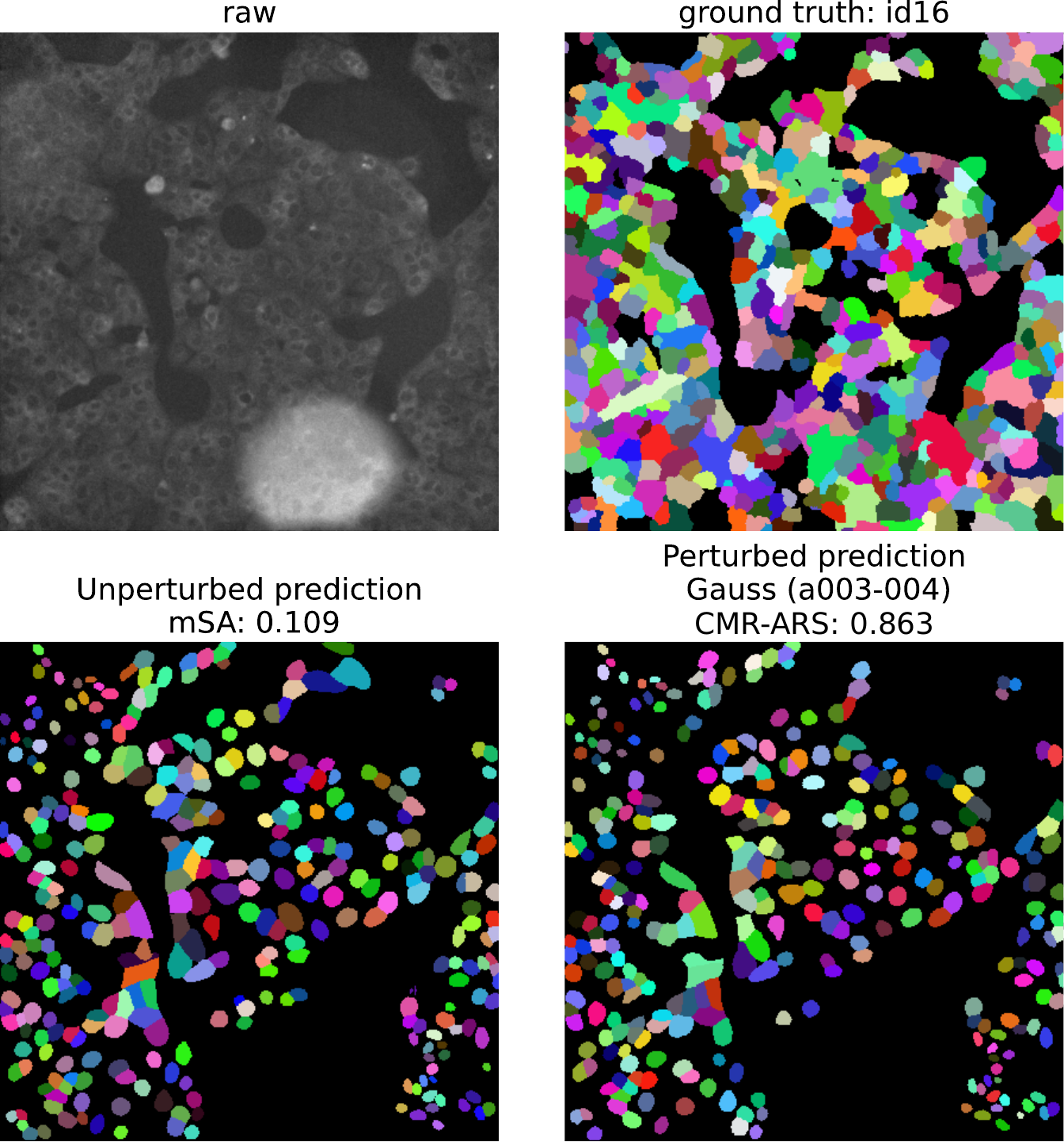}
        \caption{ID 16}
        \label{fig:cells_vs_nuclei_1}
    \end{subfigure}
    \hfill
    \begin{subfigure}[b]{0.48\linewidth}
        \centering
        \includegraphics[width=\linewidth]{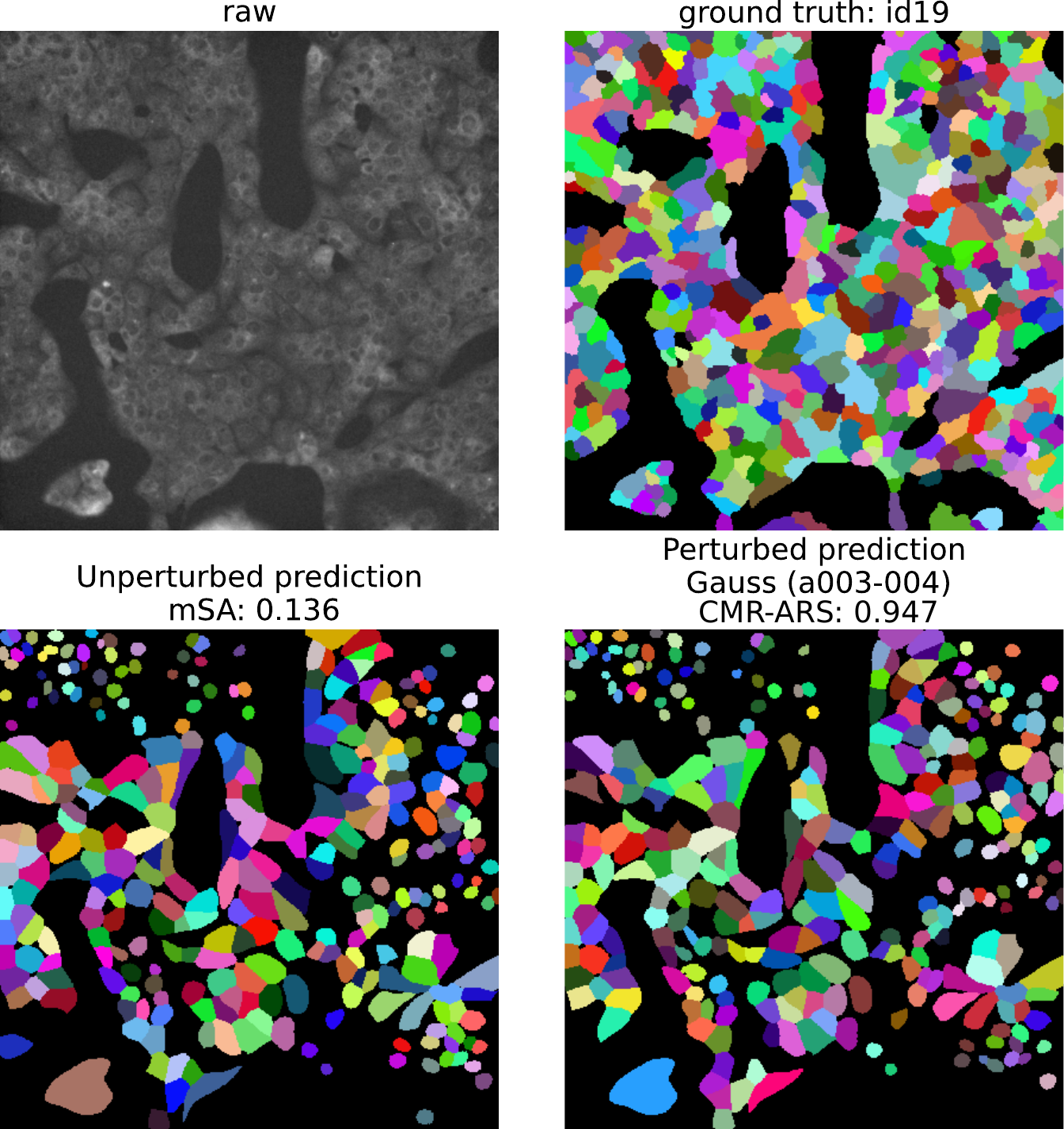}
        \caption{ID 19}
        \label{fig:cells_vs_nuclei_2}
    \end{subfigure}
    \caption{Cells vs Nuclei Prediction of Cellpose-SAM.}
    \label{fig:cells_vs_nuclei_combined}
\end{figure}

A key assumption of the proposed consistency based approach is that the source and target tasks align, i.e, if a model was trained for nuclei segmentation then it should only be evaluated for nuclei segmentation. For generalist models like $\mu\text{SAM}$~\cite{archit_segment_2025} and Cellpose-SAM~\cite{pachitariu_cellpose-sam_2025}, this assumption may not fully hold in all cases. The models are finetuned for both cell and nuclei and have no automated mechanism to control which structure is segmented. In datasets like Covid\_IF, a channel can implicitly contain information about both classes. We observed that Cellpose-SAM and $\mu\text{SAM}$ occasionally switch to segmenting nuclei instead of cells (\cref{fig:cells_vs_nuclei_1,fig:cells_vs_nuclei_2}) particularly in `challenging' sections of the image. It is hard to quantify is this always due to a `mode' switch to nuclei segmentation or just poor cell segmentation performance, however it seems clear that at least in some cases the models segment small objects (much smaller than cells) that clearly correspond to the nuclei information present in the serum channel.  This `mode' switch reduces cell segmentation performance, but not consistency scores, as predictions remain internally stable. However, despite some examples of `off-task' segmentation in general Cellpose-SAM and $\mu$SAM are strong performing cell segmentation models and in the majority of cases are able to stay on task given a good raw input signal. Hence, we can see in \cref{fig:Instance_seg_covid,tab:CMR_ars_covid_if_gauss} that our CMR based ranking of these large generalist models and specialist BMZ model~\cite{ouyang_bioimage_2022} strongly correlates with the true target performance ranking.

An important note is that the ranking in \cref{fig:Instance_seg_covid} does not show the general superiority of any one model, but is a target dataset specific ranking of automatic segmentation performance. One factor to keep in mind is that the Covid\_IF dataset that we tested ranking on was explicitly not included in the training data of $\mu SAM$, while inclusion for Cellpose-SAM is uncertain. However, this represents realistic ranking scenarios where models will vary their performance across different target datasets due to differences in source-to-target distribution and model architecture. Hence the need for target specific model ranking as provided by our method.

\clearpage

\end{document}